\documentclass[12pt]{hkputhesis}

\usepackage{booktabs}
\usepackage{graphicx}
\usepackage{amsmath}
\usepackage{amssymb}
\usepackage{amsthm}
\usepackage{csquotes}
\usepackage{multirow}
\usepackage{mathtools}
\usepackage{algpseudocode}

\algrenewcommand\algorithmicrequire{\textbf{Input:}}
\algrenewcommand\algorithmicensure{\textbf{Output:}}

\usepackage{pifont}
\usepackage{soul}
\usepackage{colortbl}
\usepackage{multirow}
\usepackage{ulem}
\usepackage{arydshln}
\usepackage{natbib}
\usepackage{wrapfig}
\usepackage{caption}
\usepackage{hyperref}
\hypersetup{
    hidelinks % 隐藏链接边框
}
\usepackage{cleveref}
\usepackage[caption=false,font=normalsize,labelfont=sf,textfont=sf,labelformat=simple]{subfig}

\usepackage{subcaption}

\usepackage[linesnumbered,ruled,vlined]{algorithm2e}

\begin{document}

\title{Towards Robust Multimodal Learning in the Open World}

\author{Fushuo HUO}
\dept{Department of Computing}
\degree{Doctor of Philosophy}
\degshort{PhD}
%\degree{Master of Philosophy}
%\degshort{MPhil}

\frontpageyear{2025} % Year of current submission

\submityear{2025} % Year of initial submission
\submitmonth{April} % Month of initial submission

% \initexam{Initial Submission for Examination Purpose} % Please delete this link when submitting the final version

\frontmatter

\prefacesection{Abstract}

The rapid evolution of machine learning has propelled neural networks to unprecedented success across diverse domains. In particular, multimodal learning has emerged as a transformative paradigm, leveraging complementary information from heterogeneous data streams (e.g., text, vision, audio) to advance contextual reasoning and intelligent decision-making. Despite these advancements, current neural network-based models often fall short in open-world environments characterized by inherent unpredictability, where unpredictable environmental composition dynamics, incomplete modality inputs, and spurious distributions relations critically undermine system reliability. While humans naturally adapt to such dynamic, ambiguous scenarios, artificial intelligence systems exhibit stark limitations in robustness, particularly when processing multimodal signals under real-world complexity. This study investigates the fundamental challenge of multimodal learning robustness in open-world settings, aiming to bridge the gap between controlled experimental performance and practical deployment requirements. 
Here, we study the multimodal learning robustness in the open world settings:

(1). Humans can extrapolate new concepts from previously learned multi-modal knowledge. 
This ability is known as compositional generalization, while neural networks have deficiencies in compositional generalization robustness, struggling to reliably handle unseen compositions due to rigid feature representations and over-reliance on training data biases.
(2). Humans can seamlessly infer unimodal inputs based on memorized contextual multimodal information, with robust inference in the absence of modality. However, neural networks hardly achieve satisfactory results when inferring unimodal inputs, based on integrated multimodal information.
(3). With the development of large language models (LLMs), large-scale multimodal large language models (MLLMs), especially large vision language models (LVLMs), have demonstrated expressing comprehensive abilities, approaching or even surpassing human abilities. However, most LVLMs are derived from LLMs by instruction tuning on multimodal datasets. LVLMs usually have the strong language modality prior or statistical bias to LLMs, which is one of the main reasons that arises the significant challenge problem known as `hallucination', even when queried by simple questions. 

In summary, 
we study above three problems to improve \textit{class-level and modality-level multimodal robustness} in terms of composition gneralization robustness (i.e., class-level), modality missing robustness (i.e., modality-level), and modality prior robustness (i.e., modality-level). Concretely, In Chapter \ref{Chapter3}, we propose a novel Progressive Cross-primitive Compatibility (ProCC) network, mimicking the human learning progress of recognizing the multimodal compositions to improve the modality composition ability. In Chapter \ref{Chapter4}, we propose the customized crossmodal knowledge distillation (C$^2$KD) to inherit multimodal knowledge during the pre-training period, and enhance the inference robustness when missing some modalities. In Chapter \ref{Chapter5}, we propose the train-free decoding strategy to alleviate language modality prior of LVLMs to mitigate the hallucination issues while not compromising general abilities of foundation models. Extensive experimental evaluations and ablation studies show the performance advantages of our works with provable advances in robustness abilities for multiple modalities.

\prefacesection{Publications arising from the thesis}
\begin{itemize}
    \item \textbf{Fushuo Huo}, Wenchao Xu, Song Guo, Jingcai Guo, Haozhao Wang, Ziming Liu, Xiaocheng Lu, ``ProCC: Progressive Cross-Primitive Compatibility for Open-World Compositional Zero-Shot Learning'', Proceedings of the AAAI Conference on Artificial Intelligence (AAAI), 2024.
    \item \textbf{Fushuo Huo}, Wenchao Xu, Jingcai Guo, Haozhao Wang, Song Guo, ``C$^2$KD: Bridging the Modality Gap for Cross-Modal Knowledge Distillation'', Proceedings of the IEEE/CVF Conference on Computer Vision and Pattern Recognition (CVPR), 2024.
    \item \textbf{Fushuo Huo}, Wenchao Xu, Zhong Zhang, Haozhao Wang, Zhicheng Chen, Peilin Zhao, ``Self-Introspective Decoding: Alleviating Hallucinations for Large Vision-Language Models'', International Conference on Learning Representations (ICLR), 2025.
    \item \textbf{Fushuo Huo}, Wenchao Xu, Song Guo, Jingcai Guo, Haozhao Wang, Yunfeng Fan, ``Non-exemplar Online Class-Incremental Continual Learning via Dual-Prototype Self-Augment and Refinement'', Proceedings of the AAAI Conference on Artificial Intelligence (AAAI), 2024.
    \item \textbf{Fushuo Huo}, Ziming Liu, Jingcai Guo, Wenchao Xu, Song Guo, ``UTDNet: A Unified Triplet Decoder Network for Multimodal Salient Object Detection'', Neural Networks (NN), 2024.
    \item Yunfeng Fan, Wenchao Xu, Haozhao Wang, \textbf{Fushuo Huo}, Jinyu Chen, and Song Guo, ``Overcome Modal Bias in Multi-modal Federated Learning via Balanced Modality Selection'', European Conference on Computer Vision (ECCV), 2024.
    \item Ziming Liu, Song Guo, Xiaocheng Lu, Jingcai Guo, Jiewei Zhang, Yue Zeng, and \textbf{Fushuo Huo}, ``(ML)$^2$P-Encoder: On Exploration of Channel-Class Correlation for Multi-Label Zero-Shot Learning'', Proceedings of the IEEE/CVF Conference on Computer Vision and Pattern Recognition (CVPR), 2023. 
    \item Jingcai Guo, Song Guo, Qihua Zhou, Ziming Liu, Xiaocheng Lu, and \textbf{Fushuo Huo}, ``Graph knows unknowns: Reformulate zero-shot learning as sample-level graph recognition'', Proceedings of the AAAI Conference on Artificial Intelligence (AAAI), 2023.
    \item Ziming Liu, Song Guo, Jingcai Guo, Yuanyuan Xu, and \textbf{Fushuo Huo}, ``Towards Unbiased Multi-Label Zero-Shot Learning with Pyramid and Semantic Attention'', IEEE Transactions on Multimedia (IEEE TMM), 2022.
\end{itemize}
			
\prefacesection{Acknowledgments}

It is highly privileged for me to take this opportunity to express my sincere gratitude
to those who have helped and supported me on my way to pursuing my Ph.D. degree.

First of all, I am immensely grateful to my supervisors, Prof. Song Guo and Prof. Wenchao Xu, for their invaluable guidance, expertise, and patience. Their mentorship and insightful feedback have been instrumental in shaping the direction of my research. I would also like to express my sincere gratitude to Prof. Bin Xiao, Prof. Jingcai Guo, and Dr. Haozhao Wang for their valuable input and constructive suggestions, which have greatly spurred me on the academic journey. Thanks to all my groupmates in the Pervasive Intelligence Lab (PEILAB). I will never forget the happy and hard times we undergo together. I am also immensely grateful to my visiting university's supervisors, Prof. Dacheng Tao and Prof. Baosheng Yu. They give me the opportunity to study at the Nanyang Technological University and inspire me a lot in many ongoing topics. Meanwhile, thanks to all the groupmates in the Generative AI Lab at the Nanyang Technological University. I will never forget the happy sports and nervous deadline times. 
I would also like to express my gratitude to Dr. Peilin Zhao, Dr. Zhong Zhang, and other friends at Tencent for spending a fulfilling and happy time there. 
Also,  I would like to thank Prof. WANG Cong and Prof. GONG Shimin for serving on my Ph.D. defense as external committees, and thank Prof. LUO Xiapu Daniel for serving on my Ph.D. defense as the BoE chair. 
Lastly, thanks to all the anonymous reviewers, whether the scoring was ACCEPT or REJECTION, I learned a lot from the submission and rebuttal periods. Thanks for all the ACCEPT and REJECTION!

I would like to express my heartfelt gratitude to my parents, who have struggled to maintain the family and have wholeheartedly supported me in pursuing my master's and doctoral degrees. Their resilience and persistence inspire me to overcome various setbacks on the academic journey. I am truly fortunate to have a mother in my life who, though physically fragile, possesses an iron will. Her unwavering support has profoundly shaped the person I am today. I would also like to thank my lovely girlfriend, Qian Zhang, whose smile has soothed my soul in the face of adversity. Thanks for my family's patience, understanding, and support. Without their support, I could not achieve this goal and accomplishment.

%% Dedication is optional
%\prefacesection{Dedication}
%Here is the dedication which is optional.

\mainmatter

% \afterpreface

\chapter{Introduction} % Main chapter title

\label{Chapter1} % For referencing the chapter elsewhere, use \ref{Chapter1} 

The pursuit of robust multimodal learning systems capable of operating in open-world environments represents a pivotal challenge at the frontier of artificial intelligence (AI). As artificial intelligent systems transition from controlled laboratory settings to real-world deployment in domains ranging from autonomous driving to AI-assisted healthcare, their ability to process heterogeneous data streams (e.g., visual, textual, auditory) with human-like adaptability becomes mission-critical. Although multimodal learning has shown remarkable success in leveraging complementary cross-modal information for tasks like visual question answering and multimodal sentiment analysis, current approaches remain fundamentally constrained by three existential limitations when confronted with open-world dynamics: (1) brittleness to novel concept compositions from known modality primitives, (2) catastrophic performance degradation under partial modality availability, and (3) systemic hallucinations induced by imbalanced modality priors in large multimodal foundation models. 
In this chapter, we first introduce the overview of our research problems in section \ref{section1.1}. Next, we describe the challenges of this research topic in Section \ref{section1.2}. Then, we present the sketch of our research framework in Section \ref{section1.3}. After that, we present the main contributions of this thesis in Section \ref{section1.4}. Finally, we give the overall organization of the thesis in Section \ref{section1.5}

\section{Overview}
\label{section1.1}

Multimodal learning represents a transformative paradigm in artificial intelligence that aims to process and integrate information from diverse data modalities (e.g., text, images, audio, video, sensor data) to mimic human-like perception and decision-making \cite{multimodal_survey, multimodal_survey23}. Unlike unimodal systems that operate on isolated data types, multimodal learning leverages the complementary strengths of heterogeneous signals to enhance contextual understanding, improve inference accuracy, and enable robust performance in real-world scenarios. This interdisciplinary field sits at the intersection of computer vision, natural language processing, and signal processing, driven by the recognition that human cognition inherently synthesizes multisensory inputs for holistic reasoning.

The concept of open-world settings represents a paradigm shift in machine learning, moving beyond the constraints of traditional closed-world assumptions where models operate within predefined, static environments with fully observed data distributions. In contrast, open-world settings reflect the inherent complexity and unpredictability of real-world scenarios, where systems must contend with dynamic data streams, unseen concept compositions, partial or corrupted inputs, and evolving contextual relationships \cite{open1, open2}. This framework is particularly critical for multimodal learning systems, as real-world applications, from autonomous robotics \cite{dai2024think} to healthcare diagnostics \cite{medical_1, medical_2}, demand adaptability to novel or even poor situations that defy the tidy boundaries of laboratory-trained models.

The robustness of multimodal learning systems from controlled laboratory environments to open-world deployment exposes fundamental limitations in current methodologies. Three interrelated challenges, compositional generalization robustness (class level), modality missing fragility (modality level), and uncontrollable hallucinations (modality level), emerge as critical barriers to reliable open-world multimodal intelligence. 
These three issues are interconnected facets of a core problem: current multimodal systems lack robustness mechanisms to dynamically balance modality-specific evidence with cross-modal causal relationships in open environments.

\section{Overall Challenges}
\label{section1.2}

The transition from controlled experimental settings to open-world deployment exposes multimodal learning systems to a spectrum of challenges that defy traditional algorithmic assumptions. These challenges stem from the inherent unpredictability, heterogeneity, and dynamic nature of real-world environments, demanding paradigm shifts in model design and evaluation. Below, we dissect the core challenges:
(1). Human-like reasoning requires extrapolating to novel compositions of known concepts when adapting to new knowledge, while neural networks might not easily generalize to new compositions composed by multimodal primitives (i.e., objects and attributes). However, humans can extrapolate new concepts from previously learned modality primitives. For instance, if the people are taught what the \emph{fried chip} and \emph{toasted bread} are, most of them can recognize the \emph{fried bread} immediately. This ability is known as \emph{compositional generalization} \cite{concept}, which is one of the ultimate targets for artificial intelligence.
In this report, such a task is formulated as \emph{Compositional Zero-Shot Learning} (CZSL). Concretely, the training set contains images with corresponding multi-modal descriptions (primitives), i.e., state and object. The model is expected to recognize unseen compositions based on known primitives, which is non-trivial because object and state are in the multi-modal formations and semantically tangled, i.e.,  objects in different states often have different appearances, and states can vary greatly conditioned on different objects. The major challenge behind the CZSL lies in how to model the interactions between state and object primitives and extrapolate seen compositions to unseen ones. 
(2). Real-world systems must operate under partial, asynchronous, or corrupted modality inputs, a stark contrast to curated datasets with aligned, complete data. For instance, autonomous vehicles may lose LiDAR signals during heavy rain while relying solely on cameras, or healthcare algorithms might face missing modality information during emergency diagnostics. This thesis focuses on developing the multimodal knowledge transfer (i.e., crossmodal knowledge distillation (CMKD)) to distill the heterogeneous modality information to another modality, which are not systematically explored. Consequently, the multimodal learning system might not degrade much when some modality is missing or even compared to all modalities available.
(3). Recently, in the era of foundation models, with the development of scaling law theory \cite{kaplan2020scaling} and Graphics Processing Unit (GPU) devices, multimodal learning has evolved into the paradigm of ``pre-training on massive datasets and fine-tuning in downstream fields". The advent of Multimodal Large Language Models (MLLMs), especially Large Vision-Language Models (LVLMs), derived from their Large Language Model (LLM) foundations, introduce a critical challenge in open-world multimodal learning: inherent modality bias, where most LVLMs are derived from LLMs with strong language modality prior \cite{vcd, icd} that dominates cross-modal reasoning, undermining robustness in dynamic, unpredictable environments. For example, given a picture of a black rotting banana, LVLMs will usually recognize the picture as ``yellow” when asked for its color. The sticky modality-level prior stems from LVLMs pretraining strategies. This thesis aims to alleviate the modality prior to enabling LVLMs to generate trustworthy answers.

\section{Research Framework}
\label{section1.3}

\begin{figure*}[t]
\centering
\includegraphics[width=0.85\textwidth]{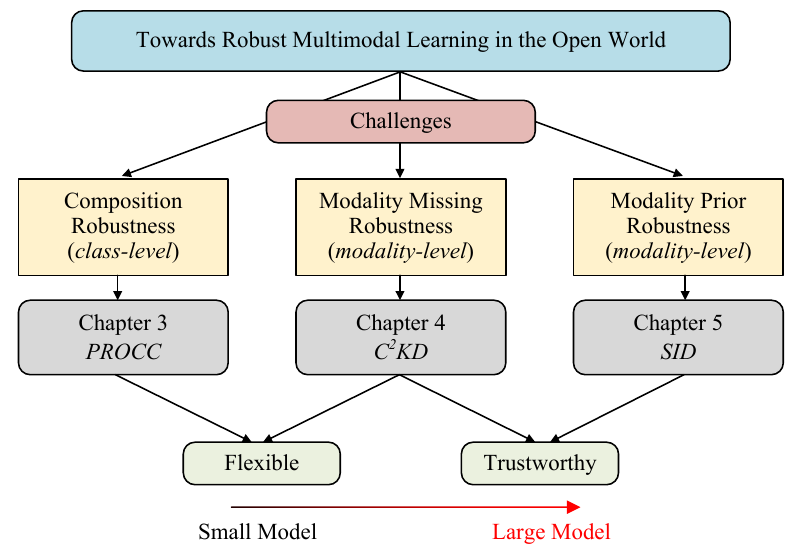}
\caption{\textbf{Research framework of this thesis.} We organize the positioning of this thesis within the field of robust multimodal learning in the open world. We classify the challenges into class-level and modality-level robustness and illustrate the contributions we focus on for each chapter.}
\label{c1_framework}
\end{figure*}

Our thesis aims to solve the above challenges and propose new frameworks for robust multimodal learning. The structural outline of the thesis is shown in Figure \ref{c1_framework}.

As shown in Figure \ref{c1_framework}, we categorize our works into two base categories in the robust multimodal learning, i.e., \textit{class-level} and \textit{modality-level} robustness. Furthermore, the evolution of the multimodal learning robustness from \textit{small-scale} to \textit{large-scale} model. 

Concretely, current approaches are fundamentally constrained by three existential limitations when faced with open-world dynamics: (1) brittleness to novel concept compositions from known modality primitives, (2) catastrophic performance degradation when only partial modalities are available, and (3) systemic hallucinations caused by imbalanced modality priors in large multimodal foundation models.
% 
% 
% In this thesis, we begin by providing an overview of our research background in Chapter \ref{Chapter2}. Then, in Chapter \ref{Chapter3}, we then introduce the novel Progressive Cross-primitive Compatibility (ProCC) network, mimicking the human learning progress of recognizing multimodal compositions. In Chapter \ref{Chapter4}, we propose the customized crossmodal knowledge distillation (C$^2$KD) to inherit multimodal knowledge during the pre-training period, and enhance the inference robustness when missing some modalities. In Chapter \ref{Chapter5}, we propose the train-free decoding strategy to alleviate language modality prior to mitigating the hallucination issues while not compromising the general abilities of LVLMs.
% 
% 
In this thesis, we begin by providing an overview of our research background in Chapter \ref{Chapter2}. Chapter \ref{Chapter3} introduces the Progressive Cross-Primitive Compatibility (ProCC) network, a framework inspired by human cognitive processes for learning multimodal compositions. By modeling cross-primitive dependencies through a curriculum-driven approach, ProCC enhances compositional generalization, enabling robust recognition of unseen object-state pairs in open-world scenarios. Chapter \ref{Chapter4} addresses modality-missing robustness via Customized Crossmodal Knowledge Distillation (C$^2$KD), which preserves cross-modal synergies during pretraining and transfers them to unimodal inference. This ensures consistent performance even when critical modalities are absent, bridging the gap between multimodal training and real-world deployment. In Chapter \ref{Chapter5}, we tackle hallucination biases in large vision-language models (LVLMs) by proposing a training-free decoding strategy, Self-Introspective Decoding (SID), which suppresses language-prior dominance without fine-tuning or compromising foundational capabilities. SID aligns model outputs with visual evidence while maintaining generative fluency.

\section{Thesis Contribution}
\label{section1.4}

 We briefly summarize the contribution of this thesis as follows:

\textbf{1. Enhancing Modality Composition Generalization Robustness.}
The first contribution addresses the challenge of recognizing novel compositions of state-object modalities in open-world scenarios (OW-CZSL), where no prior knowledge of valid compositions exists. To tackle this, the thesis introduces the Progressive Cross-Primitive Compatibility (ProCC) framework. By mimicking the human learning process, ProCC employs the Cross-Primitive Compatibility (CPC) module to explicitly model interactions between state and object features using trainable memory units, eliminating reliance on external knowledge. A progressive training paradigm further refines these interactions in an easy-to-hard manner, effectively handling partial supervision (pCZSL) where labels are incomplete. This approach achieves state-of-the-art performance across benchmarks, significantly improving generalization to unseen compositions while filtering invalid cross-modal correlations, thus enhancing \textit{class-level} robustness in dynamic environments.  

\textbf{2. Ensuring Robustness Under Modality Missing.}  
The second contribution targets the problem of modality imbalance and misalignment in cross-modal knowledge distillation (CMKD), which often degrades performance when modalities are missing during inference. The proposed Customized Crossmodal Knowledge Distillation (C$^2$KD) method bridges these gaps through the dual strategy: On-the-Fly Selection Distillation (OFSD) filters misaligned samples using Kendall Rank Correlation (KRC) metric, while bidirectional distillation between teacher-student proxies preserves cross-modal knowledge. By dynamically adapting to modality gaps, C$^2$KD ensures the inheritable crossmodal knowledge during the CMKD. Consequently, C$^2$KD outperforms traditional knowledge distillation methods on audio-visual, image-text, and RGB-depth tasks. This innovation ensures reliable performance in real-world applications, such as sensor failures or resource-constrained settings, by maintaining robustness even when critical modalities are absent, thereby addressing \textit{modality-level} reliability.  

\textbf{3. Balancing Modality Priors to Mitigate Hallucinations.}
The third contribution confronts the hallucination problem in Large Vision-Language Models (LVLMs), where over-reliance on language priors leads to factually inconsistent outputs. The Self-Introspective Decoding (SID) strategy introduces the Context and Text-aware Token Selection (CT$^2$S) mechanism to adaptively prune low-importance vision tokens in early decoder layers, amplifying vision-text association errors for contrastive suppression. This train-free approach reduces hallucinations by 12–20$\%$ on metrics like POPE and CHAIR while cutting inference costs by 30$\%$ compared to methods like VCD \cite{vcd} and ICD \cite{icd}. Crucially, SID preserves LVLMs’ general abilities, as evidenced by strong MME and MMBench scores. By rebalancing modality priors without compromising functionality, SID advances \textit{modality-level} robustness, ensuring trustworthy outputs in open-world deployment.

In summary, these contributions form a cohesive framework for robust multimodal learning: ProCC ensures generalization to unseen compositions, C$^2$KD safeguards against missing modalities, and SID mitigates modality priors. This triad equips multimodal learning systems to operate reliably in dynamic, real-world environments where compositions are novel, inputs are incomplete, and modality dominance varies, marking a significant stride toward trustworthy, open-world multimodal learning artificial intelligence.

\section{Thesis Organization}
\label{section1.5}

Thesis Organization Summary
This thesis is structured into six chapters that systematically address the critical challenges of robust multimodal learning in open-world environments, progressing from problem formulation to methodological innovation and validation:

\textbf{Chapter \ref{Chapter1}} establishes the research foundation, delineating the core challenges of compositional generalization robustness, modality missing robustness, and hallucination mitigation in multimodal models. It outlines the overarching research framework, key contributions, and thesis roadmap.

\textbf{Chapter \ref{Chapter2}} (Background) provides a comprehensive review of foundational concepts, including:
composition problem formulation, Compositional Zero-Shot Learning (CZSL), and its open-world limitations; unimodal knowledge distillation and Cross-modal knowledge distillation (CMKD); multimodal large language models, decoding strategy in LLMs, and hallucinations issues in Large Vision-Language Models (LVLMs).

\textbf{Chapter \ref{Chapter3}} (ProCC: Progressive Cross-Primitive Compatibility) introduces a novel framework to enhance compositional generalization robustness. Firstly, in Chapter \ref{c3.1}, detailed challenges and motivations are illustrated. Then, we propose the Progressive Cross-Primitive Compatibility network aligns visual and semantic primitives through curriculum learning in Chapter \ref{sec:pistis_model}. Chapter \ref{c3.3} validates ProCC in three widely-use datasets including UT-Zappos, MIT-States, and C-GQA under various settings. Detailed ablation studies confirm the relations of each modules of ProCC.

\textbf{Chapter \ref{Chapter4}} (C$^2$KD: Customized Cross-modal Knowledge Distillation) tackles modality missing inference via a customized crossmodal distillation framework. Firstly, in Chapter \ref{c4.1}, detailed challenges of knowledge distillation across modalities and motivations to develop crossmodal knowledge distillation methods are illustrated. Then, in Chpater \ref{c4.2}, we comprehensively revisit traditional knowledge distillation effectiveness in cross-modal scenario. We follow this up with a solution named Customized Cross-modal Knowledge Distillation (C$^2$KD) in Chapter \ref{c4.3}. Extensive experiments of audio-visual, image-text, and RGB-depth modalities in terms of classification and segmentation tasks are performed in Chapter \ref{c4.4}, Experimental results of C$^2$KD significantly outperform existing KD methods, demonstrating robustness in transferring knowledge even from low- to high-accuracy modalities while mitigating training instability and performance degradation caused by modality gaps. Ablation and sensitivity analysis as well as discussion are in Chapter \ref{c4.5} and \ref{discussion}, respectively.

\textbf{Chapter \ref{Chapter5}} (SID: Self-Introspective Decoding) addresses hallucination robustness in LVLMs through token-level adaptive pruning to amplify the fine-grained hallucinations then contrastively to alleviate the hallucinations. Firstly, Chapter \ref{c5.1} shows the detailed challenges of hallucination issues of LVLMs and analysis of previous contrastive decoding strategies. We then show the paradigm of LVLMs generation and comprehensively analyze contrastive decoding in LVLMs in Chapter \ref{Problem}. In Chapter \ref{c5_sec4}, we propose the Self-Introspective Decoding (SID) strategy to dynamically suppress the priors of LLM that conflict visual evidence, achieving a significant reduction in hallucinations.
Chapter \ref{sec5} illustrates the detailed experimental results on CHAIR, POPE, MME, MMBench, GPT-4 assisted benchmark. and GPT4-V assisted evaluation to validate that SID effectively alleviate the hallucination issues while preserving general abilities of LVLMs.

\textbf{Chapter \ref{Chapter6}} (Conclusion and Future Work) synthesizes the thesis contributions in Chapter \ref{c6.1} and proposes future research directions in Chapter \ref{c6.2}, including multimodal test-time adaptation (i.e., Chapter \ref{c6.2.1}), task-aware adaptation of multimodal LLMs (i.e., Chapter \ref{c6.2.2}), and developing multimodal agent as experts (i.e., Chapter \ref{c6.2.3}).
 
\chapter{Background}

\label{Chapter2}

Following the research framework we presented in Figure \ref{c1_framework} of Chapter \ref{Chapter1}, we will discuss previous and contemporary methodologies for building robust multimodal learning systems, including background of composition generalization robustness in Section \ref{sec:preliminary_czsl}, unimodal and crossmodal knowledge distillation background in Section \ref{sec:preliminary_cmkd}, and hallucination-related background in Section \ref{sec:preliminary_hallucination}.

\section{Composition Generalization Robustness}
\label{sec:preliminary_czsl}

\subsection{Problem Formulation}
Compositional Zero-Shot Learning (CZSL) aims to recognize the composition of two primitives, i.e., an state ($e.g., tiny$) and an object (e.g., $dog$). Given $S$ and $O$ as two sets of states and objects, spanning \emph{all} classes, we compose a set of possible state-object pairs, i.e., 
$C = S \times O = \left\{ {(s,o)\left| {s \in S,o \in O} \right.} \right\}$. Formally, given a training set 
${D^{s}} = \{ (i,c)|i \in {I^{s}},c \in {C^{s}}\} $, where $I^{s}$ is an training image set, and $C^{s}$ is the corresponding state-object labels. The close world CZSL follows the generalized ZSL~\cite{zsl1} that the test sample comes from either seen ($C^s$) or unseen ($C^u$) composition ($C^s \cup C^u $). For the \textbf{Open-World CZSL (OW-CZSL) setting} \cite{open_cvpr}, there assumes no prior on the set of testing compositions. It means the model must consider the full compositional space ($C$), which is much larger than $C^s \cup C^u$. Consequently, the unseen compositions are $C^u_{ow}=C \backslash C^s$. OW-CZSL introduces a more practical setting while bringing more challenging problems: 1) It is hard to generalize from small seen compositions to large unseen compositions. 2) There are a large number of less feasible compositions in the full composition space ($C$), confusing the prediction models. \cite{kgsp} recently proposes a new practical setting, i.e., only training with one of the state and object annotations, named \textbf{partial-supervision CZSL (pCZSL)}. Formally, for the training set $C^s$, The relation of the partial label of state and object primitives can be formulated as: $\{ (s,u)\}  \cup \{ (u,o)\}  = {C^s}$, where $u$ indicates unlabeled primitives. Consequently, the test set in pCZSL has the full output composition space ($C$) like OW-CZSL, while the training set in pCZSL does not have the composition knowledge about any state-object pairs.

\subsection{Composition Zero Shot Learning}

Different from typical zero-shot learning \cite{zsl1, zsl2,zsl3}, which aims to utilize attributed vectors or inherent semantic descriptions to recognize unseen instances,
\textbf{Compositional Zero-shot Learning} (\textbf{CZSL}) aims to recognize the state and object primitive (or modality) from the images, and even the state-object compositions are not ever seen in the training datasets. 
Unlike humans, which can extrapolate new concepts from previously learned knowledge, For instance, if the people are taught what the fried chip and toasted bread are, most of them can recognize the fried bread immediately, neural networks lack the compositional generalization ability. 
The main challenge of CZSL is modeling the relation and affordance of states and objects modalities, generalizing this capability to unseen compositions. Existing methods mainly deal with CZSL in two ways. The first way is inspired by Biederman’s Recognition-ByComponents theory~\cite{bio2} and Hoffman’s part theory~\cite{bio1}. For instance, Misra et al.~\cite{le} learn a transformation between individual classifiers of states and objects. Other representative methods learn hierarchical decomposition and composition of the state and object primitives~\cite{luc, laad, lsd}, model objects to be symmetric under attribute transformations \cite{sym}, and learn independent prototypical representations of visual primitives then propagated prototype via a compositional graph~\cite{prototype}. The second way tries to learn the joint representation of the state-object compositions from given images. Specially, SymNet~\cite{sym} enforces symmetries in the representation of objects given their state transformations. Graph network is also employed in~\cite{ge} to enforce the compositional information transfer from seen to unseen compositions. AoP~\cite{ao} regards attribute as the operator and models each state as a linear transformation of objects. CANet~\cite{canet} learns conditional attributes to enhance embedding space.  LAP~\cite{lap} exploits the self-attention mechanism to embed related compositions closer and unrelated far away. Differently, causality-based methods~\cite{casual, dec} explore decomposable objects and state representations.

Above methods perform well on the close-world CZSL, while suffering from severe degradation for the open-world setting \cite{open_cvpr, open_pami, kgsp}, where the output space has not imposed any limit. Mancini et al. \cite{open_cvpr} compute feasibility scores (i.e., cosine similarity) between visual features and compositional embeddings to reduce the output space. Then they further inject the feasibility scores both at the loss level and within the graph connections \cite{open_pami}. Karthik et al.~\cite{kgsp} follows the Visual Product \cite{le} and predicts state and object primitives independently with non-linear feature extractors. To refine the relation between independent primitives, Conceptnet \cite{conceptnet} is introduced as the external knowledge. We revisit the Visual Product and achieve cross-primitive compatibility in an easy-hard learning manner, avoiding the external knowledge in~\cite{kgsp} and cumbersome word embeddings in \cite{open_cvpr, open_pami}.

\section{Knowledge Distillation over Cross Modality}
\label{sec:preliminary_cmkd}
\subsection{Unimodal Knowledge Distillation}
Unimodal Knowledge Distillation (KD) transfers the knowledge of a pretrained teacher to a student by minimizing the discrepancies between output logits or intermediate features between student and teacher. Previous KD methods primarily concentrate on inheriting knowledge from the large-capacity teacher. Pioneering work \cite{kl} regularizes Kullback–Leibler (KL) divergence between student and teacher soft labels. CRD \cite{crd} develops contrastive-based objectives for knowledge transferring. SCKD \cite{sckd} automatically adjusts the KD process according to the distillation gradient similarity. Yang et al. \cite{srrl} utilize the teacher’s pre-trained classifier to regularize the student’s penultimate layer feature. Zhu et al. \cite{tllm} identify and discard the undistillable classes from the large teacher model based on the validation set. DKD \cite{dkd} decouple KD into target class and non-target class knowledge distillation to balance learning effectiveness and flexibility. Review \cite{review} proposes the review mechanism to utilize knowledge of teacher's multi-level features. 
% Considering student's noisy features, DiffKD \cite{diffusion} explicitly purifies and matches features using diffusion models. 
RKD \cite{rkd} regularizes the student with distance-wise and angle-wise structural relations to replace KL loss. DIST \cite{dist} further proposes a novel correlation-based loss to capture the inter-class and intra-class relations. L2D \cite{mlkd} extends relation-based distillation into multi-label classification. These KD methods focus on unimodal KD and learn to inherit knowledge from a fixed teacher. However, for CMKD, the modality gap impedes knowledge transfers across modalities. We \textit{argue} that teacher modality should be optimized with feedback supervision of student modality to produce \textit{receptive knowledge}. Previous online knowledge distillation methods \cite{dml, one, afd, shake} update teacher model to adapt student in unimodal scenarios. Specifically, DML \cite{dml} simply applies KD losses mutually, treating each other as teachers. ONE \cite{one} further exploits gated ensemble logits of multiple training networks. AFD \cite{afd} proposes online feature alignments via adversarial training. The recently proposed SHAKE \cite{shake} bridges offline and online KD by transferring knowledge through extra shadow heads.

\subsection{Cross-modal Knowledge Distillation}

\begin{figure*}[t]
\centering
\includegraphics[width=0.7\textwidth]{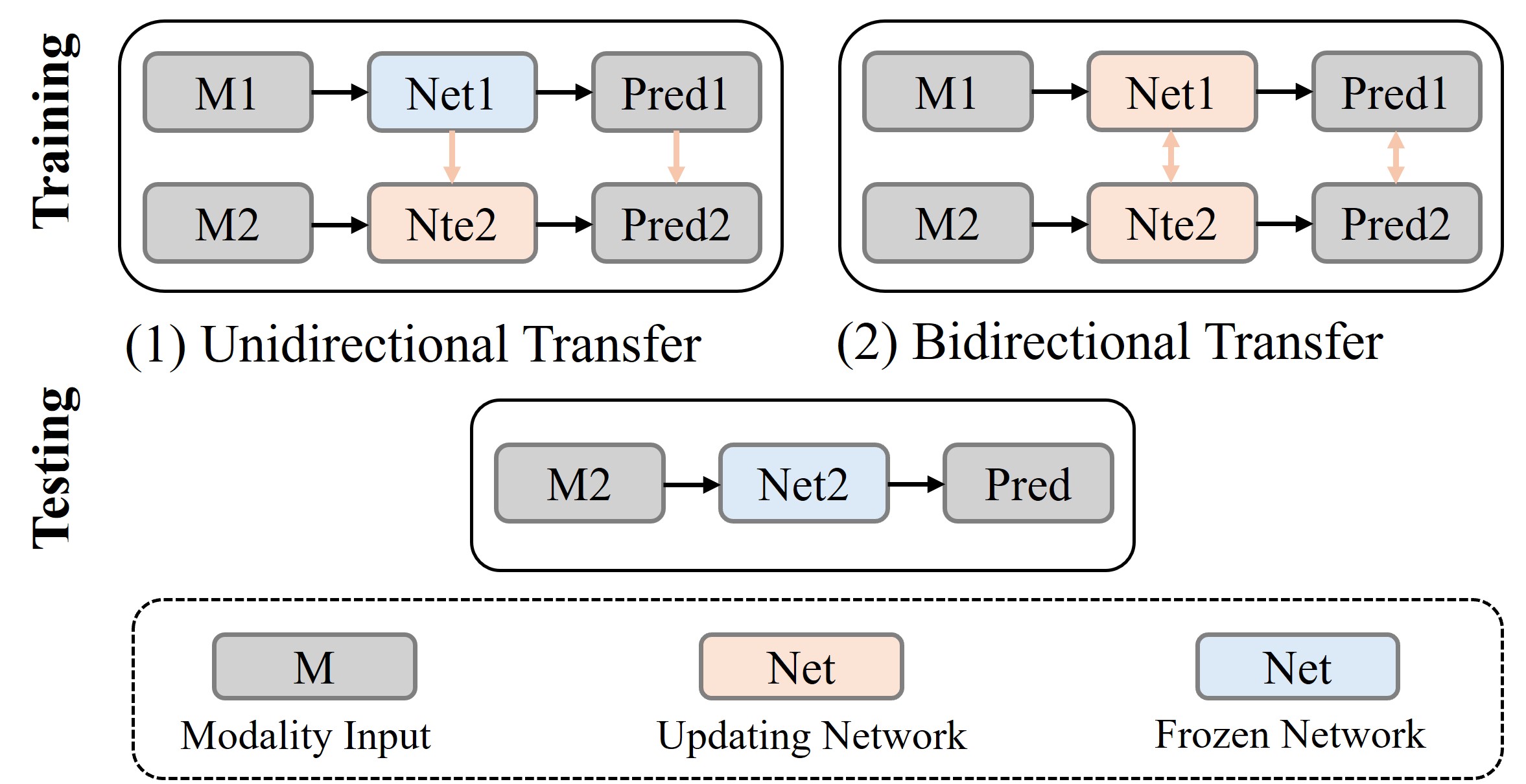}
\caption{\textbf{Intuitive presentation} of cross-modal knowledge distillation.}
\label{c2_cmkd}
\end{figure*}

With the rising prevalence of machine learning~\cite{aaai1,aaai2,iccvw,ieeesj,Liu_2023_CVPR,reqa,tmm,Guo_Guo_Zhou_Liu_Lu_Huo_2023,zhu2022spatiotemporal,wei2025responsiblediffusioncomprehensivesurvey} and mulimodal sensors~\cite{tcsvt,tim,kbs,nn,eccv,yuan2025echobenchbenchmarkingsycophancymedical,guo2025evolutionfederatedposttraininglarge,yakun2025perceptionunderstandingreasoningmultimodal}, traditional KD methods have been extended to achieve knowledge transfer across multimodal data, thereby enhancing downstream tasks \cite{survey1, unsup_crosskd, gan_crosskd, expand_crosskd, rgbt, rgb_tad, indoor_crosskd, Huo_2024_CVPR}.
Figure \ref{c2_cmkd} illustrates the protocols for cross-modal knowledge distillation, where multiple modalities are utilized and cross-modal knowledge is transferred during the training phase. In the inference phase, only one modality is available, but it benefits from the inherited multimodal knowledge acquired during CMKD, which is critical for the modality missing situations.
\textit{However}, previous methods typically utilize \textit{high-accuracy} or \textit{well-labeled} modality as the teacher to transfer knowledge to low-accuracy or unlabeled modality~\cite{survey1}. 
For example, \cite{unsup_crosskd} leverage a large labeled modality as the supervisory signal for a new unlabeled paired modality. \cite{gan_crosskd} transfers knowledge among the missing and available modalities via GANs. \cite{expand_crosskd} adapts a multimodal network to the unlabeled modality by inheriting knowledge from the well-trained unimodal teacher. \cite{rgb_tad} proposes a decomposed cross-modal distillation method to enhance RGB-based detector by transferring knowledge of the optical flow modality. \cite{indoor_crosskd} distills ImageNet pre-trained visual modality to audio modality for indoor dense prediction.
Recently, Xue et al. \cite{mfh} first perform an in-depth investigation on CMKD and propose the modality focusing hypothesis (MFH), suggesting that modality-general decisive features are crucial determinants of CMKD efficacy. \cite{mfh} contributes to MFH but doesn't develop unified solutions. In this chapter, we further quantitatively analyze the challenges of CMKD (the modality gap, i.e., \textit{modality imbalance} and \textit{soft label misalignment}) and propose effective solutions to address these issues.

\section{Multimodal Hallucination}
\label{sec:preliminary_hallucination}

\subsection{Multimodal Large Language Models}

Motivated by the success of Large Language Models (LLMs) \citep{llama, qwen, vicuna, stanford, llama2, llama3}, recent studies have extended LLMs to multimodal regions and provided Large Vision-Language Models (LVLMs) \citep{llava, minigpt, shikra, mplug, mimic, qwenvl, blip2, instructblip, llava1.5, fuyu, internvl, llavanext} powered by pre-trained LLMs. LVLMs understand and generate diverse content in a more comprehensive way by integrating user instruction and vision inputs. LLaVA \citep{llava} connects open-set vision encoder with LLMs (i.e., Vicuna \citep{vicuna}) by instruction tuning with elaborated language-image instruction-following data. Then, LLaVA-1.5 \citep{llava1.5} develops the vision-language connector that is data-efficient and powerful for better multimodal understanding. Shikra \citep{shikra} further incorporates grounding data and trains the model to understand the grounding knowledge in the given images. BLIP-2, InstructBLIP, and MiniGPT-4 \citep{blip2, instructblip, minigpt} introduce a learnable querying transformer to fusion multimodal features and largely reduce image tokens. Fuyu \citep{fuyu} proposes a vanilla decoder-only architecture without the vision encoder and adapter that makes it easier to understand, scale, and deploy. InternVL \citep{internvl} proposes three simple but effective improvements, including a strong vision encoder, dynamic high-resolution, and high-quality bilingual dataset. Recently, built on SOTA open-source LLaMA 3 \citep{llama3} and increasing the input vision resolution to 4$\times$ more pixels, LLaVA-NeXT  \citep{llavanext} exhibits excellent multimodal capabilities. Despite the impressive results, all of the above LVLMs suffer from serious hallucination problems, and we mainly conduct experiments on advanced LVLMs, including InstructBLIP, Shikra, LLaVA-1.5, and LLaVA-NeXT.

\subsection{Decoding Strategy in LLMs}
Selecting decoding strategies in language models is crucial, as it determines how models generate text. 
Greedy decoding selects the highest probability next token at each step but might lead to less varied text. 
Beam search \citep{beam} is an accumulated-score-based decoding strategy. It maintains a set of beams to enlarge the candidate range and finally selects the best one in beams, which is more sophisticated than greedy decoding. 
Sampling decoding generates the next words by randomly selecting from the output distribution, where Top-k sampling \citep{top-k} samples from Top-k likely tokens \citep{top-k} and brings diversity but sometimes induces less coherent outputs.
Top-p (Nucleus) sampling \citep{top-p} improves Top-k sampling that considers the dynamic number of words that reach the probability p, achieving a balance between randomness and relevance. 
Recently, to alleviate the hallucination issue, DoLa \citep{dola} decoding emphasizes the knowledge of mature layers and downplays that of pre-mature layers. 
OPERA \citep{opera} is established on beam-search decoding strategy and finds the interesting phenomenon of high-probability co-occurrence between the hallucination and the knowledge aggregation patterns. OPERA penalizes `Over-Trust Logit' in the beam score to alleviate aggregation patterns. 
In this thesis, we aim to contribute the decoding strategy that can be seamlessly integrated into different decoding strategies to mitigate multimodal hallucinations without sacrificing text generation quality, such as diversity, coherence, and relevance.

\subsection{Hallucination in Foundation Models}
Hallucination, defined as the generation of irrelevant, factually incorrect, or meaningless text in a given context \citep{chair, snowball, hallusionbench, ear, huo2025selfintrospectivedecodingalleviatinghallucinations}, is a significant bottleneck in current foundation models. 
This issue can stem from overfitting specific patterns in the training data, a lack of understanding world knowledge, or an inability to effectively contextualize a given input \citep{hal_survey}. In the context of LLMs, hallucinations often manifest as generated content that conflicts with world knowledge or common sense. For LVLMs, the primary concern is whether the generated answer conflicts with the provided images. To mitigate the hallucination issue, 
several solutions have been proposed, including \textbf{robust instruction tuning with curated datasets} \citep{factuality, dph, rit, hadpo, hacl, rlhfv, doctor, vista, eos}, \textbf{post-hoc utilizing auxiliary analysis networks} \citep{selfcheckgpt, lure,  pecker, halc, logical, collaboration}, and \textbf{various decoding strategies} \citep{contrastive, dola, pai, vcd, vig, icd, id, ibd}. However, robust instruction tuning requires massive high-quality datasets and advanced GPU clusters, making it resource-intensive; Post-hoc utilizing auxiliary networks heavily rely on the auxiliary network, leading to high inference costs. 
As for decoding strategies, representative LVLMs hallucination alleviation methods \citep{vcd, vig, icd} manually disturb raw inputs to induce hallucinations then contrast them to alleviate the issue. However, holistic disturbing raw inputs might bring additional noise during contrastive decoding, and double the inference cost. 
In this thesis, we propose an efficient Self-Introspective Decoding (SID) that induces and then mitigates vision-and-text association hallucination by token-level disturbances, greatly reducing the inference cost. 

\chapter{\textsc{ProCC}: Progressive Cross-Primitive Compatibility for Composition Generalization}
\label{Chapter3}

\section{Challenges and Motivations}
\label{c3.1}
Current neural networks lack the compositional generalization robustness inherent to human cognition. Specifically, the training set contains images with corresponding multimodal descriptions (primitives), namely states and objects. Since objects and states are semantically entangled, that is, objects in different states often have different appearances, and states also vary greatly depending on the object, the model needs to be based on known primitives. Identify unseen combinations. The main challenge behind CZSL is how to model interactions between state and object primitives and extrapolate seen combinations to unseen combinations.
Existing methods mainly focus on learning a shared embedding space of object-state combinations \cite{sym, ge, ao, lap} or compositional attribute and object classifiers \cite{tmn, le, sce, rac, dec}.

However, the performances of these methods degrade to some extent \cite{open_cvpr, open_pami} as for the open-world setting (OW-CZSL), where there are no priors on the unseen compositions, and the model must consider the whole possible compositions in terms of all objects and states. To deal with such a problem, existing mainstream methods utilize feasibility constraints on the composition embedding \cite{open_cvpr, open_pami} or independently predict simple state and object primitives \cite{rvp, kgsp}. While \cite{open_cvpr, open_pami} rely on different word embedding methods. The straightforward but effective Visual Product method like \cite{kgsp} predicts the state and object primitives while ignoring the compatibility between two primitives. So external knowledge is introduced to eliminate less feasible compositions, while it is cumbersome to select proper external knowledge for varying datasets.

\begin{figure*}[t]
\centering
\includegraphics[width=1.0\textwidth]{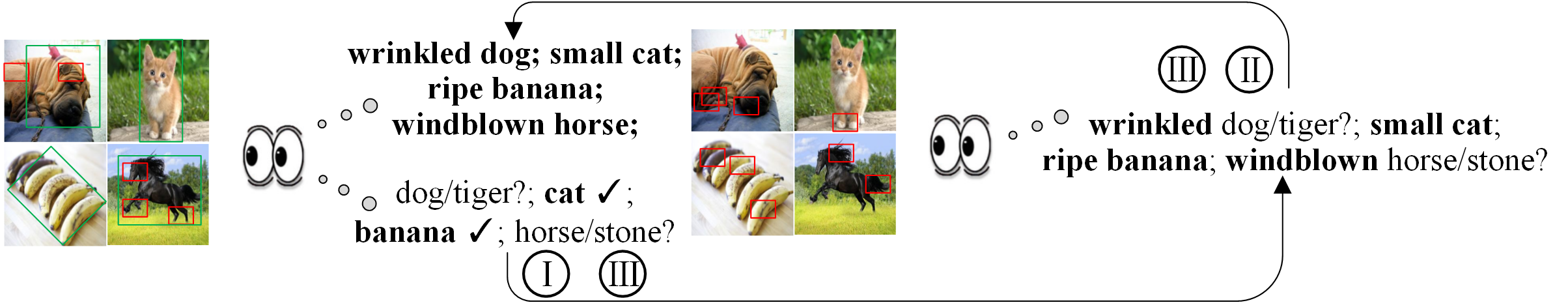}
\caption{\textbf{The overall concept of our method.} Following the principle of 'forest before trees' \cite{neuro}, human feedforward hierarchy underlies implicit processing for initial vision at a glance (i.e., green rectangle), and feedback connections add details to
explicit vision with scrutiny (i.e., red rectangle). As for composition generalization learning, humans first ($\textcircled{\scriptsize{\uppercase\expandafter{\romannumeral1}}}$) learn to recognize overall objects, then ($\textcircled{\scriptsize{\uppercase\expandafter{\romannumeral2}}}$) gradually identify the scrutiny attribute of objects, i.e., state, and finally ($\textcircled{\scriptsize{\uppercase\expandafter{\romannumeral3}}}$) reasonably compose the object and state primitives. Inspired by this, we aim to progressively recognize the object and state primitives and guide the network to exploit discriminative information conditioned on learned knowledge via the CPC module.}
\label{fig1_real11}
\end{figure*}

To address the aforementioned problems, we propose Progressive Cross-primitive Compatibility (ProCC) network to recognize compositions in the open-world setting and a more realistic setting (i.e., partial supervision), aiming at attaining cross-primitive compatibility during easy-hard recognition progress, as shown in Figure \ref{fig1_real11}. Specifically, following the route of the human learning process \cite{neuro}, we \textbf{first} learn to classify objects, which is easier than recognizing states \cite{dve, kgsp} because the same state varies greatly conditioned on objects and related contexts, i.e., ancient castle / ancient coin, and different states are sometimes less feasible composed with the same object, i.e., old dog / ripe dog. \textbf{Then}, with the learned knowledge of object primitive, we sequentially classify state primitives conditioned on object features via Cross-Primitive Compatibility (CPC) module, excavating discriminative information. \textbf{Finally}, we finetune the whole network conditioned on prior knowledge of two primitives. The ProCC achieves cross-primitive compatibility by adjusting the visual attention to filter out less feasible compositions, without the aid of external knowledge like Word2vec \cite{word2vec}, Glove \cite{glove}, Conceptnet \cite{conceptnet} etc. Also, the progressive training paradigm effectively models the interactions of primitives via conditioned features, especially for pCZSL, where only partial label results in invalid interactions.
% and alleviates the imbalance (over/under-fitting) learning of state and object classifiers compared with the jointly training. 

In summary, this chapter's contributions are four-fold:

1) We propose a novel Progressive Cross-primitive Compatibility (ProCC) network, mimicking the human learning progress of recognizing the state and object compositions without external knowledge. 

2) We revisit Visual Product methods and present a Cross-Primitive Compatibility (CPC) module to model the interactions of classifiers to exploit the discriminative visual attention conditioned on each other, guiding the model to generalize to feasible compositions. 

3) The progressive training paradigm alleviates the invalid cross-primitive interactions without the aid of cumbersome external knowledge, especially for pCZSL.

4) Comprehensive experimental results on three large-scale datasets for OW-CZSL and pCZSL tasks demonstrate the effectiveness of our proposed approach, which outperforms the state-of-the-art methods.

\begin{figure*}[t]
\centering
\includegraphics[width=1.0\textwidth]{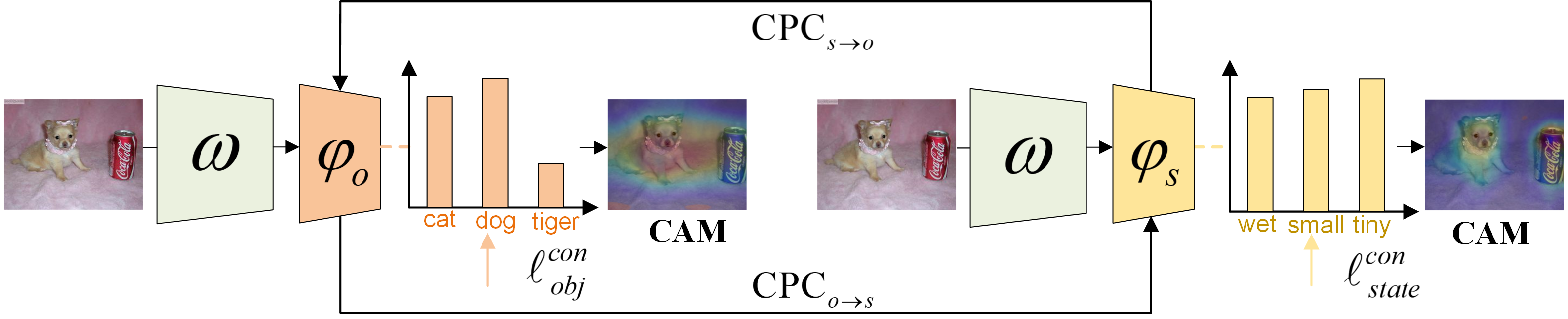}
\caption{\textbf{The framework of ProCC.} Features from the encoder ($\omega$) are respectively fed to the object and state ($\varphi_o$ and $\varphi_s$) classifiers, where the Cross-Primitive Compatibility (CPC) aims to model the cross-primitive interactions. Progressive learning strategy is proposed to gradually modulate primitive compatibility, especially for pCZSL. For detailed training procedure, please refers to \textbf{Algorithm 2}. Class Activation Maps (CAM) of input samples are illustrated to show visual attention.}
\label{fig22}
\end{figure*}

\section{Progressive Cross-primitive Compatibility (ProCC)}
\label{sec:pistis_model}

Most CZSL methods \cite{casual, sym, ao, tmn, dve, open_cvpr, open_pami, ge} explicitly modulate the interactions of states and objects to improve the generalization ability. However, it is less effective for OW-CZSL and pCZSL due to large output space and missing labels. Some methods \cite{rvp, kgsp} follow the Visual Product \cite{le} that independently predict the state and object primitives, disregarding compositional nature. Following the route of \cite{rvp, kgsp, le}, we propose Progressive Cross-primitive Compatibility (ProCC) network while achieving cross-primitive compatibility. Also, like the human learning process \cite{neuro}, ProCC trains the network in an easy-hard manner, which dynamically models interactions between state and object primitives, alleviating the negative influence of no explicit supervision on both states and objects in pCZSL. Figure \ref{fig22} shows the framework of the proposed approach.
In the following subsections, we revisit the Visual Product and introduce a cross-primitive compatibility module and progressive learning strategy.

\subsection{Revisit Visual Product}
Generally, given an image $i$, CZSL wants to model the joint probability distribution $p(s_i,o_i|i)$. The visual product simplifies this as follows:
\begin{eqnarray}
p(s_i,o_i|i) \approx p(s_i|i) \times p(o_i|i)
\end{eqnarray}
In this way, Visual Product treats the states and objects independently only from the visual cues, without side information (i.e., word embeddings). Concretely, input image $i$ is firstly encoded to obtain the feature $z$ as: $z=\omega (i)$.
Then the object 
(i.e., ${\varphi _o}\left \langle z, o\right \rangle $)
and state (i.e., ${\varphi _s}\left \langle z, s\right \rangle $)
classifiers assign $z$ to the vectors in the probability simplex $o$ and $s$, spanning all object and state classes. Visual Product minimizes the cross-entropy loss of seen compositions ($D^{s}=\{I^s, C^s\}$) for both object and state predictions:
\begin{eqnarray}
{\ell _{vp}} = {\ell _{obj}(i, o_i)} + {\ell _{state}(i, s_i)}
\label{eq_all}
\\
{\ell _{obj}} =  \mathop {\min }\limits_{\varphi _o} \sum\limits {{\ell _{ce}}({\varphi _o}\left \langle\omega (i), o\right \rangle,o_i)} 
\label{eq_o}
\\
{\ell _{state}} =  \mathop {\min }\limits_{\varphi _s} \sum\limits {{\ell _{ce}}({\varphi _s}\left \langle\omega (i), s\right \rangle,s_i)} 
\label{eq_s}
\end{eqnarray}
where $(i,(s_i, o_i)) \in {D^s}$. Thus, the prediction function is:
\begin{eqnarray}
{f }(i) = \arg \mathop {\max }\limits_{(s,o) \in C} {\varphi _s}\left \langle\omega (i), s\right \rangle\times{\varphi _o}\left \langle\omega (i), o\right \rangle
\end{eqnarray}
where $C$ represents the full state-object composition pairs in OW-CZSL. As the search space is huge, Visual Product is more effective than previous methods, which aim to produce discriminative state-object embeddings \cite{rvp, kgsp}. Recently, \cite{rvp, kgsp} expanded the visual product and equipped the classifiers with multi-layer perceptrons (MLP) to excavate discriminative features. Also, external knowledge~\cite{conceptnet} is employed in \cite{kgsp} to estimate the feasibility scores of compositions. Here, we explicitly model the composition interactions via Cross-Primitive Compatibility (CPC) module during the training procedure, without external knowledge. 
Also, considering the pCZSL setting and better modulating the primitive compatibility, the progressive learning strategy, following the human learning process \cite{neuro}, is proposed to facilitate cross-primitive compatibility in an easy-hard manner. 

\begin{figure}[t]
\centering
\includegraphics[width=0.8\columnwidth]{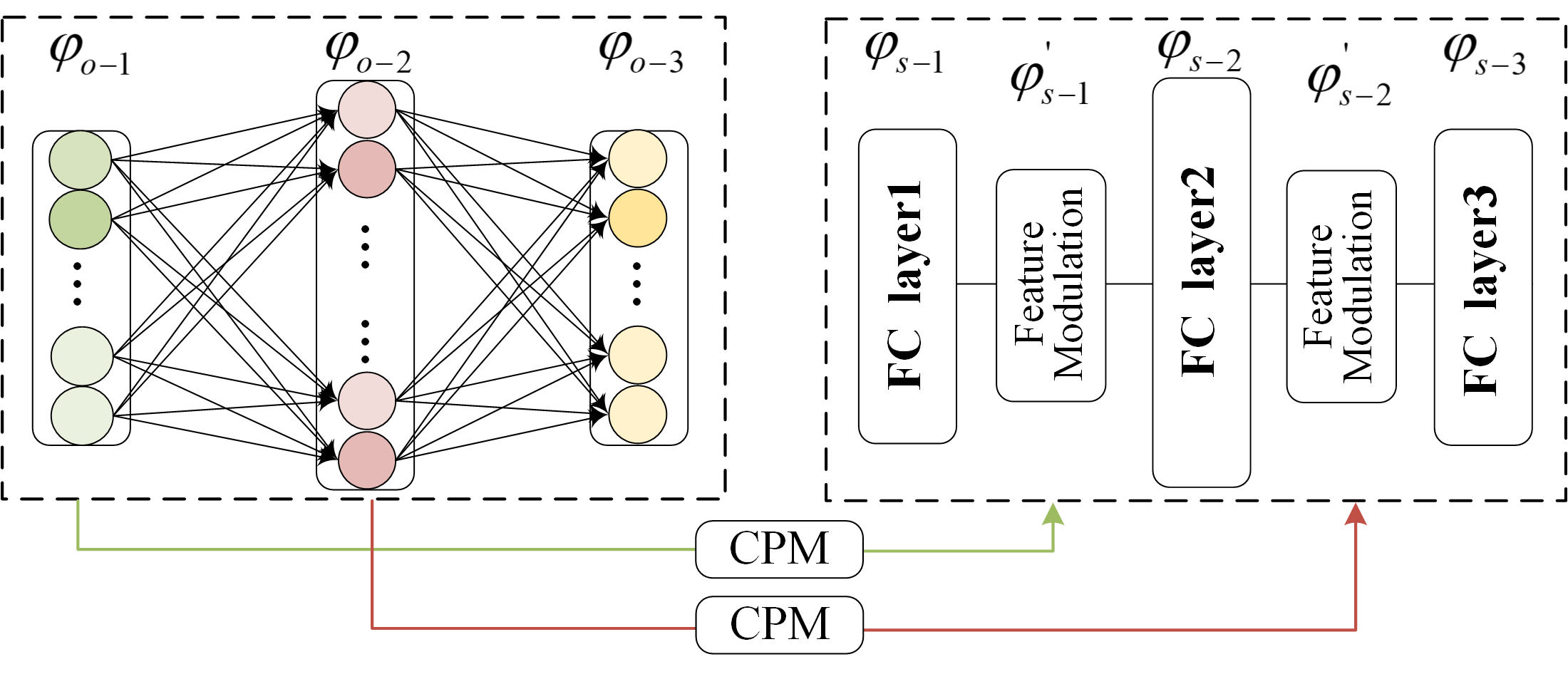}
\caption{\textbf{The detailed framework of the object-state Cross-Primitive Compatibility (CPC$_{o \rightarrow s}$).} Features from the object classifier ($\varphi_{o-1}$ and $\varphi_{o-2}$) are encoded by learnable Cross-Primitive Memory (CPM) units. Then respectively interact with state features ($\varphi_{s-1}$ and $\varphi_{s-2}$) to achieve compatibility of state features conditioned on objects.}
\label{fig33}
\end{figure}

\subsection{Cross-primitive Compatibility Module}
Visual Product methods independently predict compositions via Equation 1, which ignores the fact that the feasibility of state-object compositions is heavily conditioned on each other. A more practical compositional probability can be modeled as:
\begin{eqnarray}
p(s_i,o_i|i) \approx p(s_i|i,f{_o}(i))) \times p(o_i|i,f{_s}(i)))
\end{eqnarray}
where $f{_o}(i)$ and $f{_s}(i)$ are intermediate features of the object and state primitives.  
It is non-trivial to directly model the relationship between objects and states due to the diverse semantic entanglement and a large number of possible compositions. We integrate the feasibility reasoning into the trainable Cross-Primitive Compatibility (CPC) module, which 
facilitates interactions between two classifiers to explore informative visual attention conditioned on feature representations of each primitive. 
Specifically, the features extracted by the encoder ($\omega$) are fed to primitive classifiers (i.e., $\varphi_o$ and $\varphi_s$). The primitive classifiers follow the Visual Product methods \cite{rvp, kgsp} that consist of multi-layer perceptron (MLP), specifically three-layer MLP, for classifications. As shown in Figure \ref{fig22} and Equation 6, the network is symmetric and we take the object-state CPC ($\rm{CPC}_{o \rightarrow s}$) module for example, as shown in Figure \ref{fig33}, intermediate features (i.e., output distributions) from $\varphi_{o-1}$ and $\varphi_{o-2}$ are fed to $\varphi_s$ to interact with state features. However, direct modulation state features will induce information degradation because of the huge task diversity. 
We propose learnable Cross-Primitive Memory (CPM) units for soft interactions. Specifically, 
the learnable CPM unit introduces conditioned information to modulate corresponding features along with the residual connection, which is formulated as follows:
\begin{eqnarray}
\varphi _{o - l}^m = \sigma \left( {{\rm{Conv}}_{\rm{{1d}}}^k\left( {{\varphi _{o - l}}} \right)} \right),{\rm{ }}l \in \left( {1,2} \right)
\\
\varphi _{s - l}^, = {\varphi _{s - l}} \times \varphi _{o - l}^m + {\varphi _{s - l}},{\rm{ }}l \in \left( {1,2} \right)
\end{eqnarray}
where ${{\rm{Conv}}_{\rm{1d}}^k}$ and $\sigma$ represent the $\rm{1d}$ convolution layer and softmax activation function. Kernel size ($k$) is equal to 1/10 feature dimension to efficiently capture the long-range dependency. Then the enhanced state features are fed to the next layer of $\varphi_s$ as:
\begin{eqnarray}
{\varphi _{s - (l + 1)}} = {f_{s - l}}(W_{s - l}^T\varphi _{s - l}^{,} + {b_{s - l}}),{\rm{ }}l \in (1,2),
\end{eqnarray}
where $W$ and $b$ are weights and biases of MLP. Accordingly, the conditioned cross-primitive interactions are injected into each other, reducing less feasible primitive predictions. Therefore, Equations 3 and 4 can be re-write as:
\begin{small}
\begin{eqnarray}
{\ell _{obj}^{con}} =  \mathop {\min }\limits_{\varphi _o, \varphi _{o \to s}} \sum {{\ell _{ce}}({\varphi _o}\left \langle z|\varphi _{s \to o}({\varphi _s (z)}), o\right \rangle,o_i)} 
\label{eq_o_}
\\
{\ell _{state}^{con}} =  \mathop {\min }\limits_{\varphi _s, \varphi _{s \to o}} \sum {{\ell _{ce}}({\varphi _s}\left \langle z|\varphi _{o \to s}({\varphi _o (z)}), s\right \rangle,s_i)} 
\label{eq_s_}
\end{eqnarray}
\end{small}
where $z=\omega (i)$, $(i,(s_i,o_i)) \in {D^s}$, and ${\ell _{vp}^{con}} = {\ell _{obj}^{con}} + {\ell _{state}^{con}}$

\begin{figure}[t]
\centering
\includegraphics[width=0.85\columnwidth]{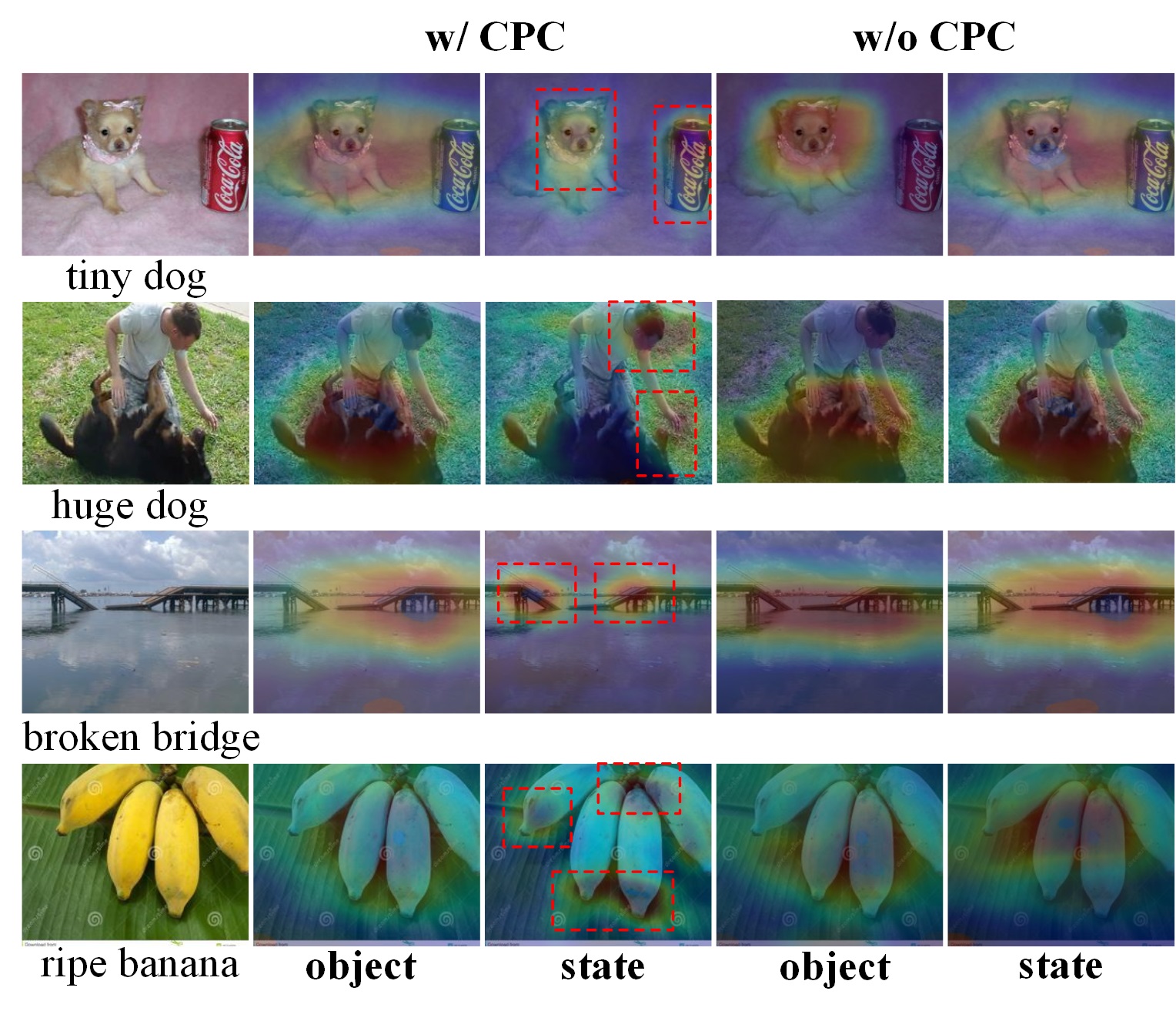}
\caption{\textbf{Visualizations of class activation maps} of ProCC with and without CPC modules on the testing dataset of MIT-States. The discriminative regions are marked with red rectangles.}
\label{fig44}
\end{figure}

\subsection{Visual Explanation} 
\begin{figure}[t]
\centering
\includegraphics[width=0.9\columnwidth]{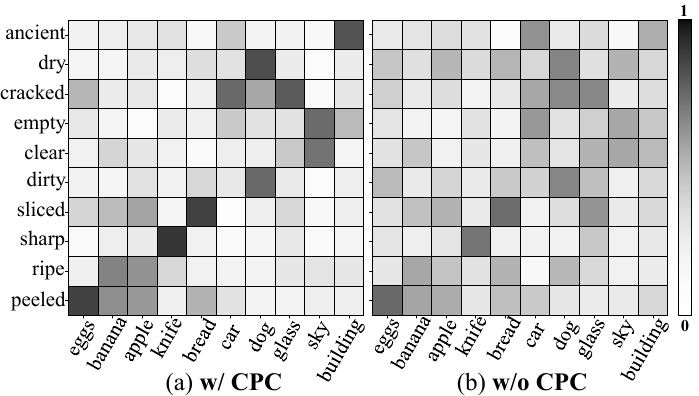}
% \vspace{-0.3cm}
\caption{\textbf{Confusion matrices} about prediction probabilities of states conditioned on objects (w/ CPC) or not (w/o CPC).}
\label{fig55}
% \vspace{-0.5cm}
\end{figure}

To further illustrate and explain the effect of the CPC module, we visualize the attention learned from the classifier via Class Activation Map (CAM) \cite{cam} in Figure \ref{fig44}. The standard CAM is formulated as:
\begin{eqnarray}
{\rm{CA}}{{\rm{M}}_c}(x,y) = {\sum_{k}} {\omega _k^c{f_k}(x,y)}  
\end{eqnarray}
where $\rm{CAM}_c$ means the class activation map that leads to the classification of an image to class $c$. ${f_k}(x,y)$ and $\omega_k^c$ stand for the activation of unit $k$ in the last layer at spatial location $(x, y)$ and the weight corresponding to class c for unit $k$. Here, $\omega_k^c$ is the final layer of the MLP (i.e., ${\varphi _{o - 3}}$ and ${\varphi _{s - 3}}$), which has been modulated by the CPC modules. 
Figure \ref{fig44} shows some visualization examples with (w/) and without (w/o) CPC module. As the encoder ($\omega$) is pre-trained for the object classification task, most CAMs for the object classifier can locate and recognize the proper attention regions. However, the CAMs for the state classifier vary greatly as state primitives are conditioned on the object primitive and related contexts. For the $tiny$ $dog$ and $huge$ $dog$ compositions, the CPC module drives the model to focus on the discriminative regions that a dog with a small head compared with other objects tends to classify to the $tiny$ otherwise classify to $huge$. For more abstract compositions, $broken$ $bridge$ and $ripe$ $banana$ compositions, the state primitives heavily depend on the object primitives otherwise may induce less feasibility compositions. The state of $broken$ is mainly reflected in the curvatures of the bridge and the $ripe$ primitive of the banana displays the black spots on the surface. 
Overall, the CPC module enables the efficient adjustment of visual attention conditioned on mutual relations. 
Moreover, Figure \ref{fig55} illustrates the confusion matrices about
% prediction probabilities of 
state and object primitives. Concretely, we select ten typical state and object primitives in the MIT-States \cite{mit} dataset. Prediction probabilities of states are accumulated then normalized with and without CPC module to formulate the confusion matrices. We can learn that the CPC module facilitates reasoning compatible compositions with high confidence.

\begin{algorithm}[]  %其中这里面不能有H不然会报错，不过不影响结果
	\caption{Training procedure of ProCC.}%算法名字
	\LinesNumbered %要求显示行号
	\KwIn{Training data ${D^{s}} = \{ (i,c)|i \in {I^{s}},c \in {C^{s}}\}$, pre-trained $\omega$, learning rate $\lambda_1$, $\lambda_2$, $\lambda_3$}%输入参数
	\KwOut{Optimal $\varphi_o$, $\varphi_s$, CPC: $\varphi_{o \to s}$, $\varphi_{s \to o}$ }%输出
	\textbf{Initialize:} $\varphi_o$, $\varphi_s$, $\varphi_{o \to s}$, $\varphi_{s \to o}$; %\;用于换行

        \textbf{Stage 1:} // train $\varphi_o$

	\While{not converged}{
		Sample a batch from $D^s$ as images ${(i_k)}_{k=1}^n$ with their object labels ${(o_k)_{k=1}^{n}}$ \;
        
        \For{samples in the batch}{
            Compute $\ell_{obj}$ via Equation 4.3.\;
            Update ${\varphi _o} \leftarrow {\varphi _o} - \lambda_1 {\nabla _{{\varphi _o}}}{\ell_{obj}}$
        }  
	}

         \textbf{Stage 2:} // train $\varphi_s$ and $\varphi_{o \to s}$

	\While{not converged}{
		Sample a batch from $D^s$ as images ${(i_k)}_{k=1}^n$ with their state labels ${(s_k)_{k=1}^{n}}$ \;
        
        \For{samples in the batch}{
            Compute $\ell_{state}^{con}$ via Equation 4.11.\;
            Update ${\varphi _{s \cup o \to s}} \leftarrow {\varphi _{s \cup o \to s}} - \lambda_2 {\nabla _{{\varphi _{s \cup o \to s}}}}{\ell_{state}^{con}}$
        }  
	}

        \textbf{Stage 3:} // finetune $\varphi_o$, $\varphi_s$, $\varphi_{o \to s}$, and $\varphi_{s \to o}$

	\While{not converged}{
		Sample a batch from $D^s$ as images ${(i_k)}_{k=1}^n$ with their object and state labels ${(o_k, s_k)_{k=1}^{n}}$ \;
        
        \For{samples in the batch}{
            Compute $\ell_{vp}^{con}$ via Equations 4.10 and 4.11.\;
            Update ${\varphi _{total}} \leftarrow {\varphi _{total}} - \lambda_3 {\nabla _{{\varphi _{total}}}}{\ell_{vp}^{con}}$
        }  
	}
\end{algorithm}

\subsection{Progressive Learning Strategy}
However, jointly training the state and object classifiers may induce two issues:
\textbf{(1)} When it comes to the more practical setting, partial supervision Compositional Zero-Shot Learning (pCZSL), where only the partial label, not both, is available \cite{kgsp}. The missing label makes the joint training strategy invalid to model the interactions between the state and object primitives. A naive way of learning from such partial supervision is to update the parameters of the state and object classifier only based on the available labels, which lacks the interaction information across primitives via the CPC module. Recent method \cite{kgsp} estimates the missing labels via pseudo-labeling \cite{plabel} as well as utilizes the external knowledge \cite{conceptnet}. The challenge of missing labels also exists in the standard Multi-Task Learning (MTL) that the traditional updating rule will give inferior results due to the missing annotations \cite{mtl_2, mtl_u1, mtl_u2, mtl_c4}. Some typical solutions propose hard knowledge distillation \cite{mtl_u1}, alternative optimization strategy \cite{mtl_u2}, and learning in the joint pairwise task spaces \cite{mtl_c4}. However, compared with the MTL task, the missing label issue matters more to the CZSL task, as the object and state primitives are heavily tangled. 
\textbf{(2)} Also, jointly training results in sub-optimal interactions as the diverse difficulty of object and state predictions. 
Concretely, classifying states is more challenging than objects \cite{dve, kgsp}. Therefore, joint training inevitably induces noisy conditioned information, which hinders to reason cross-primitive compatibility.

To enable the full interaction of state and object primitives, we propose a progressive learning strategy, mimicking the easy-hard learning process shown in Figure \ref{fig1_real11}. 
Concretely, with the features from the encoder ($\omega$), we first train the object classifier $\varphi _o$ with given labels (Equation \ref{eq_o}), to obtain object features ($\varphi_{o-l}$, $x\in(1,2)$). Then we sequentially train the state classifier $\varphi_s$ and CPC$_{o \rightarrow s}$ ($\varphi _{o \rightarrow s}$) conditioned on pre-trained object features ($\varphi_{o-l}$) (Equation \ref{eq_s_}), to interact to adjust the visual attention. Finally, we fine-tune the state and object classifiers ($\varphi _s$ and $\varphi _o$) as well as CPC modules ($\varphi _{o \rightarrow s}$ and $\varphi _{s \rightarrow o}$) conditioned on the well-trained features (Equations \ref{eq_o_} and \ref{eq_s_}). We utilize this training protocol both in the OW-CZSL and pCZSL settings. During the easy-hard recognition progress, our method alleviates invalid interactions of cross primitives, especially in the pCZSL setting, without external knowledge. For detailed training procedure, please refers to \textbf{Algorithm 2}.

\begin{table*}[ht]\centering
\scriptsize
\renewcommand\arraystretch{1}
\setlength{\tabcolsep}{1.0mm}{
\begin{tabular}{l|cccccc|cccccc|cccccc}
\multirow{3}{*}{\textbf{Method}} & \multicolumn{6}{c|}{\textbf{C-GQA}}                                                         & \multicolumn{6}{c|}{\textbf{MIT-States}}                                                   & \multicolumn{6}{c}{\textbf{UT-Zappos}}                                                        \\ \cline{2-19} 
                                 & \multicolumn{2}{c|}{Val}      & \multicolumn{4}{c|}{Test}                                   & \multicolumn{2}{c|}{Val}     & \multicolumn{4}{c|}{Test}                                    & \multicolumn{2}{c|}{Val}       & \multicolumn{4}{c}{Test}                                      \\
                                 & HM            &\multicolumn{1}{c|}{AUC}          & $S$             & $U$            & HM           & AUC           & HM           & \multicolumn{1}{c|}{AUC}          & $S$             & $U$             & HM            & AUC          & HM            & \multicolumn{1}{c|}{AUC}           & $S$             & $U$             & HM            & AUC           \\ \hline
TMN                              & NA            & NA           & NA            & NA           & NA           & NA            & 2.1          & 0.2          & 12.6          & 0.9           & 1.2           & 0.1          & 21.2          & 9.2           & 55.9          & 18.1          & 21.7          & 8.4           \\
AoP                              & NA            & NA           & NA            & NA           & NA           & NA            & 3.2          & 0.3          & 16.6          & 5.7           & 4.7           & 0.7          & 23.4          & 10.1          & 50.9          & 34.2          & 29.4          & 13.7          \\
LE+                               & 9.3           & 1.8          & 19.2          & 0.7          & 1.0          & 0.08          & 5.3          & 0.5          & 14.2          & 2.5           & 2.7           & 0.3          & 26.6          & 14.3          & 60.4          & 36.5          & 30.5          & 16.3          \\
VisProd                          & 10.5          & 2.0          & 24.8          & 1.7          & 2.8          & 0.33          & 7.2          & 1.0          & 20.9          & 5.8           & 5.6           & 0.7          & 28.8          & 15.4          & 54.6          & 42.8          & 36.9          & 19.7          \\
SymNet                           & 12.3          & 2.5          & 26.7          & 2.2          & 3.3          & 0.43          & 8.0          & 1.2          & 21.4          & 7.0           & 5.8           & 0.8          & 32.5          & 16.7          & 53.3          & 44.6          & 34.5          & 18.5          \\
CGE                              & 12.8          & 2.8          & 28.3          & 1.3          & 2.2          & 0.30          & 8.3          & 1.8          & \textbf{29.6} & 4.0           & 4.9           & 0.7          & 34.5          & 18.9          & 58.8          & 46.5          & 38.0          & 21.5          \\
CompCos                          & 12.0          & 2.4          & 28.4          & 1.8          & 2.8          & 0.39          & {\ul {8.4}}    & 1.5          & 25.4          & 10.0           & {\ul { 8.9}}           & {\ul {1.6}}          & 32.5          & 18.1          & {\ul { 59.3}}    & 46.8          & 36.9          & 21.3          \\
Co-CGE                           & 12.3          & 2.7          & {\ul { 28.7}}    & 1.6          & 2.6          & 0.37          & {\ul { 8.4}}    & \textbf{2.1} & 26.4          & {\ul { 10.4}}    & \textbf{10.1} & \textbf{2.0} & 34.8          & 19.2          & 60.1          & 44.3          & 38.1          & 21.3          \\
KGSP                              &    13.2    & {\ul { 2.9}}    & 26.6          & {\ul { 2.1}}    & {\ul { 3.4}}    & {\ul { 0.44}}    & 7.9          & 1.4          & 23.4          & 7.0           & 6.7           & 1.0          & 33.2          & {\ul { 19.8}}    & 58.0          & {\ul { 47.2}}    & {\ul { 39.1}}    & {\ul { 22.9}}    \\
CANet                             & {\ul { 14.3}}    & 2.8      & 27.3          & 1.9           &  3.2         & 0.39    & 8.3          & 1.7          & 25.3          & 6.7           & 6.6              & 1.2          & {\ul {35.1}}         & {\ul {19.8}}    & 58.7            &  46.0       &  38.7       &  22.1    \\
\textbf{Ours}                              & \textbf{16.1} & \textbf{4.0} & \textbf{29.0} & \textbf{2.6} & \textbf{3.8} & \textbf{0.54} & \textbf{8.6} & {\ul { 1.9}}    & {\ul { 27.6}}    & \textbf{10.6} & 7.8     & {\ul { 1.6}}    & \textbf{36.5} & \textbf{22.4} & \textbf{62.2} & \textbf{48.0} & \textbf{39.9} & \textbf{23.6} \\ \hline
\end{tabular}}
\caption{\textbf{Quantitative comparisons in the OW-CZSL setting.} We report the best seen ($S$), best unseen ($U$) accuracy, HM, AUC on the test and validation sub-datasets. 
The best and second-best results are bold and underlined.}
\label{table1}
\end{table*}

\section{Experimental Results}
\label{c3.3}
\subsection{Datasets and Evaluation Metrics}
We conduct experiments on three widely-use datasets including UT-Zappos \cite{utz}, MIT-States \cite{mit}, and C-GQA \cite{le}. 
UT-Zappos is a dataset for the shoes and has 50025 images. It contains 12 object classes and 16 state classes, with 83 seen compositions and a total of 192 compositional spaces. MIT-States has 53753 images with 115 state classes and 245 object classes. The seen and all output compositions are 1,262 and 28,175, respectively. C-GQA is the largest dataset that contains 186,577 images with 413 state classes and 674 object classes. It contains 5,592 seen compositions and a full output space of 278,362 compositions, which makes it the most extensive for the OW-CZSL.
For the \textbf{OW-CZSL}, we follow the splits of \cite{open_cvpr, open_pami, kgsp} and evaluate based on the generalized settings, where the test samples are from both seen and unseen compositions. Considering the performance of the model with different bias factors for the unseen compositions, we vary the bias on the seen composition ($C^s$) during the test phase and report the performance as best seen ($S$), best unseen ($U$), best harmonic mean (HM), and the Area Under the Curve (AUC). For the \textbf{pCZSL}, following \cite{kgsp}, we remove the label and calculate the metrics on the full output composition space ($C$). As we can not access the full-labeled seen compositions ($C^s$), we do not subtract any bias on $C^s$. Therefore, we use the seen (S), unseen (U), and HM metrics.

\subsection{Baselines and Implementation Details}
For \textbf{OW-CZSL}, we compare ProCC with other OW-CZSL methods, including CompCos \cite{open_cvpr}, KGSP \cite{kgsp}, and Co-CGE \cite{open_pami}. CZSL methods are also compared, including LE+ \cite{le}, AoP \cite{ao}, TMN \cite{tmn}, SymNet \cite{sym}, CGE \cite{ge}, and CANet \cite{canet}. For \textbf{pCZSL}, ProCC is compared with KGSP \cite{kgsp} as well as standard (OW-)CZSL methods like CGE \cite{ge}, CompCos \cite{open_cvpr}, and Co-CGE \cite{open_pami}, with the same partial label protocol.  

Following the standard protocols in the CZSL, we utilize the pre-trained ResNet-18 \cite{resnet} as the feature encoder ($\omega$) to extract 512-dimensional feature vectors and learn classifiers on top of these features. Following \cite{ge, kgsp}, each classifier is composed of Multi-Layer Perceptrons (MLP) with three layers with dimensions 768, 512, and the number of output classes, respectively, and comprise Layer Normalization~\cite{ln} and Dropout~\cite{dropout}. To be consistent with other methods, we 
% freeze the encoder $\omega$ and 
randomly augment input images with random crop and horizontal flip. We use PyTorch to implement our network and optimize it with Adam \cite{adam} with default settings. The batch size is 256, and the learning rate is $5.0 \times 10^{-5}$ for the first two stages and $1.0 \times 10^{-5}$ for the third stage. For the UT-Zappos, MIT-States, and C-GQA datasets, the total training time is approximately 1, 3, and 5 hours for 30/60/20, 40/80/30, and 50/100/25 epochs for three stages, respectively, with the early stop strategy.

\subsection{Open-World CZSL (OW-CZSL) Results}
The results of OW-CZSL setting are illustrated in Table \ref{table1}. Generally, closed-world CZSL methods achieve inferior performance, especially in two large datasets (i.e., C-GQA and MIT-States), due to the large cardinality of the output space. ProCC outperforms previous methods on almost all metrics in terms of three datasets. Concretely, as for the most challenging dataset, i.e., C-GQA, the proposed method exceeds the previous SOTA methods, especially for best harmonic (HM) metrics (3.4$\rightarrow$3.8: $\uparrow$12$\%$), which means that ProCC has the better ability to recognize both the seen and unseen compositions. Also, in the validation sub-dataset, Our method suppresses the best baseline (i.e., KGSP) by a large margin in two overall evaluation indexes (i.e., HM: 13.2$\rightarrow$16.1: $\uparrow$22$\%$; AUC: 2.9$\rightarrow$4.0: $\uparrow$38$\%$). As for the MIT-States dataset, our method also has comparative results. Notably, we achieve the best performance on the $U$ metric, which validates the generalization ability of ProCC. For UT-Zappos, it is specially designed for shoes and is relatively simpler than others. ProCC consistently outperforms others, i.e., $S$: 59.3$\rightarrow$62.2; $U$: 47.2$\rightarrow$48.0; HM: 39.1$\rightarrow$39.9; AUC: 22.9$\rightarrow$23.6. 
Remarkably, previous methods typically utilize word embeddings to encode the word expression, which already contains semantic knowledge of similar objects and attributes for composition learning \cite{dve}. Recent Visual Product based method \cite{kgsp} employs more complex classifiers (with hidden layers of 768 and 1024) than ours as well as uses external knowledge to eliminate the less feasibility compositions. We predict the state and object primitives with more lightweight classifiers and explicitly model the cross-primitive interactions to learn the relationship between primitives without external knowledge.

\begin{table*}[]\centering
\tiny
\renewcommand\arraystretch{1}
\setlength{\tabcolsep}{1.5mm}{
\begin{tabular}{l|cccccc|cccccc|cccccc}
\multirow{3}{*}{\textbf{Method}} & \multicolumn{6}{c|}{\textbf{C-GQA}}                                                        & \multicolumn{6}{c|}{\textbf{MIT-States}}                                                  & \multicolumn{6}{c}{\textbf{UT-Zappos}}                                                      \\ \cline{2-19} 
                                 & \multicolumn{3}{c|}{Val}                   & \multicolumn{3}{c|}{Test}                 & \multicolumn{3}{c|}{Val}                     & \multicolumn{3}{c|}{Test}                 & \multicolumn{3}{c|}{Val}                    & \multicolumn{3}{c}{Test}                     \\
                                 & S             & U           & \multicolumn{1}{c|}{HM}           & S             & U            & HM           & S             & U           & \multicolumn{1}{c|}{HM}           & S             & U            & HM           & S             & U            & \multicolumn{1}{c|}{HM}          & S             & U            & HM            \\ \hline
CGE                              & 19.2          & 2.9          & 5.6          & 17.4          & 0.4          & 0.9          & 10.0          & 2.8          & 4.3          & \textbf{19.6}          & 1.3          & 2.4          & 46.5          & 3.5          & 6.6           & 50.3          & 3.4          & 5.0           \\
CompCos                          & 18.2          & 3.0          & 5.2          & 24.3          & 0.4          & 0.7          & 11.1          & 2.9          & 4.6          & 10.8          & 2.0          & 3.6          & 50.2          & 3.9          & 7.3           & 52.4          & 4.1          & 7.6           \\
Co-CGE                           & 19.8          & 3.9          & 6.4          & 22.1          & 0.6          & 1.2          & 14.8          & {\ul { 3.3}}    & {\ul { 5.3}}    & 13.1          & 2.3          & 4.0          & 47.2          & {\ul { 6.1} }   & {\ul { 10.8}}    & 52.6          & 5.4          & 9.9           \\
KGSP                             & {\ul { 20.1}}    & {\ul { 4.8}}    & {\ul { 8.3}}    & {\ul { 22.3}}    & {\ul { 0.9}}    & {\ul { 1.7}}    & {\ul { 15.7}} & 3.2          & {\ul { 5.3}}    & 13.5    & {\ul { 2.6}}    & {\ul { 4.4}}    & {\ul { 49.4}}    & 5.9          & 9.7           & {\ul { 53.8}}    & {\ul { 6.9}}    & {\ul { 12.3}}    \\
\textbf{Ours}                             & \textbf{21.6} & \textbf{5.4} & \textbf{8.7} & \textbf{24.1} & \textbf{1.1} & \textbf{2.0} & \textbf{16.3}    & \textbf{3.5} & \textbf{5.8} & {\ul { 14.1}} & \textbf{2.9} & \textbf{4.8} & \textbf{51.0} & \textbf{7.1} & \textbf{12.5} & \textbf{55.1} & \textbf{8.1} & \textbf{14.1} \\ \hline
\end{tabular}}
\caption{\textbf{Quantitative comparisons in the pCZSL setting.} We report the seen (S), unseen (U) accuracy, and best harmonic mean (HM) on the test and validation sub-datasets.  
The best and second-best results are bold and underlined.}
\label{table2}
\end{table*}

\subsection{Partial-supervision CZSL (pCZSL) Results}
As for the more challenging setting, pCZSL, the challenges come from not only the huge output composition space but also the missing labels. As we can learn from Table \ref{table2}, our method achieves SOTA performances compared with previous CZSL, OW-CZSL, and pCZSL methods. Concretely, for the largest dataset, C-GQA, the performance of SOTAs on pCZSL severely degrades compared with OW-CZSL, even for KGSP, which is equipped with the pseudo label and external knowledge. Our method consistently exceeds them both on validation and testing datasets. For the MIT-States dataset, our method surpasses the second-best method by a large margin in HM metric (i.e., val: 5.3$\rightarrow$5.8:$\uparrow$9$\%$; test: 4.4$\rightarrow$4.8:$\uparrow$9$\%$). For the simplest dataset, UT-Zappos, our method also has the best performance. Note that we do not use any external knowledge like Word2vec, Glove, Conceptnet, and other semi-supervised learning techniques \cite{plabel, semi} for the missing annotations. The superior performance indicates even with partial labels of object and state primitives, our progressive learning strategy can also model the interactions of cross primitives with the pre-trained classifiers.

\begin{table}[] \centering
 \small
\renewcommand\arraystretch{1.0}
\setlength{\tabcolsep}{0.78mm}{
\begin{tabular}{l|cccc|cccccc}
\multicolumn{1}{c|}{\multirow{3}{*}{\textbf{Method}}} & \multicolumn{4}{c|}{\textbf{OW-CZSL}}                                         & \multicolumn{6}{c}{\textbf{pCZSL}}                                                   \\ \cline{2-11} 
\multicolumn{1}{c|}{}                                 & \multicolumn{2}{c|}{C-GQA} & \multicolumn{2}{c|}{MIT-States} & \multicolumn{3}{c|}{C-GQA}         & \multicolumn{3}{c}{MIT-States} \\ \cline{2-11} 
\multicolumn{1}{c|}{}                                 & HM    & \multicolumn{1}{c|}{AUC}   & HM                  & AUC                & S      & U   & \multicolumn{1}{c|}{HM}     & S            & U           & HM         \\ \hline
w/o CPC                                               &  3.3     & \multicolumn{1}{c|}{0.40}      &  6.2                   & 0.8                   & 17.4       & 0.5    & \multicolumn{1}{c|}{1.0}       &11.6              &2.2             &3.7            \\ 
w/o CPI                                               &  3.4     & \multicolumn{1}{c|}{0.41}      &  6.1                   & 0.9                   & 17.7       & 0.5    & \multicolumn{1}{c|}{1.0}       &12.0              &2.1             &3.6            \\ 
w/o CPM                                               &  3.5     & \multicolumn{1}{c|}{0.48}      &  6.6                   & 1.0                   & 18.9       & 0.7    & \multicolumn{1}{c|}{1.4}       &12.2              &2.5             &4.1            \\ 
\hline
w/o P-L                                             &3.7       & \multicolumn{1}{c|}{0.50}      &  7.6                   & 1.5                   &  22.4      & 0.8    & \multicolumn{1}{c|}{1.6}       &12.5              & 2.5            & 4.1           \\
w/ Ex-1\&2                                         &3.6       & \multicolumn{1}{c|}{0.48}      &  7.8                   & 1.5                   &  22.6      & 1.0    & \multicolumn{1}{c|}{1.9}       &13.2              & 2.7            &  4.4          \\
w/o Stage3                                             &3.5       & \multicolumn{1}{c|}{0.47}      &  7.4                   & 1.4                   &  23.2      & 1.1    & \multicolumn{1}{c|}{2.0}       &13.6              & 2.8            &  4.6          \\
w/ 4 Stages                                             &3.6       & \multicolumn{1}{c|}{0.50}      &  7.6                   & 1.4                   &  23.7      & 1.0    & \multicolumn{1}{c|}{1.9}       & 13.8           &2.8             & 4.7           \\
w/ 5 Stages                                             &3.7       & \multicolumn{1}{c|}{0.53}      &  7.7                   & 1.4                   &  23.9      & 1.1    & \multicolumn{1}{c|}{2.1}       & 13.6             &  2.9           &   4.8         \\
w/ 6 Stages                                             &3.8       & \multicolumn{1}{c|}{0.56}      &  7.7                   & 1.6                   &  24.0      & 1.1    & \multicolumn{1}{c|}{2.1}       &   13.8           & 2.8            &  4.7          \\\hline
\textbf{Ours}                                         & 3.8   & \multicolumn{1}{c|}{0.54}  & 7.8                 & 1.6                & 24.1   & 1.1 & \multicolumn{1}{c|}{2.0}    & 14.1         & 2.9         & 4.8        \\ \hline
\end{tabular}}
\caption{\textbf{Ablation studies} for both OW-CZSL and pCZSL.
}
\label{table3}
\end{table}

\section{Ablation Analysis}
We analyze two important components: Cross-Primitive Compatibility (CPC) module and the progressive learning strategy. We adopt the same implementation strategy and conduct the OW-CZSL and pCZSL experiments on the two largest datasets, i.e., C-GQA and MIT-States.

\noindent \textbf{Effect of the Cross-Primitive Compatibility Module.} In Table \ref{table3}, \textbf{\textcircled{\scriptsize{\uppercase\expandafter{\romannumeral1}}}} without the CPC module (w/o CPC), the performance is severely degraded both on the OW-CZSL and pCZSL settings. 
Because lacking the interaction between cross primitives makes the network degenerate to previous Visual Product baselines \cite{ rvp, kgsp}. Meanwhile, KGSP utilizes the external knowledge and surpasses the ablation configuration, especially in pCZSL setting. \textbf{\textcircled{\scriptsize{\uppercase\expandafter{\romannumeral2}}}} Moreover, to further evaluate the conditional modulation, we employ channel attention \cite{senet, ecanet} on the same primitive classifiers without cross-primitive interaction (w/o CPI). \textbf{\textcircled{\scriptsize{\uppercase\expandafter{\romannumeral3}}}} Also, we ablate the learnable cross-primitive memory (w/o CPM) and directly modulate other primitives with learned features. Results indicate that exploring internal primitives brings marginal improvement for composition learning as classifiers have extracted enough internal information, and modulating primitives via hard masks also gives sub-optimal results. Note that the CPC is extremely lightweight with two trainable \rm{1d} convolution layers. Generally, the CPC module greatly improves the performance with negligible computation burden also without external information, which is practical for real-world scenes.

\noindent \textbf{Effect of the Progressive Learning Strategy.} Another important aspect of the ProCC is the progressive learning strategy. From Table \ref{table3}, \textbf{\textcircled{\scriptsize{\uppercase\expandafter{\romannumeral1}}}} we can learn that with the traditional end-end training strategy (w/o P-L), the performance of ProCC degrades to some extent, especially in the pCZSL setting (i.e., HM: 2.0$\rightarrow$1.6 (C-GQA) and 4.8$\rightarrow$4.1 (MIT-States)). As jointly training the whole network under the pCZSL setting does not explicitly learn the relationship between state and object primitives, which is the critical issue in the CZSL task. While for the OW-CZSL setting, joint training induces
some noisy conditioned information, due to the diverse difficulty of classifying object and state primitives.  Also, we exchange the training sequence (i.e., Stage 2 $\rightarrow$ 1 $\rightarrow$ 3) (w/ Ex-1\&2) and ablate the fine-tuning stage (w/o Stage 3). \textbf{\textcircled{\scriptsize{\uppercase\expandafter{\romannumeral2}}}} For the configuration of w/ Ex-1\&2, the performance of ProCC degrades on both settings. Due to the challenge of classifying state primitives \cite{dve, kgsp}, modulation object features conditioned on noisy state features results in invalid interactions.  \textbf{\textcircled{\scriptsize{\uppercase\expandafter{\romannumeral3}}}} For the configuration of w/o Stage 3, where only CPC$_{o \rightarrow s}$ works, the performance degrades to some extent. We have two observations: CPC$_{o \rightarrow s}$ brings more improvements than CPC$_{s \rightarrow o}$; CPC$_{s \rightarrow o}$ and fine-tuning based on well-trained features also matter for the cross-primitive compatibility and global optimum. 
\textbf{\textcircled{\scriptsize{\uppercase\expandafter{\romannumeral4}}}} Moreover, following the same training protocol, we train the network for more stages, i.e., with extra Stage 1 (w/ 4 Stages), extra Stage 1 and 2 (w/ 5 Stages), and extra Stage 1, 2, and 3 (w/ 6 Stages).  We see that more training stages can not bring much accuracy improvement, as the model has converged after Stage 3. 

\section{Chapter Summary}

% In this chapter, we aim to enhance the multimodal composition generalization ability of neural networks. We propose a method named Progressive Cross-primitive Compatibility (ProCC) network for both OW-CZSL and pCZSL tasks. The simple but effective Cross-Primitive Compatibility module drives the network learning to predict feasible object and state primitives conditioned on mutual relations. Also, the progressive learning strategy significantly eliminates the invalid cross-primitive interactions in pCZSL and noisy conditioned information, in an easy-hard learning manner. Comprehensive experiments on OW-CSZL and pCZSL settings illustrate superior performance compared with other state-of-the-art methods in terms of composition generalization.

This chapter proposes the Progressive Cross-Primitive Compatibility (ProCC)  framework to enhance multimodal compositional generalization in open-world scenarios, addressing both Open-World (OW-CZSL) and partially supervised (pCZSL) Compositional Zero-Shot Learning. ProCC introduces a Cross-Primitive Compatibility (CPC) module that models conditional dependencies between object and state modality primitives (e.g., inferring "ripe" only for edible objects) through self-supervised visual-semantic correlations, eliminating reliance on external linguistic resources. Complemented by a progressive learning strategy, the framework adopts a curriculum-driven, easy-to-hard paradigm—first learning coarse-grained primitive distinctions (e.g., "metal" vs. "wood") before refining fine-grained compatibility constraints (e.g., "polished" vs. "rusty")—effectively suppressing invalid compositions (e.g., "flying tables") and noisy supervision. Experiments on benchmarks like MIT-States and CGQA demonstrate ProCC’s superiority, outperforming state-of-the-art methods by a large margin in accuracy under OW-CZSL settings and reducing invalid predictions by 27$\%$ in pCZSL. By addressing combinatorial complexity and noisy adaptation through adaptive cross-modal conditioning, ProCC advances the thesis’s core theme of robust open-world learning, and in the subsequent chapters, we introduce the cross-modal knowledge distillation for missing or invalid modality situations (Chapter \ref{Chapter4}: C$^2$KD) and Self-Introspective Decoding for multimodal large model hallucination alleviation (Chapter \ref{Chapter5}: SID).

\chapter{\textsc{C$^2$KD}: Bridging the Modality Gap for Cross-Modal Knowledge Distillation}
\label{Chapter4}

\section{Challenges and Motivations} 
\label{c4.1}
Knowledge Distillation (KD) is an effective approach to transfer knowledge from the large-capacity teacher model to the low-capacity student model during training \cite{survey1_kd, survey2_kd}. During the KD process, the student is trained to mimic the teacher’s output via the distillation loss. KD methods can be divided into two main categories: logits-based and feature-based methods. The former minimizes the discrepancy between soft labels of the teacher model and the student model \cite{kl, dml, dist}, and the latter distills knowledge from intermediate feature layers \cite{fd, review, diffusion}. 

Despite the success of traditional KD methods in single modality scenario, extending these methods to address the Cross-Modal Knowledge Distillation (CMKD) tasks remains a critical challenge.
The CMKD task involves knowledge transfer from one modality to another during the distillation phase, with inference \textit{only} on the \textit{distilled student modality}, which is crucial especially in computation-constrained and sensor-failure scenarios.
As demonstrated in Table \ref{tab1}, 
unimodal KD methods struggle to transfer knowledge from the low-accuracy visual modality to the high-accuracy audio modality, while the visual modality has only marginal gains from the audio modality.
Based on the above analysis, a pivotal and fundamental question arises: \textit{Can we effectively transfer arbitrary unimodal information to another modality?}

\begin{table}[] \centering 
\begin{tabular}{l|clc|ll}
                                     & \multicolumn{3}{c|}{\textbf{AVE}\cite{ave}}                             & \multicolumn{2}{c}{\textbf{VGGSound}\cite{vggsound}}                                    \\ \hline
 & \multicolumn{2}{c}{\textbf{Visual}} & \textbf{Audio}          & \multicolumn{1}{c}{\textbf{Visual}} & \multicolumn{1}{c}{\textbf{Audio}} \\ 
\multicolumn{1}{l|}{\textbf{Method}} & \multicolumn{2}{c}{(\textbf{A$\to$V})} & (\textbf{V$\to$A})          & \multicolumn{1}{c}{(\textbf{A$\to$V})} & \multicolumn{1}{c}{(\textbf{V$\to$A})} \\\hline
w/o KD                              & \multicolumn{2}{c}{31.6{\scriptsize±0.18}}           & 52.8{\scriptsize±0.11}                    & \multicolumn{1}{c}{38.7{\scriptsize±0.16}}            & \multicolumn{1}{c}{59.4{\scriptsize±0.16}}           \\
KD \cite{kl}                              & \multicolumn{2}{c}{32.3{\scriptsize±0.35}}                & \multicolumn{1}{c|}{46.6{\scriptsize±0.24}}   &38.5{\scriptsize±0.50}                                     & 56.3{\scriptsize±0.46}                                   \\
Review \cite{review}               & \multicolumn{2}{c}{32.1{\scriptsize±0.63}}                & \multicolumn{1}{c|}{50.6{\scriptsize±0.31}}               & 38.2{\scriptsize±0.47}                & 57.9{\scriptsize±0.33}  \\
DML \cite{dml}                                  & \multicolumn{2}{c}{31.8{\scriptsize±0.41}}       & 48.0{\scriptsize±1.31}               & 38.7{\scriptsize±0.86}                           & 58.2{\scriptsize±1.01}                          \\
SHAKE \cite{shake}                                  & \multicolumn{2}{c}{32.2{\scriptsize±0.59}}       & 47.3{\scriptsize±0.72}               &  38.3{\scriptsize±0.41}                          &  59.5{\scriptsize±0.34}                         \\
DKD \cite{dkd}                                 & \multicolumn{2}{c}{32.6{\scriptsize±0.65}}                & 48.6{\scriptsize±1.02}                        & 38.1{\scriptsize±0.43}                                    &57.2{\scriptsize±0.86}                                    \\
DIST \cite{dist}                                & \multicolumn{2}{c}{29.8{\scriptsize±0.61}}       & 49.3{\scriptsize±0.52}              & 38.5{\scriptsize±0.39}                           & 58.9{\scriptsize±0.45}                          \\ 
NKD \cite{normkd}         & \multicolumn{2}{c}{32.9{\scriptsize±0.32}}       & \multicolumn{1}{c|}{52.2{\scriptsize±0.62}}      & 39.2{\scriptsize±0.52}       & 59.3{\scriptsize±0.40}    \\ \hline
Ours                                 & \multicolumn{2}{c}{34.7{\scriptsize±0.23}}       & 54.9{\scriptsize±0.16}              & 40.9{\scriptsize±0.31}                           & 61.9{\scriptsize±0.27}                         \\ \hline
\end{tabular}
\caption{\textbf{Performances of traditional KD in CMKD.} 
The results of distilled student modality infer only on the student modality.
\textbf{A$\to$V}: Audio teacher modality distills visual student modality; \textbf{V$\to$A}: Visual teacher modality distills audio student modality.
}
\label{tab1}
\end{table}

\begin{figure}
  \centering
  \includegraphics[width=0.75\textwidth]{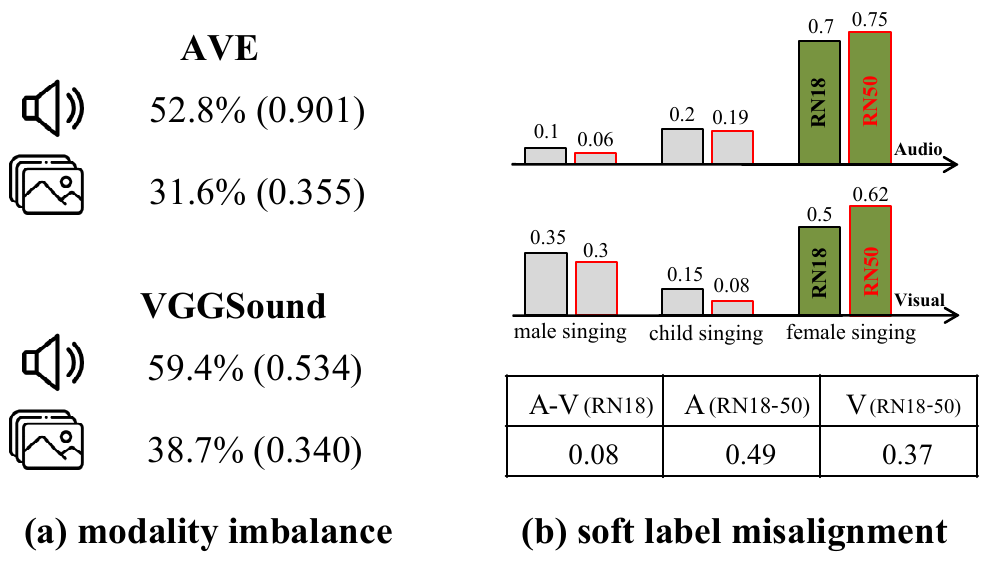}
  \caption{\textbf{The Modality Gap of CMKD.} \textbf{(a)} Top-1 accuracy (followed by average prediction probability of target classes) of each modality. Both modalities utilize ResNet-18 as the backbone. \textbf{(b)} Up: Example of three-class classification. Down: Kendall Rank Correlation \cite{kendall} of soft labels across modalities in VGGSound. A: audio; V: visual; RN: ResNet.}
  \label{motivation}
\end{figure}

To answer this question, we conduct empirical analysis to investigate why traditional KD methods fail in CMKD from the logits-based perspective, which can be attributed to the inter-modality gap that inducing \textit{modality imbalance} and \textit{soft label misalignment}, as illustrated in Figure \ref{motivation}.

For the first factor, we define \textit{modality imbalance}, akin to \cite{on_the_fly_multimodal, pmr}, as the performance disparities between modalities.
We quantitatively calculate the top-1 accuracy (followed by the average prediction probability of target classes) after training on the corresponding single modality.
Figure \ref{motivation}(a) shows that the audio modality outperforms the visual modality in AVE and VGGsound datasets, and there are significant gaps in the average prediction probability of the target class, particularly in AVE. Merely distilling knowledge from the visual to the audio modality could potentially yield adverse effects, as shown in Table \ref{tab1} (Audio columns).

For the second factor, we define \textit{soft label} as the output distributions from the teacher network, following \cite{kl, zipf}. The soft labels contain meaningful information on similarity among various classes. However, inter-modality gap leads to severe soft label misalignment between teacher and student modalities. Take three-class classification as an example (Figure \ref{motivation}(b) Up). Although both Audio and Visual modalities branches successfully predict the target class of `female singing', the non-target soft labels are rank-distorted, where the \textit{audio accent} of `child singing' is more closely related to `female singing', while the \textit{visual appearance} of `male singing' is more closely resembles `female singing'. Direct transferring soft label information across modalities is unreasonable, which could explain why distilling the audio modality to the visual modality does not yield significant improvements.
To quantitatively validate soft label misalignment, we further calculate the average Kendall Rank Correlation ($\mathrm{KRC}$) \cite{kendall} of soft labels in Figure \ref{motivation}(b) Down. A higher $\mathrm{KRC}$ indicates better rank correlation. The table indicates the $\mathrm{KRC}$ of multimodal soft label (i.e., A-V(RN-18)) is significantly lower than that of a single modality with diverse-capacity networks (i.e., A(RN18-50) and V(RN18-50)), indicating the presence of misalignment of multimodal soft labels.

To address the above issues in CMKD, this chapter proposes Customized cross-modal Knowledge Distillation (C$^2$KD). 
Concretely, instead of using the pre-trained teacher to provide supervision signals to the student, we bidirectionally update to customize both the pre-trained teacher and student via On-the-Fly Selection Distillation (OFSD) strategy, where OFSD selectively distill receptive soft labels according to the Kendall Rank Correlation, and cross-modal knowledge is transferred from non-target classes to avoid the modality imbalance issue. Furthermore, Proxy student and teacher, inheriting unimodal and cross-modal knowledge, is formulated to progressively transfer cross-modal knowledge in the bidirectional distillation form. 

The main contributions of this chapter can be summarized as follows.

\begin{itemize}
\item We empirically analyze the factors for the failure of unimodal KD in CMKD, which can be attributed to the modality imbalance and soft label misalignment.

\item To address these issues, we propose a novel method named C$^2$KD. Specifically,  OFSD produces selected crossmodel non-target class knowledge through on-the-fly bidirectionally distilling both student and teacher. Moreover, Proxy student and teacher are built to progressively transfer receptive knowledge across modalities. The proposed strategies are plug-and-play, enhancing traditional KD methods in CMKD.  

\item We conduct experiments on sparse and dense prediction tasks, including audio-visual, image-text, and RGB-Depth datasets. 
Diverse capacities and homogeneous/heterogeneous architectures are also considered. Extensive experiments validate C$^2$KD can transfer cross-modal knowledge from \textit{arbitrary} modality to another.

\end{itemize}

\section{Cross-Modal KD Effectiveness Analysis}
\label{c4.2}

First, this sub-chapter revisits traditional KD in cross-modal scenario. Given multimodal training data ([$X_1$, $X_2$], $Y$) containing multimodal samples $X_1$ and $X_2$ and labels $Y$. Let $f_T$ and $f_S$ be the output logits of the teacher $T$ and student $S$. The corresponding prediction probabilities are obtained using the softmax function ($\sigma$): $p_{S}=\sigma(f_{S})$ and $p_{T}=\sigma(f_{T})$.  Typical KD trains the student network as follows:
\begin{eqnarray}
L_{KD}=\mathcal{H}(p_{S},Y)+ \lambda \mathcal{D}(p_{S},p_{T})
\label{eq1}
\end{eqnarray}
where $\mathcal{H}$ is the supervision loss function (typical Cross-Entropy (CE) loss), $\mathcal{D}$ is the KD loss to minimize the discrepancy of output distribution between teachers and students, commonly achieved using Kullback–Leibler (KL) divergence \cite{kl}, and $\lambda$ is a balancing parameter for these two terms. Pioneering work \cite{mfh} proposes the Modality Focusing Hypothesis (MFH) and claims that modality-general decisive features are crucial for transferring knowledge across modalities during the distillation phase. In this work, we provide another fine-grained perspective to investigate the efficacy of CMKD: the modality gap, which refers to the \textit{modality imbalance} in target-class logits and \textit{soft label misalignment}, incurs the failure of CMKD. 

Regarding \textit{modality imbalance}, as depicted in Figure \ref{motivation}(a), the prediction possibility of the target class exhibits significant variations across modalities.

If simply let student modality (audio) imitate teacher modality (vision), audio will inevitably reduce prediction confidence \cite{tent} and conflict \textit{one-hot} label ($Y$). To validate our claim, we follow DKD \cite{dkd} and decouple the KD loss into Target Class (TC) and Non-target Class (NC) KD:
\begin{eqnarray} 
\begin{split}
\mathcal{D}(f_{S},f_{T})&=\alpha [\underbrace{ p_{T}^{t}log(\frac{p_{T}^{t}}{p_{S}^{t}} )+p_{T}^{\setminus t}log(\frac{p_{T}^{\setminus t}}{p_{S}^{\setminus t}} )}_{\mathrm{TCKD}}] \\
&= \beta\underbrace{\sum_{i=1,i\ne t}^{C} \hat{p}_{T}^{i} log(\frac{\hat{p}_{T}^{i}}{\hat{p}_{S}^{i}} )}_{\mathrm{NCKD}}
\label{eq2}
\end{split}
\end{eqnarray}
where $\alpha$ and $\beta$ are hyperparameters. $p^t$ denotes the target class probability: $p^{t}=\mathrm{exp} (f^{t})/\sum_{j=1}^{C} \mathrm{exp} (f^{j})$, $p^{\setminus t}$ represents the probability of all the other non-target classes $p^{\setminus t}=\sum_{k=1, k\ne t}^{C} \mathrm{exp} (f^{i})/\sum_{j=1}^{C} \mathrm{exp} (f^{j})$, and $\hat{p}^{i}$ means the probability among non-target classes: $\hat{p}^{i}=\mathrm{exp} (f^{i})/\sum_{j=1, j \ne t}^{C} \mathrm{exp} (f^{j})$. Here, $C$ is the number of classes. When \textit{only} applying TCKD in CMKD, as shown in Table \ref{tab2}, the performance of distilled audio modality severely degrades 4.8$\%$ and 3.6$\%$, respectively, while the distilled visual modality is not clearly enhanced. Therefore, \textit{modality imbalance} hinders the efficiency of CMKD, particularly when transferring knowledge from a low-accuracy modality to a high-accuracy modality.

To analyze \textit{soft label misalignment}, we only conduct NCKD (Equation \ref{eq2}) to exclude the influence of modality imbalance. As depicted in Table \ref{tab2}, the low-accuracy visual teacher modality degrades the performance of the high-accuracy audio student modality. \textit{Notably}, distilling high-accuracy audio information into the low-accuracy visual modality only results in marginal gains in the AVE, while surprisingly exhibiting a degradation in the VGGsound.
\cite{kl, lsr, zipf, dkd} investigate the mechanism of logit distillation, as soft logits provide reliable similarity information between categories. The privileged similarity information brings fine-grained supervision compared to a one-hot label. However, in the context of CMKD, the category similarities between different modalities are varied and even \textit{conflicting}. An intuitive example is the three-class classification example in Figure \ref{motivation}(b) Up, where the unreliable similarity information of non-target classes across the modalities is contradictory. Directly minimizing cross-modal distributions leads to performance degradation. To quantitatively evaluate the misalignment of soft labels, we employ the Kendall Rank Correlation ($\mathrm{KRC}$) \cite{kendall} metric to measure the rank correlation. Specifically, given teacher and student output logits $f_{T}$ and $f_{S}$, the $\mathrm{KRC}$ between $f_{T}$ and $f_{S}$ can be explicitly computed as follows:
\begin{eqnarray} \small
\mathrm{KRC}\!=\!\frac{2}{C(C-1)}\!\sum_{i<j}\!\mathrm{sign} (f_{T}^{i}\!-\!f_{T}^{j})\mathrm{sign} (f_{S}^{i}\!-\!f_{S}^{j})  
\label{eq3}
\end{eqnarray}
As depicted in Figure \ref{motivation}(b) Down, the $\mathrm{KRC}$ between multimodal networks is significantly lower than that observed in unimodal networks with different capacities. We argue that the misalignment of rank correlation is another reason for the failure of CMKD. To validate our argument, we filter out multimodal samples with $\mathrm{KRC}$ $<$ 0 (+$\mathrm{KRC}$), indicating that the count of misaligned soft label pairs is larger than aligned ones. Additionally, we randomly filter out the same number of samples (+$\mathrm{Random}$). From Table \ref{tab2}, we can see that both visual and audio modalities are improved when guided by the $\mathrm{KRC}$ metric, whereas randomly filtering out samples has almost no effect.

\begin{table}[]\centering 
\begin{tabular}{l|cccc|cccc}
                                     & \multicolumn{4}{c|}{\textbf{AVE} \cite{ave}}                                          & \multicolumn{4}{c}{\textbf{VGGsound} \cite{vggsound}}                                     \\ \hline
\multicolumn{1}{c|}{\textbf{}} & \multicolumn{2}{c|}{\textbf{Visual}} & \multicolumn{2}{c|}{\textbf{Audio}} & \multicolumn{2}{c|}{\textbf{Visual}} & \multicolumn{2}{c}{\textbf{Audio}} \\ 
\multicolumn{1}{c|}{\textbf{Method}} & \multicolumn{2}{c|}{(\textbf{A$\to$V})} & \multicolumn{2}{c|}{(\textbf{V$\to$A})} & \multicolumn{2}{c|}{(\textbf{A$\to$V})} & \multicolumn{2}{c}{(\textbf{V$\to$A})} \\ 
\hline
% $Proba.$                          & \multicolumn{2}{c|}{0.355}               & \multicolumn{2}{c|}{0.901}               & \multicolumn{2}{c|}{0.340}                          & \multicolumn{2}{c}{0.534}               \\ \hline
w/o KD                               & \multicolumn{2}{c|}{31.6}               & \multicolumn{2}{c|}{52.8}               & \multicolumn{2}{c|}{38.7}                          & \multicolumn{2}{c}{59.4}               \\ 
$proba.$                          & \multicolumn{2}{c|}{0.355}               & \multicolumn{2}{c|}{0.901}               & \multicolumn{2}{c|}{0.340}                          & \multicolumn{2}{c}{0.534}               \\ \hline
w/ KD                                & 32.3   & \multicolumn{1}{c|}{$\uparrow$0.7}   & 46.6             & $\downarrow$6.2             & 38.5   & \multicolumn{1}{c|}{$\downarrow$0.2}   & 56.3             & $\downarrow$3.1            \\ 
+${\mathrm{Random}}$                        & 32.1   & \multicolumn{1}{c|}{-0.2}   & 46.8             & +0.2                         & 38.2   & \multicolumn{1}{c|}{-0.3}               & 56.4             & +0.1            \\ 
+${\mathrm{KRC}}$                           & 32.9   & \multicolumn{1}{c|}{+0.6}            & 47.9             & +1.3                        & 39.2   & \multicolumn{1}{c|}{+0.7}             & 57.4              & +1.1            \\ \hline
TCKD                                 & 31.8   & \multicolumn{1}{c|}{$\uparrow$0.2}   & 48.0             & $\downarrow$4.8             & 37.9   & \multicolumn{1}{c|}{$\downarrow$0.8}   & 55.8             & $\downarrow$3.6            \\ \hline
NCKD                                 & 31.9   & \multicolumn{1}{c|}{$\uparrow$0.3}   & 50.1             & $\downarrow$2.7             & 38.5   & \multicolumn{1}{c|}{$\downarrow$0.2}   & 57.5               & $\downarrow$1.9            \\
+${\mathrm{Random}}$                       & 31.5   & \multicolumn{1}{c|}{-0.4}      & 50.2             & +0.1                       & 38.5   & \multicolumn{1}{c|}{-}                  & 57.6               & +0.1            \\
+${\mathrm{KRC}}$                           & 33.1   & \multicolumn{1}{c|}{+1.2}            & 51.0             & +0.9                        & 39.6   & \multicolumn{1}{c|}{+1.1}             & 58.3              & +0.8            \\ \hline
DKD \cite{dkd}                              & 32.6    & \multicolumn{1}{c|}{$\uparrow$1.0}   & 48.6             & $\downarrow$4.2             & 38.1   & \multicolumn{1}{c|}{$\downarrow$0.6}                 & 57.2             & $\downarrow$2.2            \\ \hline
\end{tabular}
\caption{\textbf{Efficacy Analysis on modality imbalance and soft label misalignment.} $proba.$ represents average prediction probability of target class. DKD is with defaulted $\{\alpha=1, \beta=8\}$ (Eq. \ref{eq2}).}
\label{tab2}
\end{table}

\section{Customized Cross-modal Knowledge Distillation (C$^2$KD)}
\label{c4.3}

\begin{figure*}
  \centering
  \includegraphics[width=1.0\textwidth]{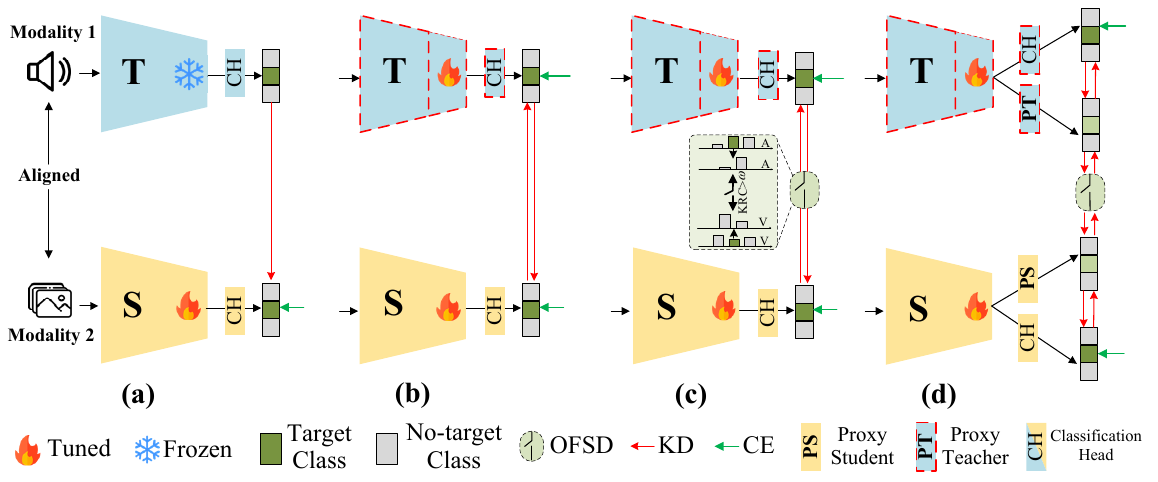}
  \caption{\textbf{Evolution of our Customized Cross-modal Knowledge Distillation (C$^2$KD) method.} (a) Traditional KD \cite{kl} with output logits from the fixed teacher. (b) We (partially) tune the teacher with the bidirectional distillation to provide customized teacher knowledge. (c) To bridge the modality gap of CMKD, On-the-Fly Selection Distillation (OFSD) is proposed to filter out samples with distorted rank correlations and perform KD on non-target classes. (d) Additionally, we introduce proxy teacher and proxy student as bridges to progressively transfer receptive cross-modal knowledge.}
  \label{framework}
\end{figure*}

Based on the aforementioned analysis, this chapter proposes a simple yet effective method named Customized Cross-modal Knowledge Distillation (C$^2$KD) to transfer cross-modal knowledge to an arbitrary single modality. To bridge the modality gap, 
we argue that both student and teacher should be tuned with the bidirectional distillation from each other, 
% we argue that the pre-trained teacher performs bidirectional distillation with student to provide customized knowledge.
in this way, teacher modality could provide receptive information for student modality. Meanwhile, the soft label misalignment samples should be filtered out otherwise induce conflicting information. Therefore, we propose the On-the-Fly Selection Distillation (OFSD) strategy to exclude non-distillable samples and inherit knowledge from non-target classes. Furthermore, dual proxies with the bidirectional distillation strategy are introduced to progressively transfer cross-modality knowledge. The evolution of our proposed framework is depicted in Figure \ref{framework}.

\subsection{Formulation of C$^2$KD} As illustrated in Figure \ref{framework}(d), C$^2$KD proposes the OFSD strategy to dynamically select receptive knowledge. 
This strategy involves distilling knowledge from non-target classes and innovatively employing the Kendall Rank Correlation ($\mathrm{KRC}$) \cite{kendall} metric to filter out samples with rank-distorted soft labels. 
Given the output logits $f_{T}$ and $f_{S}$ from the teacher and student modalities, the sample selection strategy is as follows:
\begin{eqnarray}
\eta = \begin{cases}
1,&\mathrm{KRC} (f_{T}, f_{S})>\omega  \\
0,&otherwise\\
\end{cases}
\label{eq4}
\end{eqnarray}
The $\mathrm{KRC}$ is as Equation \ref{eq3}, $\eta\in \{0,1\}$ is OFSD filter, and $\omega$ is the threshold.

Moreover, we additionally build dual proxies to progressively produce soft labels. Formally, the output features ($F$) obtained from the backbone ($B$) are fed to the original classification head and the proposed proxy as follows:
\begin{eqnarray}
\begin{split}
f_{m} &= fc_{m}^{cls}(\mathrm{GAP} (B_{m}(F_{m}))), m\in \left \{\mathrm{T} ,\mathrm{S}  \right \} \\
f_{m}^{pro}&=fc_{m}^{cls(pro)}(\mathcal{A} [\mathrm{GAP} (B_{m}(F_{m}))]), m\in \left \{\mathrm{T} ,\mathrm{S}  \right \}
\label{eq5}
\end{split}
\end{eqnarray}
where $\mathrm{GAP}$ and $fc^{cls}$ refer to global average pooling and classification head. $\mathcal{A}$ represents feature adaptation layer, akin to \cite{fitnet, shake}, consisting of the `Conv-BN-ReLU' block. To further produce customized knowledge, both student and teacher proxies serve as \textit{bridges} and get bidirectional distillation from both uni-modality and cross modality. In summary, the total loss function can be expressed as follows:
\begin{equation} 
\begin{split}\label{eq6}
L_{all} =
& \mathcal{H} (\sigma(f_{S}),Y)+\mathcal{H} (\sigma(f_{T}),Y)\\ 
&+\lambda_{1} \mathcal{D} (\sigma(f_{T}),\sigma(f_{T}^{pro}))+\lambda_{1} \mathcal{D} (\sigma(f_{T}^{pro}), \sigma(f_{T}))\\
&+\lambda_{2} \mathcal{D} (\sigma(f_{S}),\sigma(f_{S}^{pro}))+\lambda_{2} \mathcal{D} (\sigma(f_{S}^{pro}), \sigma(f_{S}))\\
&+\lambda_{3}\eta\mathcal{D} (\sigma(\hat{f}_{S}^{pro(i)}),\sigma(\hat{f}_{T}^{pro(i)})) \\
&+\lambda_{3}\eta\mathcal{D} (\sigma(\hat{f}_{T}^{pro(i)}),\sigma(\hat{f}_{S}^{pro(i)}))
\end{split}
\end{equation}
where $\lambda_{1}$, $\lambda_{2}$, and $\lambda_{3}$ are balancing parameters and $i\ne t$. $\mathcal{H}$ and $\mathcal{D}$ represent supervision and KD loss, respectively. We simply set $\{ \lambda_{1}=\lambda_{2}=\lambda_{3}=1\}$ in all experiments.

% \begin{figure}[ht] % [ht] 是放置位置参数，ht表示“here top”
%     \centering % 居中图片
%     \begin{subfigure}{0.45\textwidth} % 第一个子图，占0.45\textwidth宽度
%         \centering
%         \includegraphics[width=\textwidth]{Figures/c4_3_1} % 替换为你的图片路径
%         \caption{(a)}
%     \end{subfigure}
%     \hfill % 水平填充，确保子图之间有空间
%     \begin{subfigure}{0.45\textwidth} % 第二个子图，占0.45\textwidth宽度
%         \centering
%         \includegraphics[width=\textwidth]{Figures/c4_3_2} % 替换为你的图片路径
%         \caption{(b)}
%     \end{subfigure}
    
%     \vspace{10pt} % 垂直空间，可选，用于在两组子图之间添加空间
    
%     \begin{subfigure}{0.45\textwidth} % 第三个子图，占0.45\textwidth宽度
%         \centering
%         \includegraphics[width=\textwidth]{Figures/c4_3_3} % 替换为你的图片路径
%         \caption{(c)}
%     \end{subfigure}
%     \hfill % 水平填充
%     \begin{subfigure}{0.45\textwidth} % 第四个子图，占0.45\textwidth宽度
%         \centering
%         \includegraphics[width=\textwidth]{Figures/c4_3_4} % 替换为你的图片路径
%         \caption{(d)}
%     \end{subfigure}
    
%     \caption{\textbf{Training dynamics analysis.} The \ul{solid lines} correspond to the test accuracy, and the \dashuline{dotted lines} indicate the average number of samples with $\mathrm{KRC}<\omega$ each data batch during the training process. Here we set $\{\omega=0, \mathrm{batchsize}=64 \}$.} % 主图的标题
%     \label{train}
% \end{figure}

\begin{figure*}[] \centering
  \begin{minipage}[]{0.48\textwidth}
    \centering
    \includegraphics[width=\textwidth]{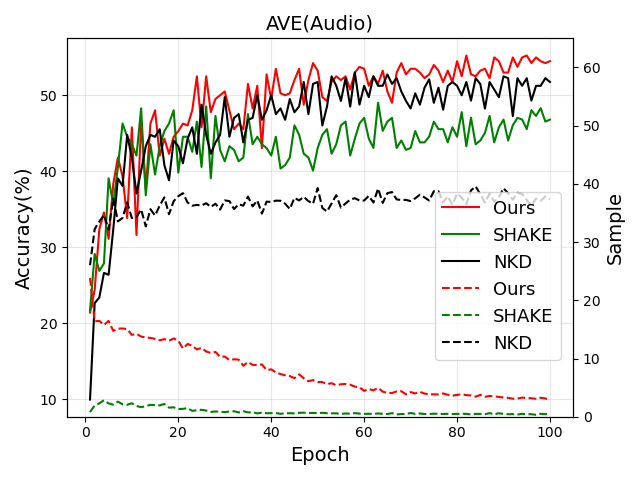}
    \caption*{(a)}
  \end{minipage}\hfill
  \begin{minipage}[]{0.48\textwidth}
    \centering
    \includegraphics[width=\textwidth]{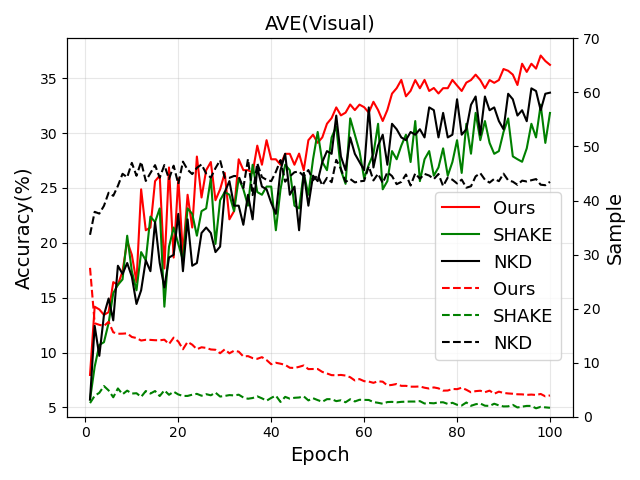}
    \caption*{(b)}
  \end{minipage}
    \begin{minipage}[]{0.48\textwidth}
    \centering
    \includegraphics[width=\textwidth]{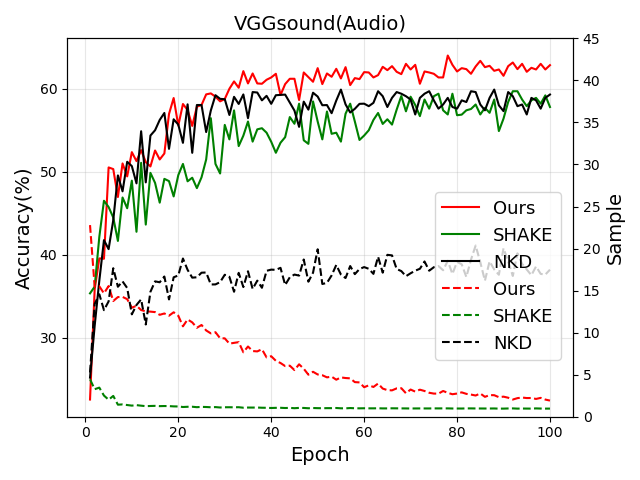}
    \caption*{(c)}
  \end{minipage}\hfill
  \begin{minipage}[]{0.48\textwidth}
    \centering
    \includegraphics[width=\textwidth]{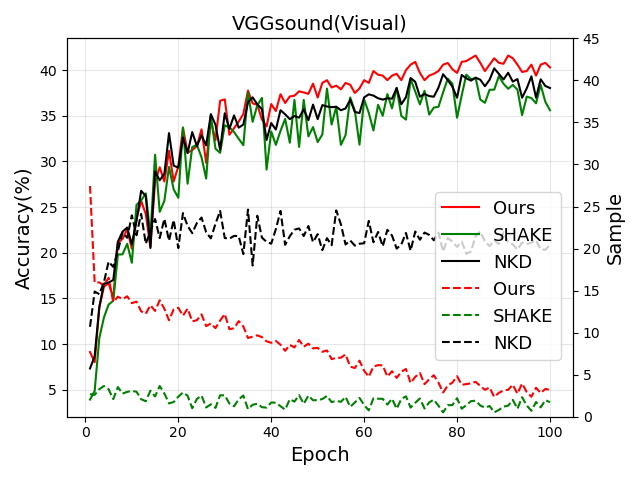}
    \caption*{(d)}
  \end{minipage}

\caption{\textbf{Training dynamics analysis.} The \ul{solid lines} correspond to the test accuracy, and the \dashuline{dotted lines} indicate the average number of samples with $\mathrm{KRC}<\omega$ each data batch during the training process. Here we set $\{\omega=0, \mathrm{batchsize}=64 \}$.}
\label{train}
\end{figure*}

\subsection{Analysis of Cross-modal Knowledge distillation}

\textbf{Understanding CMKD training dynamics.}
We visualize the training dynamics of CMKD and compare it with SHAKE \cite{shake} and NKD \cite{normkd} to demonstrate the CMKD progress. Figure \ref{train} shows the test accuracy and the average number of samples with $\mathrm{KRC}<\omega$ ($\omega=0$) during the training process.
As the advanced online KD, SHAKE gets the reverse cross-modal feedback supervision without discrimination. 
However, SHAKE suffers from severe instability of training, possibly due to conflicting cross-modal information. 
Meanwhile, the sample number of $\mathrm{KRC}<\omega$ drops to close to $0$ within initial epochs, which represents the teacher modality is influenced by the student modality and might lose teacher modality information.
In contrast, NKD minimizes the distance between student modality logits and teacher modality logits. The teacher model of NKD is not updated to cater to student modality, so the sample number of $\mathrm{KRC}<\omega$ is large, and NKD also falls into the unstable training process. As for ours, we selectively inherit cross-modal knowledge based on $\mathrm{KRC}$ and progressively update the teacher model through proxies to obtain receptive knowledge. 
During the distillation progress, the rank-distorted samples gradually reduce, and our method only filters out the non-distillable samples.

\begin{figure}[ht]
  \centering
  \includegraphics[width=0.9\textwidth]{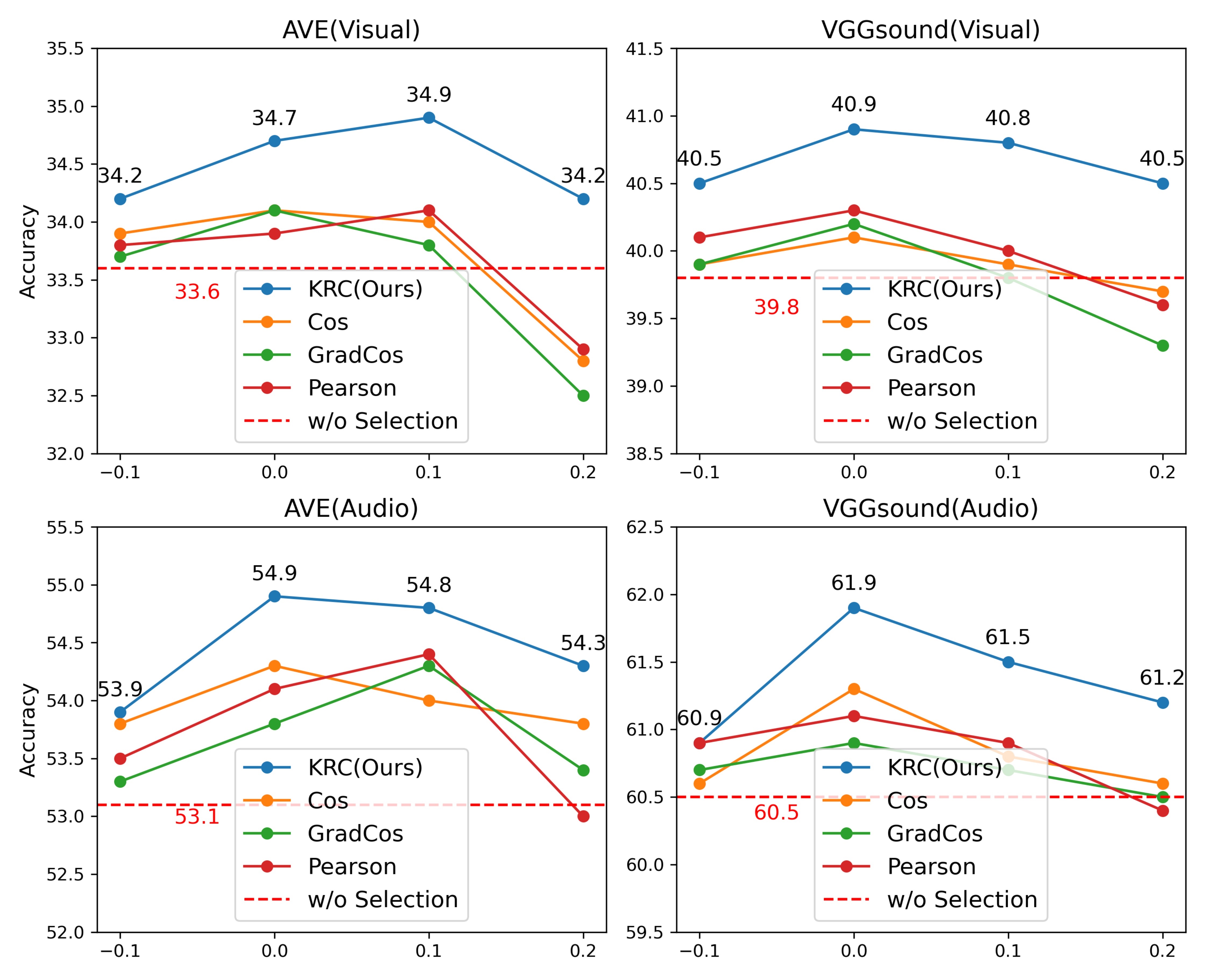}

  \caption{\textbf{Comparisons of different distance metrics.} The X-axis represents the value of $\omega$.
  % The Y-axis represents the top-1 accuracy ($\%$).
  }
  \label{omega}
\end{figure}

\textbf{Comparisons with other distance metrics.}
We select other distance metrics to verify the effectiveness of $\mathrm{KRC}$ defined in Equation \ref{eq4}. Concretely, we choose the cosine similarity ($\mathrm{Cos}$), gradient cosine similarity ($\mathrm{GradCos}$), and Pearson correlation coefficient \cite{pearson} ($\mathrm{Pearson}$) as alternatives. Cosine similarity and Pearson correlation coefficient are used to measure the distance between teacher and student logits, ranging from -1 to 1. They can be formulated as: $\amalg(f_{T}, f_{S})>\omega$, $\amalg \in \left \{{\mathrm{Cos}; \mathrm{Pearson}}\right \} $.
Similar to \cite{pcgrad,sckd}, Gradient cosine similarity regards CMKD (Equation \ref{eq1}) as two tasks: 
cross-modal distillation ($L_{cmkd}=\mathcal{D}(p_{S},p_{T})$) and unimodal task ($L_{task}=\mathcal{H}(p_{S},Y)$) and calculates the gradient cosine similarity between these two tasks as: $\mathrm{Cos}(\nabla_{\theta}L_{cmkd}, \nabla_{\theta}L_{task})>\omega$. It's worth noting that these three metrics consider \textit{both} rank and intensity between cross-modal logits, while $\mathrm{KRC}$ \textit{only} concerns about rank-distorted ones. As shown in Figure \ref{omega}, $\mathrm{KRC}$ makes the best performance among these metrics. Although inferior to $\mathrm{KRC}$ metric, other metrics with proper $\omega$ perform better than without sample selection ($\mathrm{w/o}\ \mathrm{Selection}$) strategy. The results validate the necessity of filtering out samples with misaligned soft labels.

\section{Experimental Results}
\label{c4.4}

We conduct extensive experiments to validate the effectiveness of our method. First, we compare our method with KD methods regarding multimodal classification tasks.
Also, we apply our method to the multimodal semantic segmentation. Then, we perform ablation and sensitivity analysis.

\subsection{Multimodal Classification}

We follow \cite{on_the_fly_multimodal, pmr, mcce} and conduct experiments on four visual-audio and image-text datasets: (1) \textbf{CREMA-D} \cite{crema-d} is an audio-visual dataset for speech emotion recognition, with 6 categorizations. (2) \textbf{AVE} \cite{ave} is an audio-visual dataset for audio-visual event localization, in which there are 28 event classes. (3) \noindent \textbf{VGGsound} \cite{vggsound} is a large-scale video dataset containing 309 classes covering daily life activities. We randomly choose \textit{50} class to conduct experiments due to limited computation resources. (4) \noindent \textbf{CrisisMMD} \cite{crisis} is a multimodal crisis prediction dataset and is divided into eight humanitarian categories.

\begin{table*}[] \centering \tiny
\renewcommand\arraystretch{0.85}
\begin{tabular}{c|clcl|clcl|clcl|clcl}
                & \multicolumn{4}{c|}{\textbf{CREMA-D}}                                     & \multicolumn{4}{c|}{\textbf{AVE}}                                         & \multicolumn{4}{c|}{\textbf{VGGsound}}                                    & \multicolumn{4}{c}{\textbf{CrisisMMD}}                                 \\ \hline
\textbf{Method} & \multicolumn{2}{c}{\textbf{Visual}} & \multicolumn{2}{c|}{\textbf{Audio}} & \multicolumn{2}{c}{\textbf{Visual}} & \multicolumn{2}{c|}{\textbf{Audio}} & \multicolumn{2}{c}{\textbf{Visual}} & \multicolumn{2}{c|}{\textbf{Audio}} & \multicolumn{2}{c}{\textbf{Image}} & \multicolumn{2}{c}{\textbf{Text}} \\ \hline
w/o KD          & \multicolumn{2}{c}{58.1{\tiny±0.33}}       & \multicolumn{2}{c|}{56.3{\tiny±0.22}}      & \multicolumn{2}{c}{31.6{\tiny±0.18}}       & \multicolumn{2}{c|}{52.8{\tiny±0.11}}      & \multicolumn{2}{c}{38.7{\tiny±0.16}}       & \multicolumn{2}{c|}{59.4{\tiny±0.16}}      & \multicolumn{2}{c}{66.7{\tiny±0.22}}      & \multicolumn{2}{c}{68.1{\tiny±0.21}}     \\
FitNet\cite{fitnet}          & \multicolumn{2}{c}{56.4{\tiny±0.47}}                & \multicolumn{2}{c|}{52.9{\tiny±0.32}}               & \multicolumn{2}{c}{29.6{\tiny±0.63}}                & \multicolumn{2}{c|}{48.0{\tiny±0.81}}               & \multicolumn{2}{c}{37.9{\tiny±0.39}}                & \multicolumn{2}{c|}{57.1{\tiny±0.79}}               & \multicolumn{2}{c}{-}              & \multicolumn{2}{c}{-}             \\
Review\cite{review}          & \multicolumn{2}{c}{59.6{\tiny±0.45}}                & \multicolumn{2}{c|}{55.7{\tiny±0.36}}               & \multicolumn{2}{c}{32.1{\tiny±0.63}}                & \multicolumn{2}{c|}{50.6{\tiny±0.31}}               & \multicolumn{2}{c}{38.2{\tiny±0.47}}                & \multicolumn{2}{c|}{57.9{\tiny±0.33}}               & \multicolumn{2}{c}{-}              & \multicolumn{2}{c}{-}             \\
KD\cite{kl}              & \multicolumn{2}{c}{57.4{\tiny±0.92}}       & \multicolumn{2}{c|}{53.4{\tiny±0.85}}      & \multicolumn{2}{c}{32.3{\tiny±0.35}}       & \multicolumn{2}{c|}{46.6{\tiny±0.24}}      & \multicolumn{2}{c}{38.5{\tiny±0.50}}       & \multicolumn{2}{c|}{56.3{\tiny±0.46}}      & \multicolumn{2}{c}{66.3{\tiny±0.24}}      & \multicolumn{2}{c}{68.4{\tiny±0.12}}     \\
DML\cite{dml}             & \multicolumn{2}{c}{60.3{\tiny±1.60}}       & \multicolumn{2}{c|}{56.4{\tiny±0.55}}      & \multicolumn{2}{c}{31.8{\tiny±0.41}}       & \multicolumn{2}{c|}{48.0{\tiny±1.31}}      & \multicolumn{2}{c}{38.7{\tiny±0.86}}       & \multicolumn{2}{c|}{58.2{\tiny±1.01}}      & \multicolumn{2}{c}{67.9{\tiny±0.18}}               & \multicolumn{2}{c}{69.6{\tiny±0.24}}              \\

SHAKE\cite{shake}           & \multicolumn{2}{c}{60.0{\tiny±0.35}}       & \multicolumn{2}{c|}{ \underline{58.6}{\tiny±0.61}}      & \multicolumn{2}{c}{32.2{\tiny±0.59}}       & \multicolumn{2}{c|}{ 47.3{\tiny±0.72}}      & \multicolumn{2}{c}{38.3{\tiny±0.41}}       & \multicolumn{2}{c|}{ \underline{59.5}{\tiny±0.34}}      & \multicolumn{2}{c}{68.1{\tiny±0.16}}               & \multicolumn{2}{c}{ \underline{69.7}{\tiny±0.26}}              \\
RKD\cite{rkd}             & \multicolumn{2}{c}{48.3{\tiny±0.68}}       & \multicolumn{2}{c|}{51.9{\tiny±1.36}}      & \multicolumn{2}{c}{28.2{\tiny±0.71}}       & \multicolumn{2}{c|}{44.5{\tiny±0.73}}     & \multicolumn{2}{c}{33.4{\tiny±0.49}}       & \multicolumn{2}{c|}{41.5{\tiny±1.36}}      & \multicolumn{2}{c}{67.0{\tiny±0.23}}               & \multicolumn{2}{c}{67.4{\tiny±0.21}}              \\
DKD \cite{dkd}            & \multicolumn{2}{c}{60.4{\tiny±0.82}}       & \multicolumn{2}{c|}{55.1{\tiny±0.65}}      & \multicolumn{2}{c}{32.6{\tiny±0.65}}       & \multicolumn{2}{c|}{48.6{\tiny±1.02}}      & \multicolumn{2}{c}{38.1{\tiny±0.43}}       & \multicolumn{2}{c|}{57.2{\tiny±0.86}}      & \multicolumn{2}{c}{68.0{\tiny±0.17}}               & \multicolumn{2}{c}{69.2{\tiny±0.23}}              \\
DIST \cite{dist}           & \multicolumn{2}{c}{ \underline{61.1}{\tiny±1.82}}       & \multicolumn{2}{c|}{57.9{\tiny±0.57}}      & \multicolumn{2}{c}{29.8{\tiny±0.61}}       & \multicolumn{2}{c|}{49.3{\tiny±0.29}}      & \multicolumn{2}{c}{38.5{\tiny±0.39}}       & \multicolumn{2}{c|}{58.9{\tiny±0.45}}      & \multicolumn{2}{c}{ \underline{68.3}{\tiny±0.21}}      & \multicolumn{2}{c}{67.8{\tiny±0.18}}     \\
NKD \cite{normkd}         & \multicolumn{2}{c}{60.6{\tiny±0.64}}       & \multicolumn{2}{c|}{56.1{\tiny±0.68}}      & \multicolumn{2}{c}{ \underline{32.9}{\tiny±0.32}}       & \multicolumn{2}{c|}{\underline{52.2}{\tiny±0.62}}      & \multicolumn{2}{c}{ \underline{39.2}{\tiny±0.52}}       & \multicolumn{2}{c|}{59.3{\tiny±0.40}}      & \multicolumn{2}{c}{67.2{\tiny±0.26}}               & \multicolumn{2}{c}{68.5{\tiny±0.16}}          \\ \hline
\textbf{Ours}$^{\dagger}$       & \multicolumn{2}{c}{62.4{\tiny±0.24}}       & \multicolumn{2}{c|}{60.5{\tiny±0.37}}     & \multicolumn{2}{c}{34.2{\tiny±0.28}}       & \multicolumn{2}{c|}{54.5{\tiny±0.22}}      & \multicolumn{2}{c}{40.8{\tiny±0.23}}       & \multicolumn{2}{c|}{61.6{\tiny±0.34}}      & \multicolumn{2}{c}{68.2{\tiny±0.09}}      & \multicolumn{2}{c}{69.8{\tiny±0.16}}     \\
\textbf{Ours}$^{\ddagger }$            & \multicolumn{2}{c}{\textbf{62.8}{\tiny±0.28}}       & \multicolumn{2}{c|}{\textbf{61.4}{\tiny±0.44}}     & \multicolumn{2}{c}{\textbf{34.7}{\tiny±0.23}}       & \multicolumn{2}{c|}{\textbf{54.9}{\tiny±0.16}}      & \multicolumn{2}{c}{\textbf{40.9}{\tiny±0.31}}       & \multicolumn{2}{c|}{\textbf{61.9}{\tiny±0.27}}      & \multicolumn{2}{c}{\textbf{68.8}{\tiny±0.15}}      & \multicolumn{2}{c}{\textbf{70.1}{\tiny±0.12}}     \\ \hline
\end{tabular}
\caption{\textbf{Comparison results on Visual-Audio and Image-Text datasets.} The metric is the top-1 accuracy ($\%$). Ours$^{\ddagger }$ means fully updating the teacher model, and Ours$^{\dagger}$ means partially finetuning the top 2 layers. The best is in \textbf{bold}, and the second is \underline{underlined}.}
\label{tab3}
\end{table*}

\textbf{Implementation.} For \textit{visual-audio} datasets, the preprocess strategy follows \cite{on_the_fly_multimodal, pmr}.
Concretely, for audio modality, we change the input channel from 3 to 1 as \cite{vggsound}. Audio data is transformed into a spectrogram of size 257×299 for CREMA-D, 257×1,004 for AVE, and 257×1,004 for VGGsound, respectively, with the window length of 512 and overlap of 353. For visual modality, the input channel is adjusted considering input frames \cite{sound_pixel}. Concretely, 3 frames are uniformly sampled from VGGsound, and 1 frame is extracted from AVE and CREMA-D. 
Standard augmentations are employed, including random cropping and flipping.
We train the network for 100 epochs with 1e-2 initial learning rate and decay follow the `poly' policy with the power of 0.9. We use SGD with 0.9 momentum and default hyperparameters as the optimizer. 
For the \textit{image-text} dataset, we use the same training strategies and adopt $\omega=0$ across all experiments. Here, following \cite{on_the_fly_multimodal, mcce}, we adopt the same ResNet-18 \cite{resnet} as the backbone for visual and audio modality, and BERT-base \cite{bert} for text and MobileNetV2 \cite{mobile} and image feature extractors, respectively.
All results are the average of three different seeds.
% More experiments on diverse-capacities homogeneous and heterogeneous architectures are also compared.

\textbf{Comparison Results.} 
In Table \ref{tab3}, we compare our method to some advanced KD methods with the same training settings. 
We follow \cite{on_the_fly_multimodal, pmr} and give the detailed preprocess strategy. For audio modality, we change the input channel from 3 to 1 as \cite{vggsound}. Audio data is transformed into a spectrogram of size 257×299 for CREMA-D, 257×1,004 for AVE, and 257×1,004 for VGGsound, respectively, with the window length of 512 and overlap of 353. For visual modality, the input channel is adjusted considering input frames \cite{sound_pixel}. Concretely, 3 frames are uniformly sampled from VGGsound, and 1 frame is extracted from AVE and CREMA-D. Standard augmentations are employed, including random cropping and flipping.
We initialize weights of the student model and proxies following \cite{relu}. All experiments are conducted with NVIDIA RTX3090 GPUs on CUDA 11.4 using the PyTorch framework. All results are the average of three different seeds, which are set to 1, 2, and 3, respectively.
We imply traditional unimodal knowledge distillation with their defaulted settings. Previous logits-based KD methods can be seamlessly applied to the Cross-Modal Knowledge Distillation (CMKD) task. Due to the different spatial dimensions of multimodal inputs, the intermediate features have different spatial dimensions. Feature-based KD methods cannot be directly applied to CMKD. To deal with this issue, we employ the bilinear interpolation operator to align the intermediate features of teacher and student. Besides, BERT has 12 layers while MobileNetV2 has 5 layers. We do not conduct feature-based KD on the CrisisMMD dataset for comparisons because we can not choose which layers to be distilled based on their original implementations.

We can learn from Table \ref{tab3} that our proposed method, C$^2$KD, consistently outperforms other KD methods across four datasets. Existing KD methods can not effectively distill one modality information to another modality, especially for the datasets with the significant modality imbalance issue like AVE and VGGsound. Concretely, feature-based KD (FitNet\cite{fitnet}, Review\cite{review}) methods fail in CMKD because of significant feature divergence (see Section \ref{discussion}). Online KD (DML \cite{dml} and SHAKE \cite{shake}) methods update teacher models and achieve better cross-modal knowledge transfer ability, compared with the baseline \cite{kl}. Due to soft label misalignment between modalities, the relation-based method (RKD \cite{rkd}) degrades severely in CMKD. Recent advanced logits-based methods (DKD \cite{dkd}, DIST \cite{dist}, and NKD \cite{normkd}) significantly outperform the vanilla KL loss by proposing the relaxed KD functions and logits decoupling strategies. However, these methods fail to transfer cross-modal knowledge from low-accuracy to high-accuracy modality, impeding their practical deployments in CMKD.
% As the results summarized in Table \ref{tab3}.

\subsection{Multimodal Semantic Segmentation}

We also extend C$^2$KD to the multimodal semantic segmentation, a challenging dense prediction task. Concretely, following \cite{mfh}, we conduct experiments on the NYU-Depth V2 dataset \cite{nyu}. NYU-Depth V2 contains 1,449 aligned RGB and depth pairs with 40 category labels, of which 795 pairs are used for training, and 654 pairs are used for testing.  

\textbf{Implementation.} Both teacher and student networks deploy the DeepLab V3+ \cite{deeplab} architecture with diverse backbones.
% We perform experiments on homogeneous and heterogeneous
The training settings follow \cite{cirkd} that we adopt SGD as the optimizer with a
momentum of 0.9, a batch size of 16, an initial learning rate of 0.02, and ImageNet pre-trained weights. The total training iterations is 40K, decayed by the `poly' policy with the power of 0.9. Experiments on homogeneous/heterogeneous backbones, including ResNet-18/ResNet-18 and ResNet-18/MobileNetV2 pairs, are conducted to validate our method. All results are the average of three different seeds.

\textbf{Comparison Results.} 
We compare our methods with advanced traditional KD methods (KD \cite{kl}, SHAKE \cite{shake}, DIST \cite{dist}, and NKD \cite{normkd}) as well as the semantic segmentation KD method (CIRKD \cite{cirkd}). The results compared with previous methods are summarized in Table \ref{tab4}. We can see that previous KD methods do not perform well in CMKD, especially in transferring low-accuracy modality information to high-accuracy modality. Our method can significantly improve the distilled performance of arbitrary single modality. For instance, ours consistently surpasses the advanced CIRKD in transferring depth information to RGB modality. Besides, when replacing KL loss with DIST loss \cite{dist}, our method affords clear improvements.

\begin{figure*}[t] \centering
  \begin{minipage}[t]{0.5\textwidth}
    \centering
    \includegraphics[width=\textwidth]{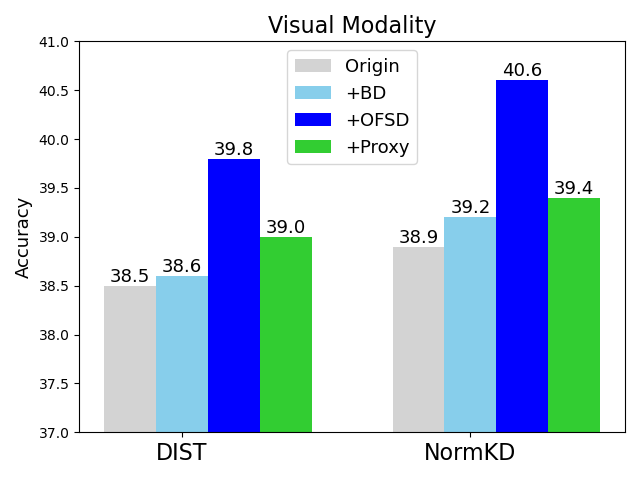}

    \caption*{(a) T: audio; S: visual}
  \end{minipage}\hfill
  \begin{minipage}[t]{0.5\textwidth}
    \centering
    \includegraphics[width=\textwidth]{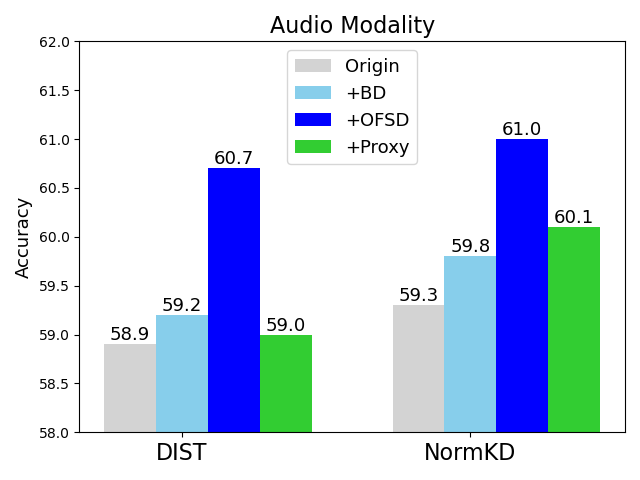}

    \caption*{(b) T: visual; S: audio}
  \end{minipage}

\caption{\textbf{Generalizability of each module.} We conduct experiments on VGGsound dataset in terms of DIST \cite{dist} and NKD \cite{normkd}.}

\label{comparison}
\end{figure*}

\begin{figure}
  \centering
  \includegraphics[width=1\textwidth]{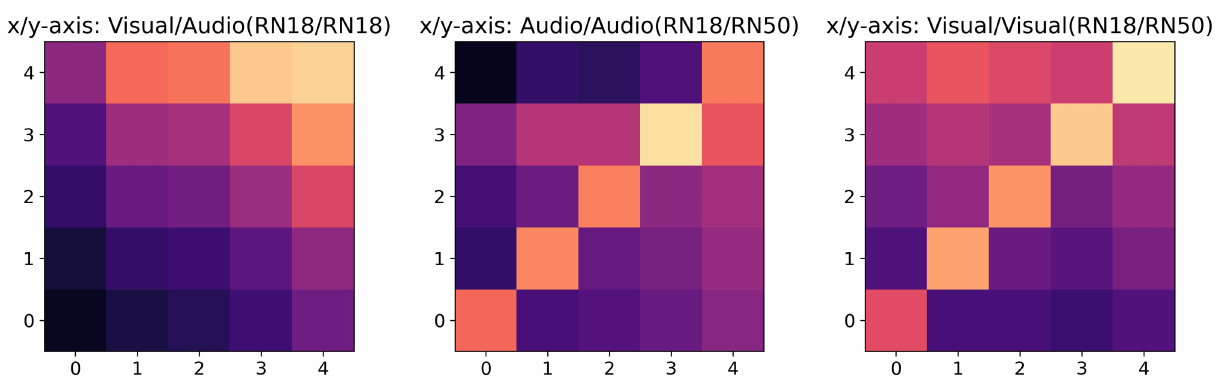}
  \caption{\textbf{The CKA score of intermediate features on AVE.}}
  \label{cka}
\end{figure}

\begin{table}[t] \small \centering
\begin{tabular}{c|cc|cc|cc}
\hline
\multirow{2}{*}{} & \textbf{RGB}      & \textbf{Depth}    & \textbf{RGB}      & \textbf{Depth}       & \textbf{RGB}         & \textbf{Depth}    \\
                  & RN18 & RN18 & RN18 & MNV2 & MNV2 & RN18 \\ \hline
w/o KD            & 36.1              & 30.5              & 36.1                  & 31.2                     &  36.3                    & 30.5                  \\\hdashline
KD \cite{kl}               &  35.8                 & 30.9                  &   36.2                &  31.9                    &  36.5                    & 31.8                  \\
SHAKE \cite{shake}            & 37.1                  &  31.2                 & \underline{37.0}   & 32.7                     & \underline{37.1}        & 32.9                  \\
DIST \cite{dist}             &  36.9                 & 32.0                  & 36.5                  & \underline{32.9}       &  36.8                    &  33.1                 \\
NKD \cite{normkd}           & 36.5                  &  30.8                 &  36.4                 & 32.2                     &   36.4                   &  32.7                 \\
CIRKD \cite{cirkd}          & \underline{37.3}  &  \underline{32.6}         & 36.9                  & 32.7                     & 36.7                     & \underline{33.4}        \\ \hline
\textbf{Ours}     & 37.5                  &32.5                   &37.2                   &  32.8                    &  37.4                    & 33.1                  \\
\textbf{Ours}+\cite{dist}    &   \textbf{38.1}      & \textbf{33.2}        & \textbf{37.7}       &  \textbf{33.5}        & \textbf{37.9}            &  \textbf{33.7}                 \\ \hline
\end{tabular}

\caption{\textbf{Comparison results on RGB-Depth semantic segmentation dataset.} The metric denotes the mean Intersection over Union (mIoU: $\%$). Ours+\cite{dist} means we replace KL loss with the
advanced DIST loss. RN18: ResNet-18; MNV2: MobileNetV2.}

\label{tab4}
\end{table}

\begin{table}[t] \centering

\begin{tabular}{c|cccc|cc|cc|cc}
\hline
    \multirow{2}{*}{\makebox[0.005\textwidth][c]{}} & \multirow{2}{*}{\makebox[0.04\textwidth][c]{\scriptsize\textbf{Proxy}}} & \multirow{2}{*}{\makebox[0.005\textwidth][c]{\scriptsize\textbf{OFS}}} &\multirow{2}{*}{\makebox[0.005\textwidth][c]{\scriptsize\textbf{NT}}} & \multirow{2}{*}{\makebox[0.005\textwidth][c]{\scriptsize\textbf{BD}}} & \multicolumn{2}{c|}{\makebox[0.2\textwidth][c]{\textbf{Proxy}}} & \multicolumn{2}{c|}{\textbf{VGGsound}} & \multicolumn{2}{c}{\textbf{CrisisMMD}} \\ \cline{6-11} 
                  &                                 &                               &                               &                                & \textbf{Visual}  & \textbf{Audio} & \textbf{Visual}    & \textbf{Audio}    & \textbf{Image}     & \textbf{Text}     \\ \hline
\textbf{(e)}                     &                                 &                               &                              &                                & 31.6             & 52.8           & 38.7               & 59.4              & 66.7               & 68.1              \\                  
\textbf{(a)}               &                                 &                               &                              &                                & 32.3             & 46.6           & 38.5               & 56.3              & 66.3               & 69.2              \\
\textbf{(b)}               &                                 &                               &                              & $\checkmark$                   & 32.7             & 47.9           & 38.8               & 57.6              & 67.3               & 69.2              \\
\textbf{(c)}               &                                 & $\checkmark$                  & $\checkmark$                 & $\checkmark$                   & 34.6             & 54.3           & 40.4               & 61.5              & 68.5               & 69.8              \\
-                          &                                  &                              &$\checkmark$                 & $\checkmark$                   & 33.2             & 52.9           & 39.4               & 60.3              & 67.9               & 68.9              \\
-                          &                                  & $\checkmark$                  &                             & $\checkmark$                   & 34.4             & 52.5           & 40.0               & 59.9              & 68.0               & 69.5              \\\hline
% -                 & $\checkmark$                              &                              &                              & $\checkmark$                   & 33.2             & 53.6           & 39.3               & 59.5              & 67.5               & 69.0              \\ \hline
\textbf{(d)}               & $\checkmark$                   & $\checkmark$                  & $\checkmark$                      & $\checkmark$                    & 34.7             & 54.9           & 40.9               & 61.9              & 68.8               & 70.1              \\ \hline
\end{tabular}

\caption{\textbf{Ablation studies on each module.} (a), (b), (c), and (d) represent the evolution steps of C$^2$KD (Figure \ref{framework}). (e) indicates the results without KD. The metric is the top-1 accuracy ($\%$).}

\label{tab5}
\end{table}

\section{Ablation and Sensitivity Analysis}
\label{c4.5}

\noindent \textbf{Effectiveness and generalizability of each module.}
We analyze how the proposed modules improve CMKD. Table \ref{tab5} reports the results of ablation studies on AVE, VGGsound, and CrisisMMD with the same backbones. The configurations of (a), (b), (c), and (d) correspond to the evolution steps shown in Figure \ref{framework}. Compared to the vanilla KD (i.e., (a)), the Bidirectional Distillation (BD) updates the teacher model (i.e., (b)) to mitigate the model gap. 
Furthermore, to validate the effectiveness of OFSD, we decouple OFSD into the On-the-Fly Selection (OFS) strategy and Non-Target (NT) classes distillation approach. We can learn that both OFS and NT benefit CMKD, and the combination of both brings significant improvement compared to (b).
% Considering modality imbalance and soft label misalignment, we devise OFSD and significantly improve the performances. The distilled modality consistently outperforms individual training, which validates the selected samples based on $\mathrm{KRC}$ bring receptive cross-modal knowledge for the student modality. 
The proxy teacher and student circumvent the direct imitation of cross-modal logits, serving as bridges for inheriting unimodal and cross-modal knowledge and facilitating the transfer of integrated knowledge through bidirectional distillation.  The progressive KD strategy further improves the CMKD results. 
The structure of proxies adheres to \cite{fitnet, shake}. 

To ascertain the generalizability of each component, we incorporate the proposed plug-and-play modules into advanced KD methods (i.e., DIST \cite{dist} and NKD \cite{normkd}). We can learn from Figure \ref{comparison} that the proposed modules consistently improve the performances of traditional KD methods in the CMKD task, especially for the OFSD strategy.

\begin{table}[t] \small \centering
\begin{tabular}{c|clcl|clcl|clcl}
                     & \multicolumn{4}{c|}{\textbf{AVE}}                                         & \multicolumn{4}{c|}{\textbf{VGGsound}}                                    & \multicolumn{4}{c}{\textbf{CrisisMMD}}                                 \\ \hline
\textbf{Method}      & \multicolumn{2}{c}{\textbf{Visual}} & \multicolumn{2}{c|}{\textbf{Audio}} & \multicolumn{2}{c}{\textbf{Visual}} & \multicolumn{2}{c|}{\textbf{Audio}} & \multicolumn{2}{c}{\textbf{Image}} & \multicolumn{2}{c}{\textbf{Text}} \\ \hline
w/o KD               & \multicolumn{2}{c}{31.6}            & \multicolumn{2}{c|}{52.8}           & \multicolumn{2}{c}{38.7}            & \multicolumn{2}{c|}{59.4}           & \multicolumn{2}{c}{66.1}           & \multicolumn{2}{c}{68.1}          \\ \hdashline
DLB\cite{dlb}                  & \multicolumn{2}{c}{32.6}            & \multicolumn{2}{c|}{53.3}           & \multicolumn{2}{c}{39.1}            & \multicolumn{2}{c|}{60.2}           & \multicolumn{2}{c}{66.9}           & \multicolumn{2}{c}{68.6}          \\
ZipfKD\cite{zipf}               & \multicolumn{2}{c}{33.3}            & \multicolumn{2}{c|}{53.5}           & \multicolumn{2}{c}{40.2}            & \multicolumn{2}{c|}{60.3}           & \multicolumn{2}{c}{67.4}           & \multicolumn{2}{c}{68.9}          \\
USKD\cite{normkd}                   & \multicolumn{2}{c}{33.1}            & \multicolumn{2}{c|}{53.2}           & \multicolumn{2}{c}{40.0}            & \multicolumn{2}{c|}{60.1}           & \multicolumn{2}{c}{67.1}           & \multicolumn{2}{c}{69.0}          \\ \hline
\textbf{Ours}        & \multicolumn{2}{c}{34.7}            & \multicolumn{2}{c|}{54.9}           & \multicolumn{2}{c}{40.9}            & \multicolumn{2}{c|}{61.9}           & \multicolumn{2}{c}{68.8}           & \multicolumn{2}{c}{70.1}          \\\hline
\end{tabular}
\caption{\textbf{Comparison results of different Self-KD methods.}}
\label{tab6}
\end{table}

\noindent \textbf{Necessity of cross-modal KD.}
Considering the challenges of cross-modal KD, a question may arise: Do we really need CMKD rather than fully explore self-knowledge? Self-knowledge distillation (Self-KD) techniques \cite{tfkd, mixkd, dlb, zipf, normkd} have been proposed to utilize the information within the student model to facilitate its learning process. Specially, DLB \cite{dlb} leverages the soft targets generated in the last mini-batch backup for training consistency and stability. ZipfKD \cite{zipf} and USKD \cite{normkd} generate soft labels following the Zipf’s law distribution \cite{law}. We conduct experiments on these advanced Self-KD methods to validate the necessity of CMKD. From Table \ref{tab6}, we can learn that although Self-KD improves the performance of each modality, our method consistently outperforms Self-KD methods by a clear margin. The results indicate the necessity of cross-modal KD.

\begin{figure}
  \centering
  \includegraphics[width=1\textwidth]{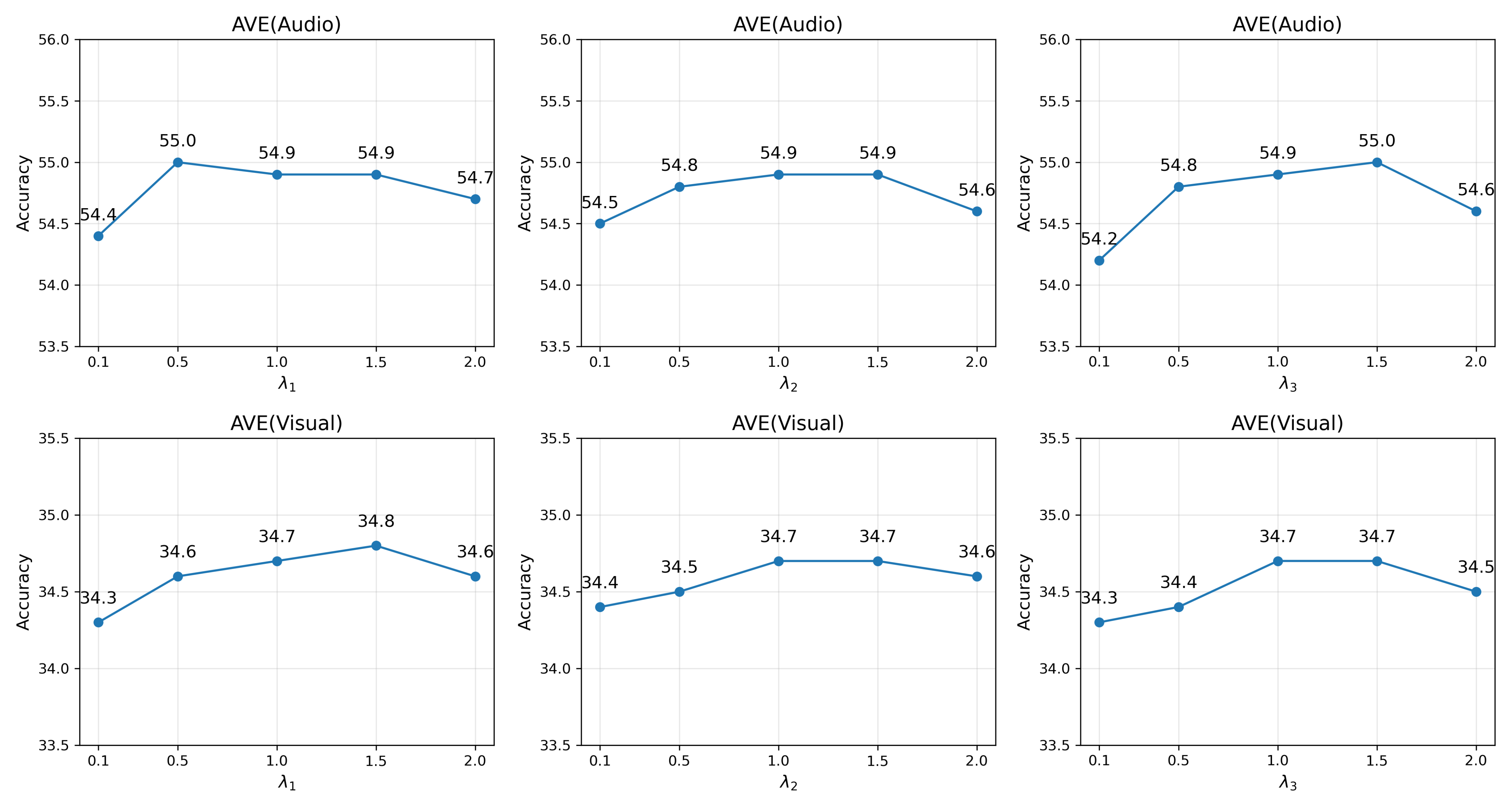}
  \caption{\textbf{Analysis of $\lambda_1$, $\lambda_2$, $\lambda_3$.} We conduct experiments on the AVE \cite{ave} dataset with ResNet-18 as the multimodal backbones.}
  \label{lambda}
\end{figure}

\noindent \textbf{Parameter sensitivity.}
Here, we conduct a sensitivity study on $\mathrm{KRC}$ threshold $\omega$. Results are in Figure \ref{omega}. 
% $\omega$ in Equation \ref{eq4} determines when to filter out rank-distorted samples. Intuitively, samples with more receptive cross-modal information should be preserved. 
% Larger $\omega$ filters out more samples, which might hinder corssmodal KD. 
% Small $\omega$ filters out more samples, which might hinder corssmodal KD. 
% Our method is sensitive to the hyper-parameters $\omega$ if it is set to the other value.
Large $\omega$ filters out more samples, which might hinder cross-modal knowledge transfer, while low $\omega$ preserves more samples, which might contain rank-distorted samples that induce adverse effects. We heuristically set $\omega$ to $0$ and achieve balanced results. Note that, as shown in Tables \ref{tab3} and \ref{tab5}, even the worst accuracy of varying $\omega$ is still competitive with the baselines, we think the studies show the necessity of on-the-fly filtering out rank-distorted samples based on $\mathrm{KRC}$. More analyses of $\lambda_{1}$, $\lambda_{2}$, and $\lambda_{3}$ are given in Figure \ref{lambda}. Our method is robust in terms of different hyperparameters. As our method can effectively transfer corssmodal information, large and small values of $\lambda$ could hinder knowledge transfer. Therefore, we adopt $\{ \lambda_{1}=\lambda_{2}=\lambda_{3}=1\}$ in all experiments.

\definecolor{whitegray}{RGB}{240,240,240}
\begin{table*}[] \tiny \centering
\renewcommand\arraystretch{0.95}
\begin{tabular}{c|cc|cc|cc|cc}
                            & \multicolumn{2}{c|}{\textbf{CREMA-D} \cite{crema-d}}                                                                                            & \multicolumn{2}{c|}{\textbf{AVE} \cite{ave}}                                                                                                & \multicolumn{2}{c|}{\textbf{VGGsound} \cite{vggsound}}                                                                                           & \multicolumn{2}{c}{\textbf{CrisisMMD} \cite{crisis}}                                                                                         \\ \cline{2-9} 
                            & \textbf{\begin{tabular}[c]{@{}c@{}}Visual\\ (A$\rightarrow$ V)\end{tabular}} & \textbf{\begin{tabular}[c]{@{}c@{}}Audio\\ (V$\rightarrow$ A)\end{tabular}} & \textbf{\begin{tabular}[c]{@{}c@{}}Visual\\ (A$\rightarrow$V)\end{tabular}} & \textbf{\begin{tabular}[c]{@{}c@{}}Audio\\ (V$\rightarrow$A)\end{tabular}} & \textbf{\begin{tabular}[c]{@{}c@{}}Visual\\ (A$\rightarrow$V)\end{tabular}} & \textbf{\begin{tabular}[c]{@{}c@{}}Audio\\ (V$\rightarrow$A)\end{tabular}} & \textbf{\begin{tabular}[c]{@{}c@{}}Image\\ (T$\rightarrow$I)\end{tabular}} & \textbf{\begin{tabular}[c]{@{}c@{}}Text\\ (I$\rightarrow$T)\end{tabular}} \\ \cline{2-9} 
                            & \textbf{RN18}                                                   & \textbf{RN50}                                                  & \textbf{RN18}                                                   & \textbf{RN50}                                                  & \textbf{RN18}                                                   & \textbf{RN50}                                                  & \textbf{BERT}                                                  & \textbf{SNV2}                                                 \\ \hline
\multicolumn{1}{l|}{w/o KD} & \multicolumn{1}{l}{58.1{\scriptsize±0.33}}                                   & \multicolumn{1}{l|}{57.9{\scriptsize±0.19}}                                 & \multicolumn{1}{l}{31.6{\scriptsize±0.18}}                                   & \multicolumn{1}{l|}{53.7{\scriptsize±0.16}}                                 & \multicolumn{1}{l}{38.7{\scriptsize±0.16}}                                   & \multicolumn{1}{l|}{60.1{\scriptsize±0.18}}                                 & \multicolumn{1}{l}{66.7{\scriptsize±0.22}}                                  & \multicolumn{1}{l}{68.0{\scriptsize±0.12}}                                 \\
KD \cite{kl}                         & 57.1{\scriptsize±0.57}                                                       & 54.1{\scriptsize±0.43}                                                      & 32.6{\scriptsize±0.62}                                                       & 48.5{\scriptsize±0.35}                                                      & 39.0{\scriptsize±0.46}                                                       & 57.8{\scriptsize±0.51}                                                      & 66.2{\scriptsize±0.38}                                                      & 68.4{\scriptsize±0.22}                                                     \\
Review \cite{review}                     & 59.4{\scriptsize±0.52}                                                       & 56.9{\scriptsize±0.62}                                                      & 32.0{\scriptsize±0.53}                                                       & 51.3{\scriptsize±0.57}                                                      & 38.5{\scriptsize±0.53}                                                       & 58.7{\scriptsize±0.60}                                                      & -                                                              & -                                                             \\
SHAKE \cite{shake}                      & 60.2{\scriptsize±0.36}                                                       & 58.9{\scriptsize±0.63}                                                      & 32.5{\scriptsize±0.67}                                                       & 48.6{\scriptsize±0.46}                                                      & 38.9{\scriptsize±0.51}                                                       & 59.9{\scriptsize±0.38}                                                      & 68.2{\scriptsize±0.23}                                                      & 69.6{\scriptsize±0.25}                                                     \\
NKD \cite{normkd}                        & 60.5{\scriptsize±0.62}                                                       & 56.9{\scriptsize±0.43}                                                      & 33.0{\scriptsize±0.36}                                                       & 52.5{\scriptsize±0.36}                                                      & 39.2{\scriptsize±0.67}                                                       & 59.6{\scriptsize±0.54}                                                      & 67.3{\scriptsize±0.31}                                                      & 68.6{\scriptsize±0.25}                                                     \\ \hline
\rowcolor{whitegray}Ours                        & 63.1{\scriptsize±0.25}                                                       & 62.1{\scriptsize±0.37}                                                      & 35.0{\scriptsize±0.21}                                                       & 55.3{\scriptsize±0.12}                                                      & 41.0{\scriptsize±0.22}                                                       & 62.0{\scriptsize±0.23}                                                      & 68.9{\scriptsize±0.12}                                                      & 70.0{\scriptsize±0.09}                                                     \\ \hline \hline
                            & \textbf{RN50}                                                   & \textbf{RN18}                                                  & \textbf{RN50}                                                   & \textbf{RN18}                                                  & \textbf{RN50}                                                   & \textbf{RN18}                                                  & \textbf{BERT}                                                  & \textbf{RN18}                                                 \\ \hline
\multicolumn{1}{l|}{w/o KD} & \multicolumn{1}{l}{59.7{\scriptsize±0.20}}                                   & \multicolumn{1}{l|}{56.3{\scriptsize±0.22}}                                 & \multicolumn{1}{l}{32.7{\scriptsize±0.25}}                                   & \multicolumn{1}{l|}{52.8{\scriptsize±0.11}}                                 & \multicolumn{1}{l}{39.3{\scriptsize±0.13}}                                   & \multicolumn{1}{l|}{59.4{\scriptsize±0.16}}                                 & \multicolumn{1}{l}{66.7{\scriptsize±0.22}}                                  & \multicolumn{1}{l}{68.1{\scriptsize±0.13}}                                 \\
KD \cite{kl}                         & 58.2{\scriptsize±0.53}                                                       & 54.0{\scriptsize±0.36}                                                      & 33.0{\scriptsize±0.43}                                                       & 46.9{\scriptsize±0.42}                                                      & 38.9{\scriptsize±0.52}                                                       & 56.4{\scriptsize±0.61}                                                      & 66.2{\scriptsize±0.42}                                                      & 68.5{\scriptsize±0.21}                                                     \\
Review \cite{review}                     & 60.4{\scriptsize±0.58}                                                       & 55.9{\scriptsize±0.39}                                                      & 32.7{\scriptsize±.0.56}                                                      & 51.2{\scriptsize±0.61}                                                      & 38.2{\scriptsize±0.43}                                                       & 58.1{\scriptsize±0.61}                                                      & -                                                              & -                                                             \\
SHAKE \cite{shake}                      & 60.5{\scriptsize±0.53}                                                       & 59.0{\scriptsize±0.48}                                                      & 33.4{\scriptsize±0.53}                                                       & 47.5{\scriptsize±0.43}                                                      & 38.6{\scriptsize±0.41}                                                       & 59.8{\scriptsize±0.49}                                                      & 68.0{\scriptsize±0.19}                                                      & 69.8{\scriptsize±0.23}                                                     \\
NKD \cite{normkd}                        & 60.9{\scriptsize±0.54}                                                       & 58.4{\scriptsize±0.62}                                                      & 33.2{\scriptsize±0.47}                                                       & 52.8{\scriptsize±0.55}                                                      & 39.5{\scriptsize±0.53}                                                       & 59.1{\scriptsize±0.46}                                                      & 67.4{\scriptsize±0.26}                                                      & 68.6{\scriptsize±0.22}                                                     \\ \hline
\rowcolor{whitegray}Ours                        & 63.5{\scriptsize±0.28}                                                       & 61.6{\scriptsize±0.23}                                                      & 35.5{\scriptsize±0.30}                                                       & 55.1{\scriptsize±0.22}                                                      & 41.3{\scriptsize±0.28}                                                       & 62.1{\scriptsize±0.24}                                                      & 68.8{\scriptsize±0.16}                                                      & 70.2{\scriptsize±0.16}                                                     \\ \hline
\end{tabular}
\caption{\textbf{Comparison results on Visual-Audio and Image-Text datasets.} The metric is the top-1 accuracy ($\%$). RN18: ResNet-18; RN50: ResNet-50; SNV2: ShuffleNet V2 \cite{snv2}.}
\label{tab1_appendix}
\end{table*}

\noindent \textbf{Different Backbones Evaluations.}
Besides the results in Tables \ref{tab3} and \ref{tab4}, we conduct more experiments to further demonstrate the effectiveness of our method across diverse-capacities homogeneous and heterogeneous architectures. We compare C$^2$KD with vanilla KD \cite{kl}, the state-of-the-art feature-based KD (Review \cite{review}), online KD (SHAKE \cite{shake}), logits-based KD (NKD \cite{normkd}). The results in Table \ref{tab1_appendix} illustrate C$^2$KD can effectively transfer crossmodal knowledge across diverse-capacities homogeneous architectures (i.e., ResNet-18-ResNet-50) and heterogeneous architectures (i.e., BERT-ResNet-18 and BERT-ShuffleNet V2).

\noindent\textbf{Proxy Analysis.}
\begin{table}[] \centering
\begin{tabular}{l|cc|cc}
\multicolumn{1}{c|}{} & \multicolumn{2}{c|}{\textbf{AVE} \cite{ave}} & \multicolumn{2}{c}{\textbf{VGGsound} \cite{vggsound}} \\ \hline
                      & \textbf{Visual}  & \textbf{Audio} & \textbf{Visual}    & \textbf{Audio}   \\
\textbf{Method}       & \textbf{(A$\rightarrow$V)}   & \textbf{(V$\rightarrow$A)} & \textbf{(A$\rightarrow$V)}     & \textbf{(V$\rightarrow$A)}   \\ \hline
w/o FA                                    & 34.4{\scriptsize±0.36}             & 53.0{\scriptsize±0.22}           & 40.2{\scriptsize±0.25}                   & 60.6{\scriptsize±0.24}                 \\
w/ CFA                                    & 34.7{\scriptsize±0.18}             & 55.0{\scriptsize±0.20}           & 40.8{\scriptsize±0.28}               & 62.0{\scriptsize±0.21}             \\ \hline
\rowcolor{whitegray}\textbf{Ours}         & 34.7{\scriptsize±0.23}             & 54.9{\scriptsize±0.16}           & 40.9{\scriptsize±0.31}               & 61.9{\scriptsize±0.27}             \\ \hline
\end{tabular}
\caption{\textbf{Analysis of the structure of the proxy.} We conduct experiments on the AVE \cite{ave} and VGGsound \cite{vggsound} datasets with ResNet-18 as the multimodal backbones.}
  \label{proxy}
\end{table}
We provide the detailed analysis of the student and teacher proxies. The proxy consists of the feature adaptation layer and the linear classification head, as shown in Equation 5. The feature adaptation layer follows the feature-based KD methods \cite{fitnet, review}, consisting of `Conv-BN-ReLU' block. Specifically, the kernel size of `Conv' is set to 1$\times$1, and input and output channel dimensions remain the same. Here, we analyse the structure of the proxy. We ablate the feature adaptation layer (w/o FA) and employ a complicated feature adaptation layer (`Conv-BN-Conv-BN-ReLU', i.e., w/ CFA). Table \ref{proxy} illustrates that without the feature adaptation layer (w/o FA), the linear classification head can not effectively transfer crossmodal information, possibly due to the degradation of nonlinear ability. However, the complicated feature adaptation layer does not bring obvious improvement. Therefore, the feature adaptation layer and linear classification head constitute the proxy.

\section{Discussion}
\label{discussion}
\noindent \textbf{Feature-based CMKD Analysis.}
We analyze the modality gap of CMKD in the \textit{logits-based} perspective and propose C$^2$KD to mitigate the issues. Furthermore, we provide the analysis of the challenge of CMKD from the \textit{feature-based} perspective. We adopt the Center Kernel Alignment (CKA) \cite{cka}, a feature similarity metric that measures input similarity with different dimensions. As shown in Figure \ref{cka}, compared to unimodal features, cross-modal features have significant feature divergence, and directly using feature-based distillation methods for CMKD is unreasonable. We leave the exploration of feature-based CMKD as the future work.

\begin{table}[] \centering
\scalebox{0.94}{
\begin{tabular}{l|ccccl|ccccl}
\multicolumn{1}{c|}{} & \multicolumn{5}{c|}{\textbf{AVE}}                                                                                      & \multicolumn{5}{c}{\textbf{VGGsound}}                                                                                 \\ \hline
\textbf{Method}       & \textbf{V}(S) & \multicolumn{1}{c|}{T}    & \textbf{A}(S) & \multicolumn{2}{c|}{T}    & \textbf{V}(S) & \multicolumn{1}{c|}{T}    & \textbf{A}(S) & \multicolumn{2}{c}{T}    \\ \hline
NKD \cite{normkd}                  & 32.9               & \multicolumn{1}{c|}{\textit{52.8}}    & 52.2              & \multicolumn{2}{c|}{\textit{31.6}}    & 39.2               & \multicolumn{1}{c|}{\textit{59.4}}    & 59.3              & \multicolumn{2}{c}{\textit{38.7}}    \\ \hdashline
+Cat                  & 34.0               & \multicolumn{1}{c|}{\textit{59.6}}    & 52.0              & \multicolumn{2}{c|}{\textit{60.2}}    & 39.9               & \multicolumn{1}{c|}{\textit{62.9}}    & 59.9              & \multicolumn{2}{c}{\textit{63.2}}    \\
+FiLM \cite{film}                & 33.2               & \multicolumn{1}{c|}{\textit{57.4}}    & 51.7              & \multicolumn{2}{c|}{\textit{57.6}}    & 39.0               & \multicolumn{1}{c|}{\textit{62.1}}    & 58.6              & \multicolumn{2}{c}{\textit{62.8}}    \\
+OGM  \cite{on_the_fly_multimodal} & 33.6               & \multicolumn{1}{c|}{\textit{60.9}}    & 52.7              & \multicolumn{2}{c|}{\textit{61.6}}    & 40.2               & \multicolumn{1}{c|}{\textit{65.2}}    & 59.8              & \multicolumn{2}{c}{\textit{64.3}}    \\ \hline
\textbf{Ours}         & 34.7               & \multicolumn{1}{c|}{\textit{54.2}}    & 54.9              & \multicolumn{2}{c|}{\textit{34.6}}    & 40.9               & \multicolumn{1}{c|}{\textit{59.0}}    & 61.9              & \multicolumn{2}{c}{\textit{41.6}}    \\ \hline
\end{tabular}}

\caption{\textbf{Comparison results of different multimodal teachers.} The \textit{italic} numbers mean teachers' accuracy. S: student; T: teacher.}

\label{mt}
\end{table}

\noindent \textbf{Multimodal Teacher Efficacy Analysis.} We provide analysis of the efficacy of multimodal teacher in CMKD. 
Concretely, the multimodal teacher is formulated by fusing the teacher and the student modalities with the supervision loss.
Following the multimodal learning \cite{uni_multimodal, on_the_fly_multimodal, pmr}, we adopt the fusion strategies including Concatenation (Cat), FiLM \cite{film}, and OGM \cite{on_the_fly_multimodal}. Pretrained teachers are updated for the better information fusion. Table \ref{mt} indicates multimodal teachers generate high-accuracy soft labels, while don't necessarily improve the distilled modality, especially for the high-accuracy modality (i.e., audio). Our customized teacher integrates receptive corssmodal information and ensures effective knowledge transfer. Inspired by \cite{mfh}, developing the multimodal learning method that contains more modality-general decisive information is a possible solution. We leave this intriguing challenge to future work.

\section{Chapter Summary}

% This chapter contribution targets the problem of modality imbalance and misalignment in cross-modal knowledge distillation (CMKD), which often degrades performance when modalities are missing during inference. we conduct thorough investigation toward the efficacy of CMKD and reveal that the modality imbalance and soft label misalignment induced by the inter-modality gap are the main factors for the failure of CMKD. Based on our analyses, we propose a simple yet effective method, \textbf{\underline{C}}ustomized \textbf{\underline{C}}rossmodal \textbf{\underline{K}}nowledge \textbf{\underline{D}}istillation (C$^2$KD). Specifically, we propose On-the-Fly Sample Selection (OFSD) strategy to filter out rank-distorted samples based on the $\mathrm{KRC}$ metric and distill knowledge from non-target classes. Meanwhile, the pre-trained teacher conducts bidirectional distillation with the student. Proxy student and teacher, inheriting unimodal and cross-modal knowledge, progressively transfer cross-modal knowledge. Extensive experiments demonstrate the effectiveness of our method.

Chapter \ref{Chapter4} contribution targets the problem of modality imbalance and misalignment in cross-modal knowledge distillation (CMKD), which often degrades performance when modalities are missing during inference. 
This chapter reveals that modality imbalance and soft label misalignment in cross-modal knowledge distillation (CMKD) are critical bottlenecks that degrade performance when modalities are missing during inference. Through systematic analysis, we identify the inter-modality gap—divergent feature distributions between modalities (e.g., text vs. vision) as the root cause of distorted knowledge transfer. To address this, we propose Customized Crossmodal Knowledge Distillation (C$^2$KD), specifically, we propose On-the-Fly Sample Selection (OFSD) strategy to filter out rank-distorted samples based on the $\mathrm{KRC}$ metric and distill knowledge from non-target classes. Meanwhile, the pre-trained teacher conducts bidirectional distillation with the student. Proxy student and teacher, inheriting unimodal and cross-modal knowledge, progressively transfer cross-modal knowledge. Extensive experiments demonstrate the effectiveness of our method. By resolving knowledge misalignment through adaptive sample selection and proxy-guided distillation, C$^2$KD advances the thesis’s theme of robust multimodal integration, extending Chapter \ref{Chapter3}’s compatibility principles to cross-modal distillation while setting the stage for Chapter \ref{Chapter5}’s self-introspective decoding to tackle hallucination in the multimodal inference.

\chapter{\textsc{SID}: Self-Introspective Decoding for Modality Prior to Alleviate Hallucinations for Multimodal Large Models}

\label{Chapter5}

\section{Challenges and Motivations}
\label{c5.1}

Chapters \ref{Chapter3} and \ref{Chapter4} demonstrate how to improve the robustness abilities of multimodal learning in terms of composition ability and inference with missing modalities situations. In the past few years, large-scale models pre-trained on massive amounts of data and then applied on downstream tasks have become the new paradigm, illustrating the robust and flexible solutions to diverse tasks. Concretely, recent advancements in Large Language Models (LLMs) \cite{llama, qwen, vicuna, llama2, llama3} have demonstrated great success over the past few years. Many efforts have been made to extend LLMs to Multimodal Large Language Models (MLLMs), especially Large Vision-Language Models (LVLMs) \cite{mplug, mimic, qwenvl, blip2, instructblip, llava1.5, fuyu, yi, llavanext}, achieving impressive performance across various vision tasks \cite{llavamed, gpt4roi} as well as more complex tasks like content comprehension \cite{ lisa} and generation \cite{instructdiffusion}.

Despite their extraordinary versatility, LVLMs face a significant challenge known as the `hallucination'. Concretely, hallucinated texts are fluent and semantically coherent but contain incorrect or non-existent statements about the given image, e.g., generating irrelevant or meaningless responses, identifying inaccurate colors, numbers, and locations of objects not present in the image \cite{opera}. 
This flaw in LVLMs poses a significant risk for real-world applications to become trustworthy AI assistants. For instance, in model-assisted computer-aided diagnosis scenarios \cite{chatcad}, such misinterpretation of medical images could lead to serious medical accidents.

One mainstream approach to alleviating hallucinations in LVLMs involves developing training-free decoding strategies known as Contrastive Decoding (CD) \cite{vcd, vig, icd, id}, which adjusts the next-token logits in a contrastive manner. Concretely, Vision CD (VCD) manipulates vision inputs with Gaussian noise \cite{vcd} or directly ablates visual inputs \cite{vig} to amplify language modality priors. Instruction CD (ICD) \cite{icd, id} designs negative prompt.\footnote{negative prompts like \texttt{`You are a confused object detector.'} and \texttt{`Always respond with the opposite of what you're asked.'} for different tasks.} 
The rationale is that disturbed inputs significantly exacerbate hallucinations, and CD subtracts hallucinated concepts from the original distribution to mitigate hallucinations.

\begin{figure} \centering
  \centering
  \includegraphics[width=1\textwidth]{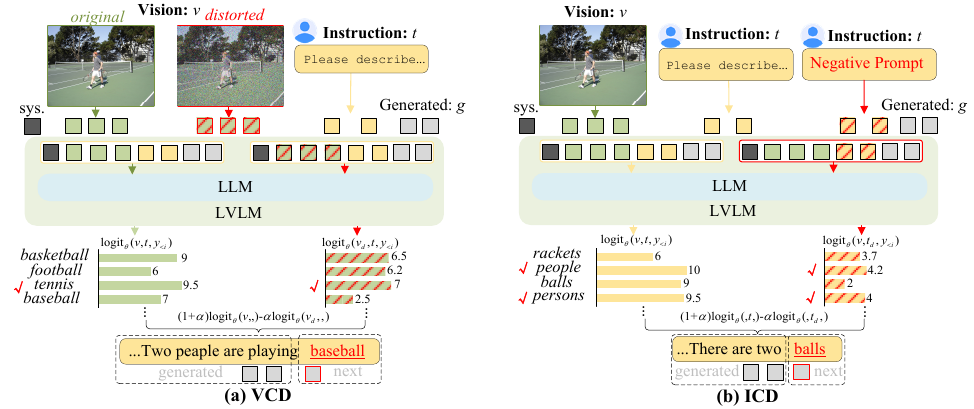}
  \caption{\textbf{Contrastive Decoding strategies:} (a) Visual Contrastive Decoding (VCD) \cite{vcd}  \textbf{manually} distort vision inputs. (b) Instruction Contrastive Decoding (ICD) \cite{icd, id} also \textbf{manually} design noisy instruction (negative prompt). We ablate other modules like the vision encoder and tokenizer for clarity. \textit{t}: \texttt{`Please describe this image in detail.'}; sys.: system prompt. $g$: generated text tokens.}
  \label{fig1}
\end{figure}

\textit{However}, input disturbances require elaborate designs for various downstream tasks, and the inference cost is inevitably doubled. \textit{Moreover}, the contrastive distributions are \textit{vision-and-text agnostic}, not necessarily amplify desired hallucinations but sometimes induce potential uncertainty noise for CD. Intuitive examples are illustrated in Figure  \ref{fig1}, and detailed analyses are in Sec. \ref{sec3.2}. 
In Figure  \ref{fig1} (a) and (b), LVLMs directly infer the correct next token from multimodal inputs. For Vision CD, distorted vision input exacerbates hallucinated object logits such as \textit{football} and \textit{basketball}, while the holistic noise suppresses \textit{baseball} to a low logit value. Consequently, VCD might compromise normal decoding. Similarly, for Instruction CD, LVLMs tend to refuse to answer negative prompts in open-end generation task (as seen in Figure  \ref{fig_example1} and \ref{fig_example2}), and also suffer from potential uncertainty noise similar to VCD.

To address the aforementioned issues, we propose a novel decoding strategy called \textit{Self-Introspective Decoding} (SID). 
Our empirical investigations reveal that pre-trained LVLMs can introspectively assess the importance of vision tokens adaptively, based on preceding vision and text (both instruction and generated) tokens. SID leverages this capability to amplify and then subtract \textit{vision-and-text association} hallucinations by proposing token-level disturbances named Context and Text-aware Token Selection (CT$^2$S) strategy. This strategy induces multimodal contextual hallucinations, rather than aimless ones, by conducting token selection in the early decoder layers.

In summary, this chapter's main contributions are three-fold:

{\footnotesize \textbullet\ } We re-think CD methods in LVLMs and attribute their failure cases to vision-and-text agnostic input distributions that induce potential uncertainty noise.

{\footnotesize \textbullet\ } To address this, we propose Self-Introspective Decoding (SID), where the CT$^2$S strategy adaptively amplifies and then subtracts vision-and-text association hallucinations. This approach is grounded in our investigations that pre-trained LVLMs can introspectively assess visual importance informed by preceding tokens.

{\footnotesize \textbullet\ } Through comprehensive comparisons, we demonstrate that SID generates high-quality texts with fewer hallucinations. Additionally, SID significantly reduces inference cost of contrastive decoding.

\section{Preliminary and Discussions}\label{Problem}
% \vspace{-0.4cm}
In the following, this sub-chapter first illustrates the generation paradigm of LVLMs to facilitate the understanding of SID. We then re-think the contrastive decoding in LVLMs and propose our motivation for SID. 
% \vspace{-0.4cm}

\subsection{Paradigm of LVLMs Generation}\label{sec3.1}
\textbf{Vision and Language Inputs.}
The inputs of LVLMs consist of both image ($v$) and text ($t$). Generally, the raw images are commonly fed to the visual encoder, and then the cross-model projection module maps vision information into LLMs’ input space, which is denoted as vision tokens $v=\left \{ v_{1}, v_{2}...v_{n} \right \} $ ($n$ is the length of vision tokens). Similarily, text is processed by tokenizer and embedding modules, which is denoted as text tokens $t=\left \{ t_{1}, t_{2}...t_{m} \right \} $ ($m$ is length of text tokens). Then, the image ($v$) and text ($t$) tokens are concatenated as the final input of LLMs.

\noindent \textbf{LVLMs Forward.}
The backbone networks of LVLMs are pre-trained LLMs like Vicuna \cite{vicuna} and LLaMA 2 \cite{llama2}, parameterized by $\theta$. Given multimodal tokens $\left \{  v , t\right \}$, LVLMs predict the next token probability ($y_{i}$) at $i$ time step in an auto-regressive manner following the methodology of LLMs, over the vocabulary set $\nu$: 
\begin{equation}
p(y_{i}|v,t,y_{<i})=\mathrm{softmax}(logit_{\theta }(y_{i}|v,t,y_{<i})),  y_{i} \in \nu
\label{eq1_}
\end{equation}

\noindent \textbf{Next Token Decoding.} After obtaining the next token probability $p(y_{i}|v,t,y_{<i})$, different decoding strategies are proposed to predict next token. 
The decoded token is concatenated to the last original input token, for the next round of generation until the end of the generation process.

\subsection{Re-thinking Contrastive Decoding in LVLMs}\label{sec3.2}

\begin{table}[] \centering \small
\caption{\textbf{Efficacy Analyses on CD strategies} on MSCOCO dataset. The \textit{Random} setting means objects
absent from the image are chosen randomly, while the \textit{Adversarial} setting prioritizes co-occurring objects which are not present in the image.}
\begin{tabular}{cl|cccc}
\hline
\multicolumn{2}{c}{}                                 & \multicolumn{2}{|c}{\textbf{Greedy}} & \multicolumn{2}{c}{\textbf{Sampling}} \\ 
\multicolumn{1}{l}{\textbf{Setting}} & \textbf{Method} & Accuracy $\uparrow$  & F1 Score $\uparrow$  & Accuracy $\uparrow$  & F1 Score $\uparrow$ \\ \hline
\multirow{7}{*}{\textit{Random}}     & Normal            & 88.8{\tiny±0.05} &  88.6{\tiny±0.08}   & 84.9{\tiny±0.03} & 83.2{\tiny±0.01}\\ \cdashline{2-6}
                                     & VCD               & 87.8{\tiny±0.02} &  87.9{\tiny±0.06}   &87.73 & 83.28        \\
                                     & w/o Eq. \ref{c5_eq3} &   -              &  -                 & 83.3{\tiny±0.04}  &82.2{\tiny±0.02}\\ \cdashline{2-6}
                                     & ICD               & 87.9{\tiny±0.04} &  88.1{\tiny±0.02}    &  86.9{\tiny±0.03}  &  85.2{\tiny±0.04}        \\
                                     &w/o Eq. \ref{c5_eq3} &   -              &  -                 & {82.7\tiny±0.02} & {81.8\tiny±0.03} \\\cdashline{2-6}
                                     & \textbf{Ours}     & \textbf{89.3}{\tiny±0.08} & \textbf{89.5}{\tiny±0.02}  & \textbf{88.8}{\tiny±0.03} & \textbf{88.7}{\tiny±0.02} \\
                                      & w/o Eq. \ref{c5_eq3}     &   -              &  -           & 87.2{\tiny±0.01} & 88.0{\tiny±0.02}  \\ \hline
\multirow{7}{*}{\textit{Adversarial}}  & Normal            & {79.3\tiny±0.05} & 80.9{\tiny±0.09}  &78.7{\tiny±0.03} &78.9 {\tiny±0.02} \\\cdashline{2-6}
                                     & VCD               & {80.9\tiny±0.06} & {81.0\tiny±0.04}  & 80.88 & 81.33        \\
                                     &w/o Eq. \ref{c5_eq3}   &   -              &  -     &{76.2\tiny±0.04} & {76.0\tiny±0.04}\\\cdashline{2-6}
                                     & ICD               & {80.2\tiny±0.03} &  {81.3\tiny±0.01}         &79.1{\tiny±0.02} & 80.4{\tiny±0.04}        \\
                                     & w/o Eq. \ref{c5_eq3}  &   -              &  -            &75.4{\tiny±0.02} & 76.4{\tiny±0.04}        \\\cdashline{2-6}
                                     &\textbf{Ours}      & \textbf{83.3}{\tiny±0.07} & \textbf{82.5}{\tiny±0.06}  & \textbf{82.6}{\tiny±0.05} & \textbf{82.1}{\tiny±0.06}\\
                                     &w/o Eq. \ref{c5_eq3}   &   -              &  -     & 82.2{\tiny±0.03} & 81.9{\tiny±0.01}\\\hline
\end{tabular}
\label{c5_table1}
\end{table}

Following the seminal works \cite{contrastive} in natural language processing, which introduced the Contrastive Decoding (CD) mechanism to enhance coherence and informativeness by considering the differences between expert and amateur models, various studies have adapted this strategy to LVLMs by distorting the visual or instruction inputs for contrastive purposes.
As the vision and instruction contrastive processes are symmetrical, we use visual contrastive decoding as an example. The contrastive decoded probability of next-token ($p_{cd}$) can be generally formulated as follows:
\begin{equation}
p_{cd}(y_{i}|v, v_{d}, t, y_{<i})=\mathrm{softmax}[(1+\alpha)logit_{\theta }(y_{i}|v,t,y_{<i}) - \alpha logit_{\theta  }(y_{i}|v_{d},t,y_{<i})]
\label{c5_eq2}
\end{equation}
where $d$ and $\alpha$ indicate distortion operation and hyperparameter, respectively. \textit{Generally}, CD methods employ an adaptive plausibility constraint to calibrate the entire output distribution, preventing implausible outputs from the augmented distribution \cite{contrastive, dola, vcd, vig, icd, id, ibd}:
\begin{equation}
\begin{split}
\nu _{token}(y_{<i})=\left\{  y_i \in \nu: p_{\theta}(y_{i}|v, t, y_{<i}) \ge \beta \max_{\omega }  p_{\theta}(\omega|v, t, y_{<i}) \right\},
\\
p_{cd}(y_{i}|v,v_{d},t, y_{<i}))=0, \;\mathrm{if}\; y_{i}\notin \nu _{token}(y_{<i})
\label{c5_eq3}
\end{split}
\end{equation}
where $\nu$ and $\nu_{token}$ are the output vocabulary and selected tokens.
$\beta$ controls the strength of truncation, with larger $\beta$ indicating more aggressive truncation that retains only high-probability tokens.

\begin{figure*} [] \centering
  \centering
  \includegraphics[width=1\textwidth]{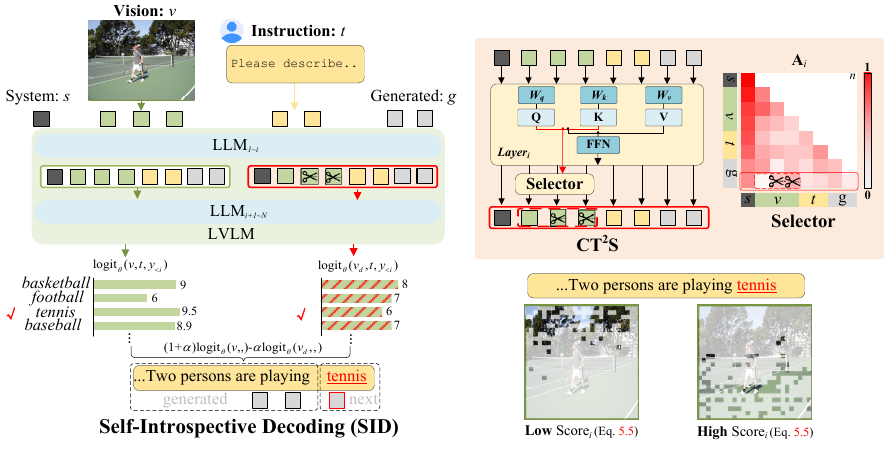}
  \caption{\textbf{Overview of Self-Introspective Decoding (SID).} CT$^2$S: Context and Text-aware Token Selection strategy. 
  LLaVA-1.5 7B is utilized as an example to visualize visual tokens with low and high scores (Equation \ref{c5_eq5}).
  }
  \label{fig2}
\end{figure*}

However, we argue that manually disturbing raw inputs might not trigger the desired hallucinations, while holistic disturbances will bring uncertainty noise that compromises the normal decoding. To validate our claim, we analyze the performances of normal decoding, VCD, and ICD using the POPE \cite{pope} metric, under both \textbf{sampling} and \textbf{greedy} decoding settings.
POPE quantitatively converts the hallucination evaluation into a binary classification problem by using the question format to prompt the model: \texttt{`Is there a <object> in the image?'}, with expected answers being \texttt{`Yes'} or \texttt{`No'}. 
From Table \ref{c5_table1}, 
under the \textbf{greedy} decoding setting, CD methods improve performance in the \textit{adversarial} setting, which are more challenging as they prioritize co-occurring confusing objects. CD methods achieve this by exacerbating and subtracting hallucinated concepts from the original distribution. However, in \textit{random} settings, where objects absent from the image are chosen randomly and are easily recognized, CD methods slightly underperform normal greedy decoding, which indicates that the correct token logit is somewhat compromised during contrastive decoding. In the \textbf{sampling} decoding setting, CD methods clearly outperform the normal sampling decoding.
However, CD methods rely on the adaptive plausibility constraint (Equation \ref{c5_eq3}) to filter out low-probability tokens. Without Equation \ref{c5_eq3}, CD methods are inferior to normal decoding in both \textit{random} and \textit{adversarial} settings, validating that vision-and-text agnostic input distributions induces potential uncertainty noise after Equation \ref{c5_eq2}. To address these issues, we propose a decoding strategy named \textit{Self-Introspective Decoding} (SID). SID \textit{adaptively} amplifies \textit{vision-and-text association} hallucinations informed by generated tokens to guide LVLMs in exploring factualness. Details are illustrated in the Sec. \ref{c5_sec4} and Figure  \ref{fig2}.

\begin{figure*} [ht]
\footnotesize
  \centering
  \includegraphics[width=1\textwidth]{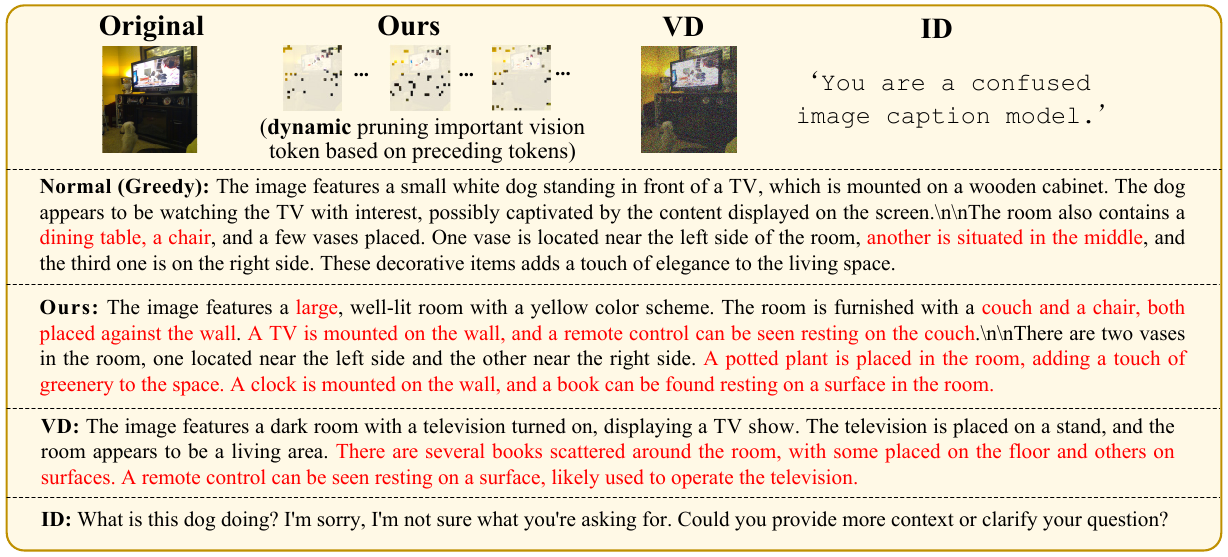}
  \caption{\textbf{Instance Illustration of Different Disturbance Results.} Examples are from MSCOCO inferred by LLaVA-1.5 7B with $i=3$ and Top-k=50. Hallucinations are marked in \textcolor{red}{red}.}
  \label{fig_example1}
\end{figure*}

\begin{figure*}[ht]
  \centering
  \includegraphics[width=1.0\textwidth]{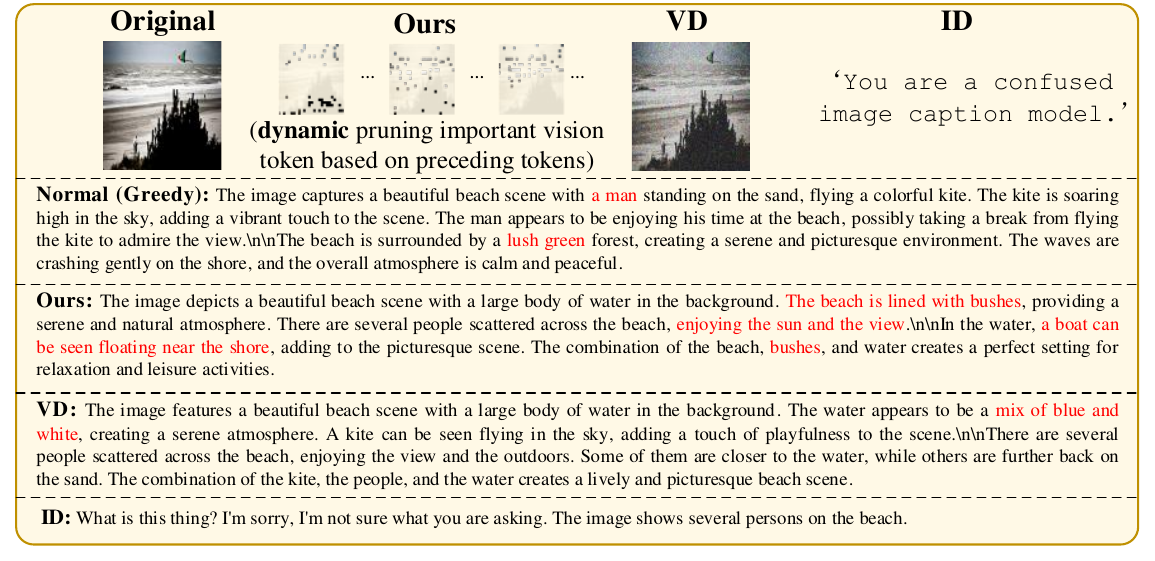}
\caption{\textbf{Instance Illustration of Different Disturbance Results.} Examples are from MSCOCO \cite{mscoco} inferred by LLaVA-1.5 7B with $i=3$ and Top-k=50. Hallucinations are marked in \textcolor{red}{red}.}
  \label{fig_example2}
\end{figure*}

\section{Self-Introspective Decoding (SID)}\label{c5_sec4}

\subsection{Understanding the Self-Introspective Pre-trained LVLMs} \label{sec4.1}
LLMs \cite{qwen, vicuna, llama2, llama3} have been scaled up to billions of paramters and pre-trained on trillions of tokens, endowing LLMs with encyclopedic ability like in-context learning \cite{icl}, zero \cite{zero-shot}/few-shot \cite{few-shot} ability. LVLMs extend LLMs to multimodal understanding capabilities by visual instruction tuning. Some works \cite{pumer, llava_pru, 0.5token} pointed out that vision information is redundant in LVLMs, and develop vision token reduction technologies to prune \cite{dynamicvit} and merge \cite{merging} tokens guided by importance metrics without further re-training. 
Regarding the hallucination issue, we argue that vision tokens with low attention scores induce \textbf{vision-and-text association hallucination}.
Formally, for the transformer block \cite{transformer} in the auto-regressive decoder \footnote{Here we illustrate the transformer block without KV Cache for better understanding.}, vision ($v$), text instruction ($t$), and generated tokens ($g$) are concatenated and projected into three distinct vectors: the query vector \textbf{Q}, the key vector \textbf{K}, and the value vector \textbf{V}, utilizing three linear transformations \textbf{$W_q$}, \textbf{$W_k$}, and \textbf{$W_v$}. The self-attention ($SA$) mechanism computes the relevance of each item to other items as follows:
\begin{equation}
\begin{split}
\mathbf{R}= SA(\mathbf{Q},\mathbf{K},\mathbf{V})=\mathbf{A}\cdot \mathbf{V},
\\
\mathbf{A}=\mathrm{softmax}(\frac{\mathbf{Q}\cdot \mathbf{K}^{T}}{\sqrt{d_{l}} } + M )
\label{c5_eq4}
\end{split}
\end{equation}
where $d_{l}$ represents the dimension of \textbf{Q}, \textbf{K}, \textbf{V}, $M$ represents the casual mask. $\mathbf{A} \in R^{(b, h, n, n)}$, where $b$, $h$, and $n$ denote batch size, number of key-value heads, and total token number, respectively. We denote the $\mathbf{A}_{i}$ as the attention matrix after \textit{Layer} $i$ of LVLMs. We then calculate vision token importance scores ($\mathrm{Score}_{i}(v)$) as shown in Figure  \ref{fig2} (\textbf{Selector}) based on $\mathbf{A}_{i}$:
\begin{equation}
\mathrm{Score}_{i}(v)  = \frac{1}{h}\sum_{j=1}^h\mathbf{A}_{i}^{(\cdot,j,\cdot,\cdot)}[-1]
\label{c5_eq5}
\end{equation}
where $v$ means vision token indexes. Contrary to token pruning/merging \cite{dynamicvit, merging} strategies, we preserve a certain number of the least important vision tokens based on Equation \ref{c5_eq5}.

\begin{figure*}[b]
  \centering
  \includegraphics[width=1\textwidth]{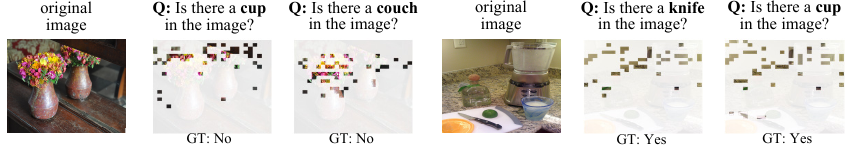}
  \caption{\textbf{Visualization Results} of the \textbf{least} important vision tokens on discrimination tasks informed by preceding vision and text tokens. LLaVA-1.5 7B with Layer $i=3$ is utilized.}
  \label{fig3}
\end{figure*}

\begin{figure*}
  \centering
  \includegraphics[width=1\textwidth]{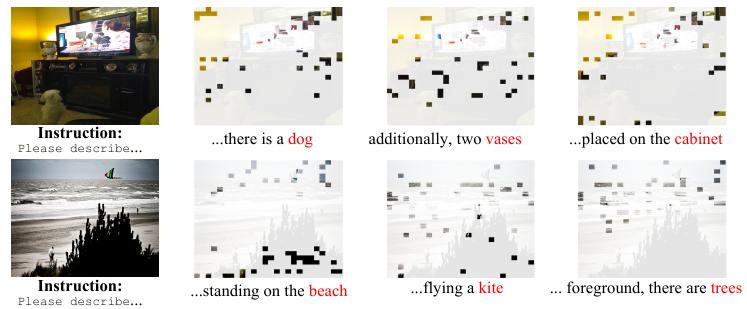}
  \caption{\textbf{Visualization Results of \textit{Adaptively} Selecting} the \textbf{least} important vision tokens on open-end generative tasks informed by preceding vision and text tokens. LLaVA-1.5 7B with Layer $i=3$ is utilized.}
  \label{fig4}
\end{figure*}

\noindent \textbf{Analyses.} Figure  \ref{fig3} and \ref{fig4} preliminarily \textcolor{black}{validate} the efficacy of $\mathrm{Score}_{i}(v)$ qualitatively. In Figure  \ref{fig3}, the preserved \textbf{least} important tokens mainly reflect areas opposite to the query. For instance, when querying \texttt{`cup'} in Figure  \ref{fig3} (left), LVLMs focus on \texttt{`cup'} in the foreground, thus preserving background tokens with low $\mathrm{Score}_{i}(v)$.
Conversely, LVLMs pay attention to background items when querying \texttt{`couch'}. When querying existing items in Figure  \ref{fig3} (right), vision tokens of unrelated regions are mainly preserved. For open-end generative tasks in Figure  \ref{fig4}, auto-regressive decoded tokens are generated based on preceding vision ($v$), instruction ($t$), and generated text ($g$) tokens. The preserved vision tokens are \textbf{adaptively adjusted} according to preceding tokens at each decoding step, primarily focusing on spurious related regions.
To quantitative analyses the attention score, we select the
top-100 and least-100 important vision tokens out of a total of 576 vision tokens of LLaVA-1.5 7B based on attention score based on Equation \ref{c5_eq5}. Visual and Instruction Disturbance (VD and ID) are also employed as inputs for analyses. Quantitative results in Table \ref{c5_table2} illustrate that 100 out of 576 vision tokens with high attention scores greatly maintain original ability, while low attention score tokens reach almost 50$\%$ accuracy for the binary classification problem, which indicates attention scores are a good indicator for
vision token importance. As for VD and ID, disturbance in raw input does not obviously harm the LVLMs' discrimination ability, as indicated by the POPE metric. However, VD and ID significantly compromise the open-end generation tasks reflected by the CHAIR metric (LVLMs tend to refuse to ID as shown in Figure \ref{fig_example1} and \ref{fig_example2}).
Above evaluations suggest that Equation \ref{c5_eq5} effectively assesses the importance of vision tokens.

We further demonstrate the \textit{open-end generated hallucinations} induced by ours, Vision Disturbance (VD) \citep{vcd}, and Instruction Disturbance (ID) \citep{icd} in Figure \ref{fig_example1} and \ref{fig_example2}. The hallucinations we amplified are more vision-and-text association compared to VD, while LVLMs usually refuse to response to ID. 
Additionally, we demonstrate the quantitative results for discrimination and generation tasks with \textit{VD and ID as inputs} in Table \ref{c5_table2}. Interestingly, VD and ID do not degrades much especially in discrimination tasks. Experiments imply that disturbed target logits still have the highest probability in most cases, and therefore, contrastive decoded target logits are not enhanced much after Equation \ref{eq2}, while CD methods are susceptible to potential uncertainty noise.

\begin{table}[] 
\centering \small
\caption{\textbf{Efficacy Analyses on Vision Token Attention Scores} with POPE metric on MSCOCO dataset and CHAIR metric. We select the Top-100 and Least-100 important vision tokens out of a total of 576 vision tokens of LLaVA-1.5 7B, based on Equation \ref{eq5} ($i$=3). \textbf{VD}: Visual Disturbance; \textbf{ID}: Instruction Disturbance.}
\begin{tabular}{l|cccc|cc}
\hline
\multicolumn{1}{c|}{\multirow{2}{*}{\textbf{Setting}}} & \multicolumn{2}{c|}{\textit{Random}}     & \multicolumn{2}{c|}{\textit{Adversarial}} & \multirow{2}{*}{CHAIRs $\downarrow$} & \multirow{2}{*}{CHAIRi $\downarrow$} \\ 
\multicolumn{1}{c|}{}                                  & Accuracy $\uparrow$ &  \multicolumn{1}{c|}{F1 Score $\uparrow$} & Accuracy $\uparrow$           & F1 Score $\uparrow$           &                         &                         \\ \hline
\textbf{Greedy}                                        & 88.8     & 88.6                          & 79.3               & 80.9               & 49.6                    & 14.4                    \\ \cdashline{1-7} 
+Top-100                                      & 85.6     & 83.9                          & 77.1               & 76.3               & 52.7                    & 15.2                    \\
+Least-100                                    & 55.3     & 66.1                          & 54.0               & 65.3               & 63.2                   & 38.7                    \\\cdashline{1-7} 
+\textbf{VD}                                 & 88.0    & 87.6                         & 78.9              & 79.8              &  56.7                       & 16.9                        \\
+\textbf{ID}                                & 88.2      &87.7                          & 79.1              &80.1              & -                        & -                        \\ \hline
\end{tabular}
\label{c5_table2}
\end{table}

\subsection{Context and Text-aware Token Selection (CT$^2$S) Strategy} \label{4.2}
\begin{figure} % 'r'表示右侧，0.4为图片宽度
    \centering
      \centering
      \includegraphics[width=0.9\textwidth]{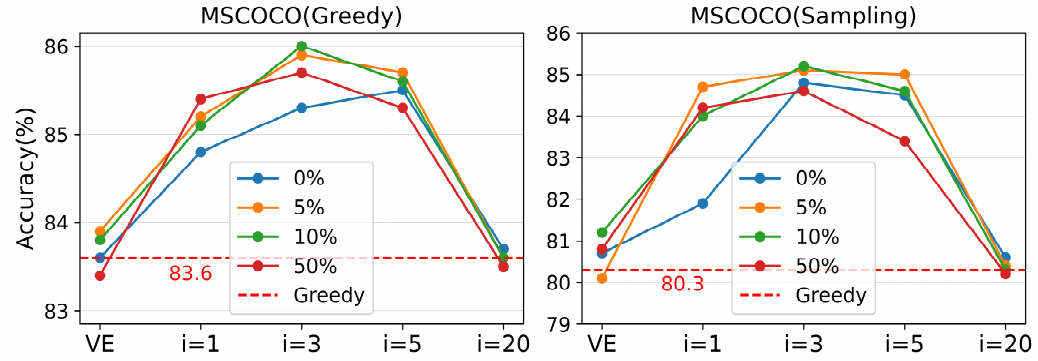}
      \caption{Analyses of varying $i$ and preserved ratios in CT$^2$S. 
      VE: vision encoder; \textbf{$i$}: $i$-th decoder layer.}
\label{fig-i}
\end{figure}

Based on the above investigations, we argue that to induce context- and text-aware hallucinations for contrastive decoding, only a small percentage 
of vision tokens with low attention scores should be preserved after the early decoder layers. 
To validate our claims, we conduct the following experiments: 1) In Vision Encoder (VE), we preserve tokens with low attention values between the $\mathbf{[CLS]}$ token and vision tokens in the penultimate layer, calculated as: $\mathbf{A}=\mathrm{softmax}(\frac{\mathbf{[CLS]}\cdot \mathbf{K^{T}}}{\sqrt{d_{k}} } )$. 2) In the LLM decoder, we preserve tokens with low importance score (Equation \ref{c5_eq5}) across varying layers ($i$). Additionally, we adjust the number of preserved vision tokens. As shown in Figure  \ref{fig-i}, \textbf{firstly}, pruning vision tokens in VE based on $\mathbf{[CLS]}$ may not always yield positive gains, as the $\mathbf{[CLS]}$ token lacks information about instructions and generated texts, which are crucial for multimodal understanding. 
Specifically, pruning all vision tokens resembles VIG \cite{vig}, which contrastively amplifies the vision importance over the language prior by ablating vision inputs. 
\textbf{Secondly}, 
aggressive pruning of vision tokens (i.e., 0$\%$) after $Layer_{i=1}$ is not optimal. As the ideal induced hallucination distributions are \textit{target-co-occurring} but suppress \textit{target logits}, the loss of visual information for subsequent decoding results in visual context diminishing, which can lead to aimless hallucinations due to insufficient grounding in visual information.
\textbf{Thirdly}, selecting tokens in the late decoder layers degrades contrastive decoding to normal decoding, as preceding layers of LVLMs already decode and understand multimodal information, which is consistent with LLMs' early-exiting mechanisms \cite{calm, exit}. In summary, the proposed CT$^2$S strategy selects Top-k least important vision tokens after the early layers based on attention score (Equation \ref{c5_eq5}), where the induced hallucinations are aware of both visual contexts and text information. Finally, following CD methods (Sec. \ref{sec3.2}), we contrastively subtract amplified vision-and-text association hallucinations for the next token prediction.

\noindent \textbf{Discussion.} Based on analyses of the self-introspective pre-trained LVLMs, could we enhance vision information informed by the proceeding vision and text tokens rather than utilizing the contrastive decoding? To explore this, we rewrite Equation \ref{eq2} as follows:
\begin{equation}
p_{add}(y_{i}|v,v\uparrow,t,y_{<i})=\mathrm{softmax}[(1-\alpha)logit_{\theta }(y_{i}|v,t,y_{<i}) + \alpha logit_{\theta  }(y_{i}|v\uparrow,t,y_{<i})]
\label{c5_eq6}
\end{equation}
where we preserve vision tokens with high importance scores (Equation \ref{c5_eq5}) denoted
\begin{table}
\centering 
% \footnotesize
\caption{\textbf{Analyses of Contrastive Decoding Mechanisms} on the POPE metric. Hyperparameters are consistent with CD settings.}
\begin{tabular}{lcc}
\hline
\multicolumn{1}{c}{\textbf{Setting}} & \textit{Random} & \textit{Adversarial} \\ \hline
Sampling                             & 84.7            & 78.7               \\
VCD                                  & 87.7            & 80.9               \\
\texttt{Add}                         & \textbf{88.9}   & 79.4               \\
\textbf{Ours}                        & 88.8            & \textbf{82.6}      \\ \hline
Greedy                               & 88.8            & 79.1               \\
VCD                                  & 87.8            & 80.9               \\
\texttt{Add}                         & 89.1            & 80.1               \\
\textbf{Ours}                        & \textbf{89.3}   & \textbf{83.3}      \\ \hline
\end{tabular}
\label{figadd}
\end{table}
as $v\uparrow$. From Table \ref{figadd}, we observe that enhancing vision information (i.e., \texttt{Add}) alleviates hallucinations to some extent, which also implicitly validates the efficacy of Equation \ref{c5_eq5}. 
However, in the adversarial setting, enhancing vision information does not bring much benefits compared to ours. 
Because our amplified hallucinations effectively associate co-occurring objects,
reflected in \textbf{high logit values of hallucination token}, and then contrastively suppress them. 
In contrast, enhancing vision information primarily boosts the original prediction's \textbf{target} logits grounded in attention scores, which does not significantly improve discrimination, especially in the adversarial setting.

\section{Experimental Results}\label{sec5}

\subsection{Experimental Settings}\label{sec5.1}

\noindent  \textbf{Models and Baselines.} We utilize four representative LVLMs: InstructBLIP \cite{instructblip}, Shikra \cite{shikra}, LLaVA-1.5 \cite{llava1.5} at the 7B scale, and LLaVA-NeXT \cite{llavanext} at the 8B scale. 
For the detailed model descriptions, InstructBLIP employs Q-former \citep{blip2} to condense image tokens to 32, as a result, we are unable to visualize the dynamic token pruning process of InstructBLIP like Figure  \ref{fig3} and \ref{fig4}. Shikra, LLaVA-1.5, and LLaVA-NeXT directly leverage linear projection layers as vision-language connectors to align multimodal features. Shikra and LLaVA-1.5 encode 256 and 576 image tokens to LVLMs. LLaVA-NeXT increases the input vision resolution by 4$\times$ to capture more visual details, resulting in ~4$\times$ more encoded vision tokens than LLaVA-1.5. All LVLMs utilize pre-trained vision encoders like CLIP \citep{clip} vision encoder, as well as pre-trained LLMs as language decoders, such as Vicuna v1.1 \citep{vicuna}, LLaMA 2 \citep{llama2}, and recently released LLaMA 3 \citep{llama3}. We provide results at the 7 Billion (B) scale, and larger-scale results are in Table \ref{table_large}.
Since our method aims to propose \textit{training-free} LVLM decoding strategies \textcolor{black}{\textit{without} the aid of auxiliary networks}, we compare six decoding methods: Sampling (Top-p=1), Greedy, Dola \cite{dola}, and LVLM decoding strategies (VCD \cite{vcd}, ICD \cite{icd}, and OPERA \cite{opera}). For comprehensive comparisons, we apply VCD and ICD in both sampling (Top-p=1) and greedy decoding settings.

\noindent  \textbf{Implementation Details.} As analyzed in Sec. \ref{4.2}, we set Layer $i$=3 and preserve top 10$\%$ least important vision tokens for Shikra, LLaVA-1.5, and LLaVA-NeXT and $i$=5 and top 10$\%$ least important vision tokens for Q-former based LVLMs (InstructBLIP) to induce fine-grained hallucinations.
% Hyperparameters in Equation \ref{eq2} and \ref{eq3} follow VCD and ICD. 
For sampling and greedy decoding, we adopt the default hyperparameter settings. As for Dola \citep{dola}, it is designed to alleviate hallucinations (i.e., improve factuality) of LLM by contrasting the differences in logits obtained from projecting the later layers versus premature layers. Dola is sensitive to the premature layer selection, we adapt Dola to LVLMs, following OPERA \citep{opera} to utilize ``0,2,4,6,8,10,12,14” as the indexes of candidate premature layers and ``32” as the index of the mature layer. The repetition penalty is set to 1.2, as Dola suggests. OPERA, VCD, and ICD are proposed for LVLMs and we adopt the default settings. For fair comparisons, SID's hyperparameters of Equation \ref{eq2} and \ref{eq3} follow VCD and ICD. Moreover, we apply SID, VCD, and ICD in both sampling (Top-p=1) and greedy decoding settings for comprehensive comparisons.
Note that due to amplified fine-grained hallucinations, SID is more robust to hyperparameters compared to other CD methods (Sec. \ref{sec6}). Experiments are performed on NVIDIA V100/A100 GPUs.

\subsection{Evaluation Results}\label{sec5.2}

In this section, we follow previous methods \cite{vcd, icd, opera} to evaluate the SID on \textbf{CHAIR} \cite{chair} and \textbf{POPE} \cite{pope} metrics. Besides manually designed metrics, we also leverage \textbf{GPT-4 assisted benchmark} \cite{hadpo} to evaluate attribute, location, and relation hallucinations. \textbf{MME} \cite{mme} and \textbf{MMBench} \cite{mmbench} benchmarks are employed to assess the LVLM's general ability. Moreover, \textbf{GPT4-V assisted evaluation} on both hallucination alleviation and generated text quality and \textbf{Case study} of several pictures are illustrated.

\begin{table}[t] 
\centering 
\caption{\textbf{Results on the CHAIR metric.} $^{*}$ and $^{\star}$ denote adopting the same sampling and greedy decoding strategies, respectively.}
\begin{tabular}{l|cccccccc} \hline
\multicolumn{1}{c|}{\multirow{2}{*}{\textbf{Setting}}} & \multicolumn{2}{c}{\textbf{LLaVA-1.5}} & \multicolumn{2}{c}{\textbf{InstructBLIP}} & \multicolumn{2}{c}{\textbf{Shikra}}    & \multicolumn{2}{c}{\textbf{LLaVA-NeXT}} \\ 
\multicolumn{1}{c|}{}                                  & $C_S$$\downarrow$            & \multicolumn{1}{c}{$C_I$$\downarrow$} & $C_S$$\downarrow$              & \multicolumn{1}{c}{$C_I$$\downarrow$}  & $C_S$$\downarrow$            & \multicolumn{1}{c}{$C_I$$\downarrow$} & $C_S$$\downarrow$                 & $C_I$$\downarrow$                 \\ \hline 
Sampling                                               & 51.3          & 16.8                    & 51.0            & 24.2                     & 48.9          & 14.7                    & 42.6               & 14.1               \\
ICD$^{*}$                                             & 48.7          & 13.9                     & 48.3                & 16.7                         & 47.8          & 14.5                    & 42.7               & 13.6               \\
VCD$^{*}$                                               & 48.0          & 14.3                   &47.9                 & 17.2                         & 48.1          & 13.8                    & 41.3               & 12.9               \\
\textbf{Ours}$^{*}$                                              & 45.0          & \textbf{11.7}          & 43.6            & 13.1                     & 46.0          & 12.9                    & 38.4               & 11.4               \\ \cdashline{1-9} 
Greedy                                                 & 49.6          & 14.4                    & 54.6            & 13.6                     & 47.1          & 13.9                    & 42.9               & 13.2               \\
Dola$^{\star}$                                         & 47.1          & 13.8                    & 52.7            & 14.0                     & 46.8          & 14.2                    & 40.9               & 13.1               \\
OPERA                                                  & 45.2          & 12.7                    & 47.4                & 12.9                  & \textbf{44.4} & 13.6                    & 39.4               & 11.8               \\
ICD$^{\star}$                                           & 47.4          & 13.9                    &46.3                 & 15.3                 & 47.3          & 14.1                    & 42.1               & 12.6               \\
VCD$^{\star}$                                          & 46.8          & 13.2                    & 44.0            & 13.6                     & 47.8          & 14.0                    & 41.1               & 12.9               \\
\textbf{Ours}$^{\star}$                                                   & \textbf{44.2} & 12.2                    & \textbf{42.3}   & \textbf{12.4}            & 44.8          & \textbf{12.8}           & \textbf{38.1}      & \textbf{11.3}      \\ \hline
\end{tabular}
\label{chair}
\end{table}

\noindent\textbf{CHAIR and POPE Evaluations.}
CHAIR \cite{chair} and POPE \cite{pope} are quantitative metrics to assess objection hallucinations of VLMs. 
The Caption Hallucination Assessment with Image Relevance (CHAIR) \citep{chair} metric is specially designed to assess objection hallucinations in the image caption tasks. Concretely, CHAIR quantifies the degree of hallucinations in a generated image caption by calculating the proportion of all objects mentioned in the caption that are not present in the ground truth label pool. There are two common variants of CHAIR: CHAIRi ($C_{I}$) and CHAIRs ($C_{S}$), which evaluate the degree of object hallucination in the instance and sentence level, respectively. These two metrics are formulated as follows:
\begin{equation}
C_{I}=\frac{| \{ \mathrm{hallucinated  \ objects \}} |}{| \{ \mathrm{all \ mentioned  \ objects} \} |} , \ C_{S}=\frac{| \{ \mathrm{captions \ with \ hallucinated \ objects} \} |}{|\mathrm{\{ all \ captions} \}|}
\end{equation}
The smaller the value of $C_{I}$ and $C_{S}$, the better the hallucination alleviation performance.
The Polling-based Object Probing Evaluation (POPE) \citep{pope} was recently developed to assess hallucination problems in LVLMs. POPE queries the LVLMs with the template: \texttt{Is there a <object> in the image?} The ratio between queries about existing and no-existing objects is balanced (i.e., 50$\%$-50$\%$). This benchmark consists of three sampling settings: \textit{random}, \textit{popular}, and \textit{adversarial}, each differing in the construction of negative samples. Specially, in the \textit{random} setting, objects that are not present in the image are selected at random. The \textit{popular} setting selects missing objects from the high-frequency pool, whereas in the \textit{adversarial} setting, co-occurring objects that are not present in the image are prioritized. POPE consists of three different datasets, including MSCOCO \citep{mscoco}, A-OKVQA \citep{aokvqa}, and GQA \citep{gqa}. POPE involves 500 images from each dataset with six questions each, ultimately yielding 27,000 query-answer pairs. Accuracy and F1 score are chosen as evaluation metrics. The larger the value of Accuracy and F1 score, the better the hallucination alleviation performance.
% Please refer to the Appendix \ref{eval_metric} for detailed descriptions of CHAIR and POPE.
As for \textbf{CHAIR}, Following \cite{icd, opera, eos}, we randomly select 500 images from the validation set of the MSCOCO \cite{mscoco} dataset and query different LVLMs with the prompt: \texttt{`Please describe this image in detail.'}. We set the max new tokens to 512 to generate responses for fair comparisons. As shown in Table \ref{chair}, our method outperforms other baselines in most cases, validating the effectiveness of SID in open-end generation tasks. Compared to CD methods, SID \textit{online adaptively} prunes attention-important vision tokens informed by instruction and generated text to induce fine-grained hallucinations for contrastive decoding during open-end text generations. 
For the \textbf{POPE} metric, which comprises three datasets, we average the results in Table \ref{pope}. Our method performs best overall in \textit{random}, \textit{popular}, and \textit{adversarial} sampling settings. Specifically, in the sampling decoding setting, SID surpasses the normal sampling decoding by a large margin in a train-free manner. SID also clearly outperforms CD methods (Dola, ICD, and VCD) because the self-introspective decoding strategy amplifies \textit{vision-and-text association} hallucinations then subtracts them, rather than coarsely disturbing raw inputs. Additionally, owing to the context and text-aware token selection strategy, SID is more computation-efficient than CD methods, as analyzed in Table \ref{table_e}. Note that beam-search based OPERA \cite{opera} shows almost no gain in the POPE metric, primarily because answering the binary classification only requires a few tokens and selecting the best beam score in a decoded sequence ($N$=5) brings little improvement.

\begin{table}[] 
\centering \scriptsize
\caption{\textbf{Average results on the POPE metric.} $^{*}$ and $^{\star}$ denote adopting the same sampling and greedy decoding strategies, respectively. Results are from the original papers or re-implemented based on official codes.}
\begin{tabular}{cccc|cc|cc}
\hline
\multicolumn{2}{c}{\textbf{Setting}}               & \multicolumn{2}{c|}{\textit{Random}}                         & \multicolumn{2}{c|}{\textit{Popular}}                        & \multicolumn{2}{c}{\textit{Adversarial}}                      \\ \bottomrule
\textbf{Model}                          & \textbf{Decoding} & \multicolumn{1}{l}{Accuracy$\uparrow$} & \multicolumn{1}{l|}{F1 Score$\uparrow$} & \multicolumn{1}{l}{Accuracy$\uparrow$} & \multicolumn{1}{c|}{F1 Score$\uparrow$} & \multicolumn{1}{l}{Accuracy$\uparrow$} & \multicolumn{1}{l}{F1 Score$\uparrow$} \\ \hline
\multirow{10}{*}{\rotatebox{90}{\textbf{LLaVA-1.5}}}    & Sampling          & 84.77                        & 82.28                         & 79.98                        & 79.34                         & 76.03                        & 76.26                        \\
                                        & ICD$^{*}$         & 87.51                        & 83.28                         & 83.15                        & 83.91                         & 79.13                        & 80.41                        \\
                                        & VCD$^{*}$         & 86.84                        & 86.83                         & 82.65                        & 83.37                         & 77.31                        & 79.28                        \\
                                        & \textbf{Ours}$^{*}$        & 88.91                        & 88.84                         & 83.97                        & 85.42                         & 82.54                        & 81.98                        \\ \cdashline{2-8}
                                        & Greedy            & 88.81                        & 88.52                         & 82.76                        & 83.36                         & 79.11                       & 80.92                        \\
                                        & Dola$^{\star}$    & 87.94                        & 87.97                         & 83.87                        & 84.68                         & 80.35                        & 81.21                        \\
                                        & OPERA            & 88.85                        & 88.67                         & 82.77                        & 83.40                         & 79.16                        & 80.93                        \\
                                        & ICD$^{\star}$         & 87.97                        & 87.84                         & 84.03                        & 84.22                         & 80.21                        & 80.97                        \\
                                        & VCD$^{\star}$       & 87.02                        & 86.96                         & 83.53                        & 84.56                         & 78.12                        & 80.16                        \\
                                        & \textbf{Ours}$^{\star}$      & \textbf{89.46}               & \textbf{89.62}                & \textbf{85.13}               & \textbf{85.94}                & \textbf{83.24}               & \textbf{82.21}               \\\bottomrule
\multirow{10}{*}{\rotatebox{90}{\textbf{InstructBLIP}}} & Sampling          & 80.42                        & 80.94                         & 76.09                        & 77.65                         & 72.37                        & 75.42                        \\
                                        & ICD$^{*}$         & 85.78                        & 85.73                         & 81.12                        & 82.25                         & 76.82                        & 78.99                        \\
                                        & VCD$^{*}$         & 84.11                        & 84.13                         & 79.96                        & 80.80                         & 76.32                        & 78.08                        \\
                                        & \textbf{Ours}$^{*}$        & 86.56                        & 85.94                         & 80.26                        & 81.75                         & 77.64                        & 80.41                        \\ \cdashline{2-8}
                                        & Greedy            & 84.56                        & 83.75                         & 78.23                        & 79.16                         & 74.58                        & 76.34                        \\
                                        & Dola$^{\star}$     & 84.67                        & 83.38                         & 78.21                        & 79.19                         & 75.69                        & 77.98                        \\
                                        & OPERA            & 84.57                        & 83.74                         & 78.24                        & 79.15                         & 74.59                        & 76.33                        \\
                                        & ICD$^{\star}$         & 84.36                        & 83.82                         & 77.88                        & 78.70                         & 75.17                        & 77.23                        \\
                                        & VCD$^{\star}$      & 84.52                        & 83.63                         & 78.04                        & 78.45                         & 75.95                        & 77.76                        \\
                                        & \textbf{Ours}$^{\star}$       & \textbf{87.23}               & \textbf{86.90}                & \textbf{81.16}               & \textbf{82.57}                & \textbf{78.51}               & \textbf{81.26}               \\ \bottomrule
\multirow{10}{*}{\rotatebox{90}{\textbf{Shikra}}}       & Sampling          & 81.42                        & 82.46                         & 79.60                        & 80.78                         & 73.85                        & 76.39                        \\
                                        & ICD$^{*}$         & 82.34                        & 82.82                         & 78.17                        & 80.43                         & 74.96                        & 77.68                        \\
                                        & VCD$^{*}$        & 82.31                        & 82.73                         & 79.34                        & 80.93                         & 75.61                        & 77.96                        \\
                                        & \textbf{Ours}$^{*}$        & 83.87                        & 83.94                         & 80.26                        & 82.07                         & 77.85                        & 78.94                        \\ \cdashline{2-8}
                                        & Greedy            & 83.00                        & 83.19                         & 81.39                        & 81.90                         & 76.69                        & 78.31                        \\
                                        & Dola$^{\star}$      & 82.87                        & 82.98                         & \textbf{82.42}             & 82.50                         & 76.85                        & 78.09                        \\
                                        & OPERA             & 83.05                        & 83.20                         & 81.40                        & 81.89                         & 76.73                        & 78.31                        \\
                                        & ICD$^{\star}$        & 82.67                        & 82.64                         & 80.73                        & 81.58                         & 75.98                        & 78.43                        \\
                                        & VCD$^{\star}$       & 82.96                        & 82.63                         & 80.68                        & 81.27                         & 76.94                        & 78.32                        \\
                                        & \textbf{Ours}$^{\star}$        & \textbf{84.46}               & \textbf{84.62}                & 82.38                    & \textbf{82.73}                & \textbf{78.67}               & \textbf{79.34}               \\  \bottomrule
\multirow{10}{*}{\rotatebox{90}{\textbf{LLaVA-NeXT}}}   & Sampling          & 86.32                        & 83.11                         & 82.27                        & 81.03                         & 77.32                        & 76.96                        \\
                                        & ICD$^{*}$         & 87.32                        & 84.03                         & 83.62                        & 83.54                         & 80.31                        & 80.41                        \\
                                        & VCD$^{*}$         & 86.97                        & 86.71                         & 83.07                        & 83.65                         & 79.42                        & 80.28                        \\
                                        & \textbf{Ours}$^{*}$        & 89.16                        & 88.92                         & 84.38                        & 85.76                         & 82.95                        & 81.98                        \\ \cdashline{2-8} 
                                        & Greedy            & 89.37                        & 88.82                         & 83.68                        & 84.62                         & 80.08                        & 80.74                        \\
                                        & Dola$^{\star}$     & 88.73                        & 88.67                         & 84.56                        & 84.96                         & 80.32                        & 80.68                        \\
                                        & OPERA            & 89.36                        & 88.80                         & 83.65                        & 84.60                         & 80.10                        & 80.75                        \\
                                        & ICD$^{\star}$        & 87.40                        & 87.96                         & 84.11                        & 83.79                         & 80.94                        & 80.67                        \\
                                        & VCD$^{\star}$       & 87.83                        & 87.09                         & 82.68                        & 83.55                         & 79.61                        & 81.20                        \\
                                        & \textbf{Ours}$^{\star}$        & \textbf{90.05}               & \textbf{89.97}                & \textbf{86.13}               & \textbf{85.69}                & \textbf{84.06}               & \textbf{82.95}               \\ \bottomrule

\end{tabular}

\label{pope}
\end{table}

\begin{figure*}[]
  \centering
  \includegraphics[width=1.05\textwidth]{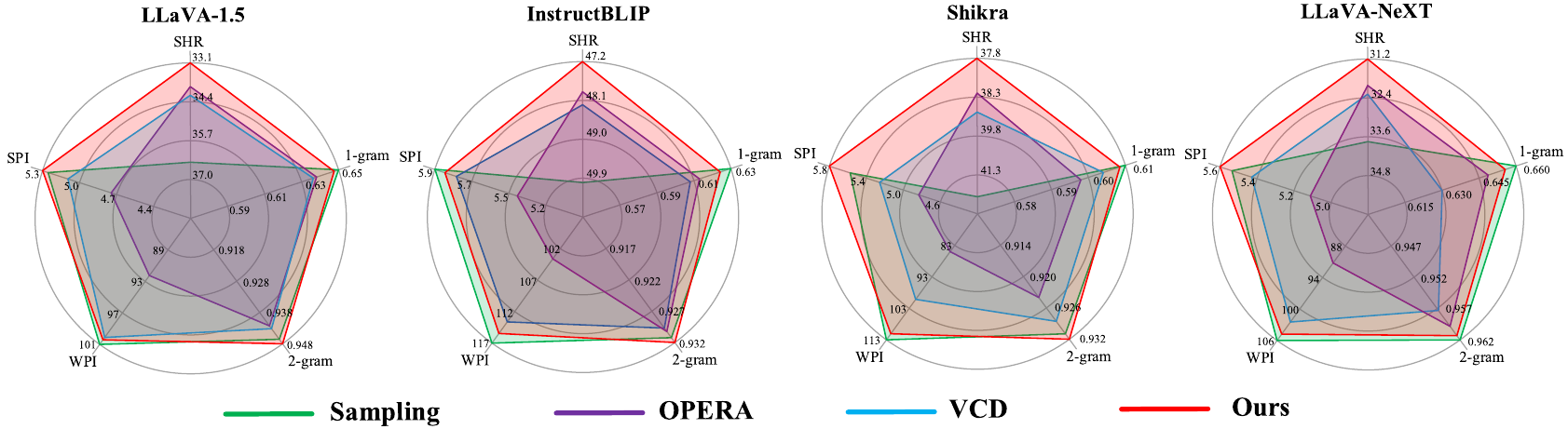}
  \caption{\textbf{GPT-4 assisted benchmark} \cite{hadpo}. \textbf{Hallucination} (SHR), \textbf{fluency} (1$\&$2-gram), and \textbf{detailness} (WPI and SPI) aspects are compared.  % Larger WPI, SPI, 1/2-gram and lower SHR are better. 
  Larger areas mean better performances. VCD and ours adopt the same sampling decoding. Please zoom in for details.}
  \label{gpt4}
\end{figure*}

\begin{figure*}[ht]
  \centering
  \includegraphics[width=1\textwidth]{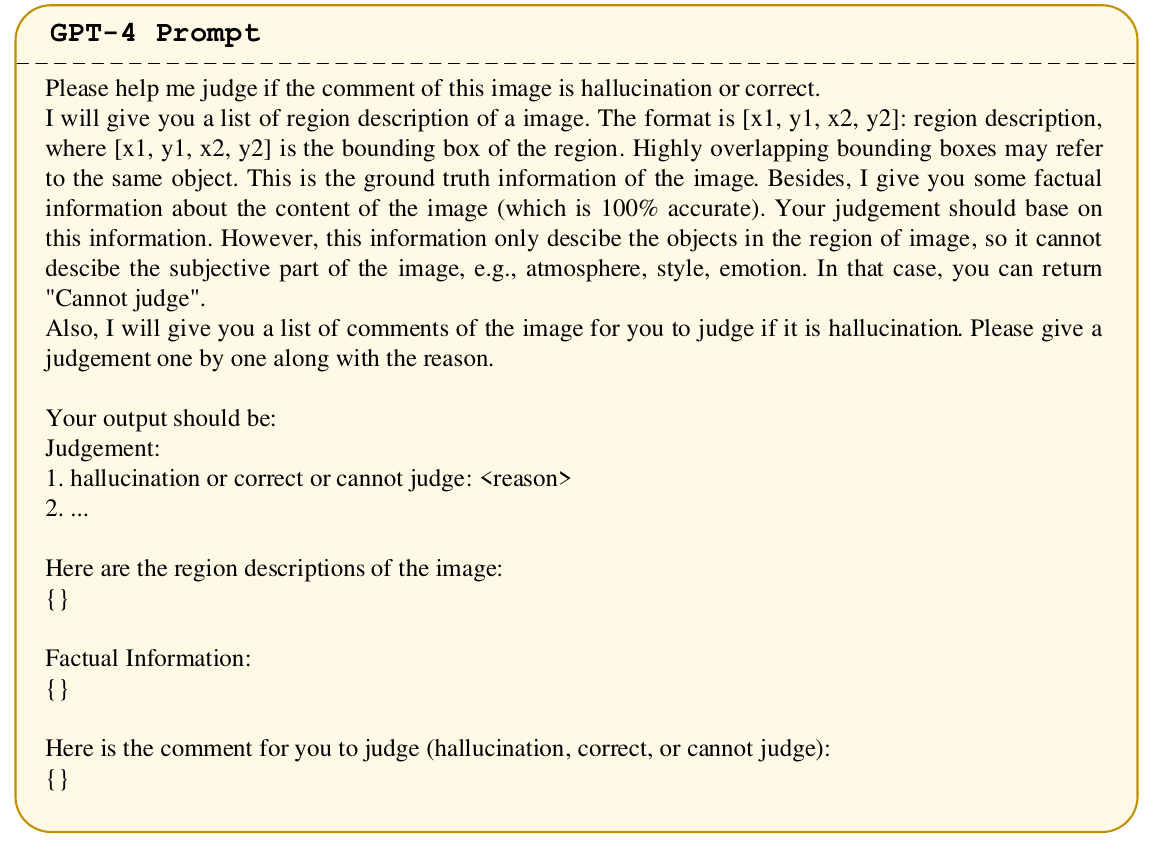}
  \caption{\textbf{Prompts of GPT-4 for evaluations.}}
  \label{fig-gpt4}
\end{figure*}

\noindent\textbf{GPT-4 Assisted Benchmark.} While CHAIR and POPE evaluate object-existence-level hallucinations, these metrics are unable to identify other types of hallucination, such as \textit{positional}, \textit{relational}, and \textit{attribute} hallucinations. Therefore, the GPT-4 assisted benchmark \cite{hadpo} utilizes the fine-grained object-level descriptions in the Visual Genome (VG) dataset \cite{vg} as ground-truth and relies on the advanced GPT-4 to judge the fine-grained hallucinations and calculate Sentence-level Hallucination Ratio (SHR). Concretely, besides object-existence-level hallucinations evaluated by CHAIR and POPE, GPT-4 assisted benchmark \citep{hadpo} utilizes the fine-grained object-level description in the Visual Genome (VG) dataset \citep{vg} as ground-truth and relies on the advanced GPT-4 to judge the detailed (such as \textit{positional}, \textit{relational}, and \textit{attribute}) hallucinations and calculate Sentence-level Hallucination Ratio (SHR). With the generated sentences and manually annotated factual information, GPT-4 is prompted to evaluate whether existing hallucinations sentence by sentence. The prompt template is provided in Figure  \ref{fig-gpt4}. Following \citep{hadpo}, we utilize 200 images from the VG dataset and set max new tokens to 512, with the prompt of  \texttt{`Please describe this image in detail.'} We conduct experiments on sampling decoding strategies and representative LVLMs decoding strategies: VCD \citep{vcd} and OPERA \citep{opera}. 
Moreover, we employ n-gram fluency (n = 1 and 2) metrics to measure the smoothness of generated text, and the number of generated words/sentences per image (WPI/SPI) to compare the detailedness of generated texts.
As shown in Figure  \ref{gpt4}, SID achieves the best results in the SHR metric among the four LVLMs, outperforming others by a clear margin. 
Regarding the quality of the generated texts, 
Sampling decoding outperforms ours slightly in terms of 1-gram fluency and WPI. However, compared to other baselines, our approach alleviates hallucinations with minimal sacrifice in text generation quality regarding smoothness and detailness. For instance, OPERA generates text with fewer words and sentences due to penalization of the over-trust mechanism, and VCD impairs text fluency, possibly arising from the holistic and fixed disturbance of contrastive inputs.

\noindent\textbf{MME and MMBench Evaluations.}
Besides, we test on two popular LVLMs' general ability benchmarks: MME and MMBench. MME comprises ten subtasks to evaluate
models’ perceptual capabilities and four subtasks for assessing recognitive abilities in the form of the yes/no question. MMBench systematically evaluates twenty ability dimensions of LVLMs. We
present the results of LLaVA-1.5 7B as a representative in Table \ref{benchmark}, SID can maintain and improve the multimodal ability on LVLMs benchmarks. In contrast, other CD methods tend to compromise the general multimodal ability. 

\begin{figure*}[ht]
  \centering
  \includegraphics[width=1\textwidth]{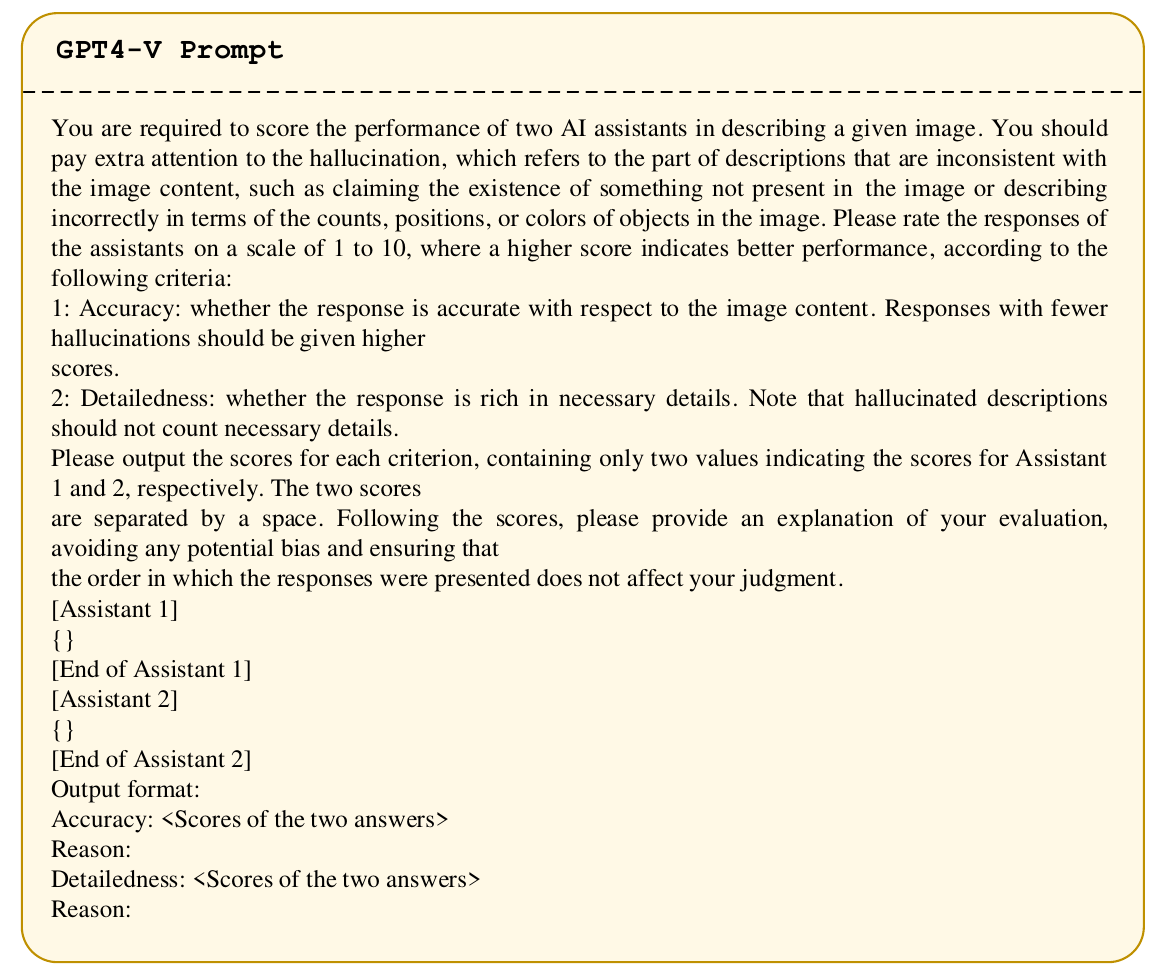}
  \caption{\textbf{Prompts of GPT4-V for evaluations.}}
  \label{fig-gpt4v}
\end{figure*}

\noindent\textbf{GPT4-V Assisted Evaluation.}
To further analyze the hallucinations and text quality for open-end generation tasks, following \cite{opera, pecker}, we utilize the strong multi-modal assistant GPT4-V, which simultaneously processes input from vision and text modalities. 
% We randomly sample 500 images from MSCOCO’s validate set and ask different MLLM models to describe these images
We strictly follow \cite{opera}, which utilizes 500 images from the MSCOCO dataset and prompts LVLM:\texttt{`Please describe this image in detail.'} with the maximum number of 512.
To mitigate the impact of the sequential order fed to GPT4-V, we simultaneously compare the generated texts obtained from two decoding methods and instruct GPT4-V to judge the correctness and detailedness score on a scale of 0-10 based on the input image. The detailed GPT4-V prompt is in Figure  \ref{fig-gpt4v}. We set up three representative pairs of comparison experiments: greedy decoding and ours, CD-based VCD \cite{vcd} and ours, and OPERA \cite{opera} and ours. As shown in Table \ref{gpt4-v}, our SID achieves the best results in terms of most metrics. Concretely, our method improves correctness by about 15-20$\%$ compared to sampling decoding while not compromising the detailedness level. Compared to advanced hallucination mitigation methods VCD and OPERA, SID generates text with obvious more details and better mitigates the hallucination issue. Since the perceptual and reasoning capabilities of GPT4-V are very close to those of humans, the results of the GPT4-V evaluation reflect, to some extent, the strong performance of the compared methods in terms of mitigating hallucinations and generating text quality from a human perceptual perspective.

\begin{table}[] 
\centering 
\caption{\textbf{GPT4-V assisted hallucination evaluations} \cite{opera, pecker}. VCD and ours adopt the same sampling decoding strategy. $C$: correctness; $D$: detailedness}
\vspace{-0.2cm}
\begin{tabular}{lcccccccc}
\hline
\multicolumn{1}{c}{\multirow{2}{*}{\textbf{Setting}}} & \multicolumn{2}{c}{\textbf{LLaVA-1.5}} & \multicolumn{2}{c}{\textbf{InstructBLIP}} & \multicolumn{2}{c}{\textbf{\  \  \  \  \  \  Shikra\  \  \  \  \  \ }} & \multicolumn{2}{c}{\textbf{LLaVA-NeXT}} \\ 
\multicolumn{1}{c}{}                                  & \textit{C}$\uparrow$         & \textit{D}$\uparrow$        & \textit{C}$\uparrow$          & \textit{D}$\uparrow$          & \textit{C}$\uparrow$       & \textit{D}$\uparrow$       & \textit{C}$\uparrow$         & \textit{D}$\uparrow$         \\ \hline
Sampling                                                & 5.18               & 5.79              & 4.73                & 5.10                & 5.03             & \textbf{5.17}    & 5.34               & 5.67               \\
\textbf{Ours}                                         & \textbf{5.97}      & \textbf{6.01}     & \textbf{5.62}       & \textbf{5.16}       & \textbf{5.78}    & 5.10             & \textbf{6.47}      & \textbf{5.85}      \\ \hline
VCD                                                   & 5.46               & 5.63              & 4.98                & 5.21                & 5.31             & 5.24             & 5.92               & 5.47               \\
\textbf{Ours}                                         & \textbf{6.16}      & \textbf{5.9}4              & \textbf{5.37}       & \textbf{5.46}       & \textbf{5.61}    & \textbf{5.29}    & \textbf{6.12}      & \textbf{5.78}      \\ \hline
OPERA                                                 & \textbf{6.16}      & 5.57              & 5.29                & 4.86                & 5.34             & 4.87             & 6.11               & 5.24               \\
\textbf{Ours}                                         & 6.15               & \textbf{5.94}     & \textbf{5.76}       & \textbf{5.42}       & \textbf{5.97}    & \textbf{5.88}    & \textbf{6.63}      & \textbf{6.23}      \\ \hline
\end{tabular}
\label{gpt4-v}
\end{table}

\begin{table}[]
\centering \footnotesize
\caption{\textbf{LVLM benchmark evaluations.} DoLa, ICD, VCD, and SID employ the same greedy decoding. }
\begin{tabular}{lccccccc}
\hline
        & Greedy & Sampling & DoLa   & ICD    & VCD    & OPERA  & \textbf{SID}    \\ \hline
MME     & 1510.8\textcolor{black}{{\tiny±1.2}} & 1471.5\textcolor{black}{{\tiny±5.6}}   & 1480.7\textcolor{black}{{\tiny±1.3}} & 1473.2\textcolor{black}{{\tiny±1.2}} & 1488.5\textcolor{black}{{\tiny±0.8}} & 1515.2\textcolor{black}{{\tiny±1.1}} & \textbf{1520.4}\textcolor{black}{{\tiny±0.9}} \\
MMbench & 64.4\textcolor{black}{{\tiny±.22}}   & 63.9\textcolor{black}{{\tiny±.81}}     & 63.7\textcolor{black}{{\tiny±.22}}   & 63.0\textcolor{black}{{\tiny±.24}}   & 63.8\textcolor{black}{{\tiny±.22}}   & 64.4\textcolor{black}{{\tiny±.13}}   & \textbf{65.0}\textcolor{black}{{\tiny±.23}}   \\ \hline
\end{tabular}
\label{benchmark}
\end{table}

\begin{figure*}[ht]
  \centering
  \includegraphics[width=1\textwidth]{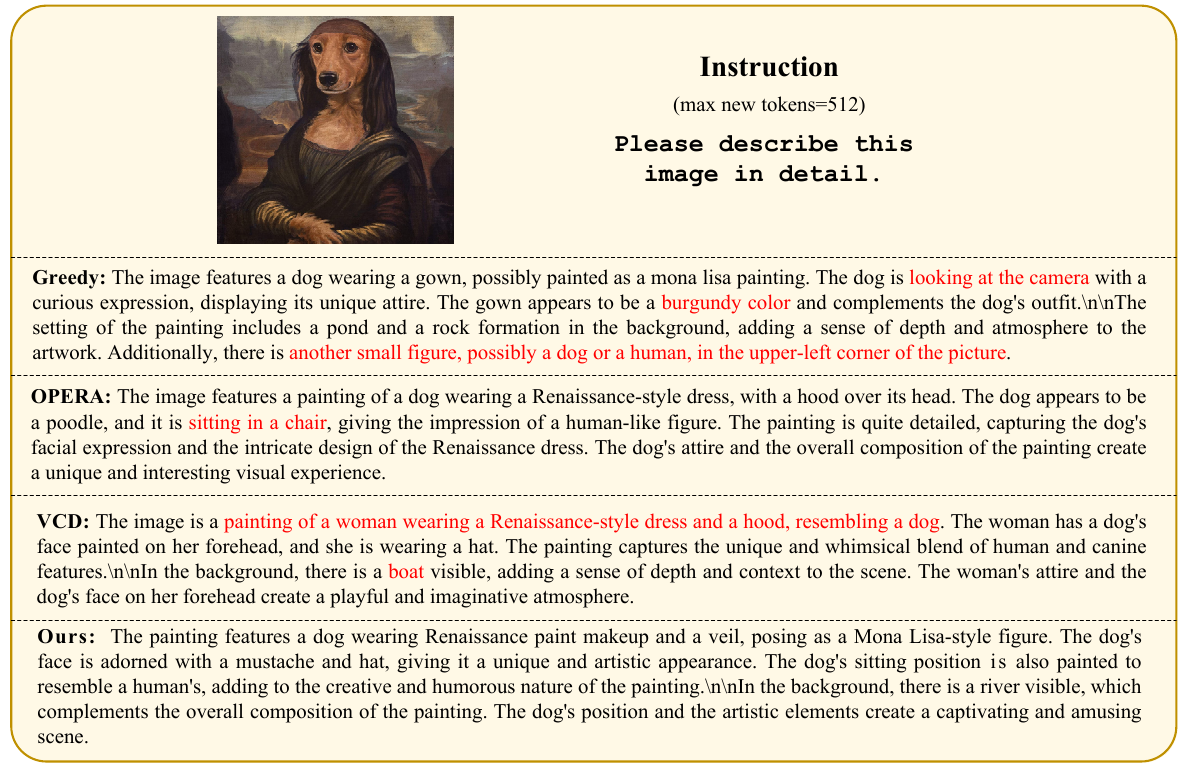}
  \caption{\textbf{Case Study from LLaVA-Bench-in-the-Wild} on LLaVA-1.5 7B. Hallucinations are marked in \textcolor{red}{red}. }
  \label{case1}
\end{figure*}

\begin{figure*}[ht]
  \centering
  \includegraphics[width=1\textwidth]{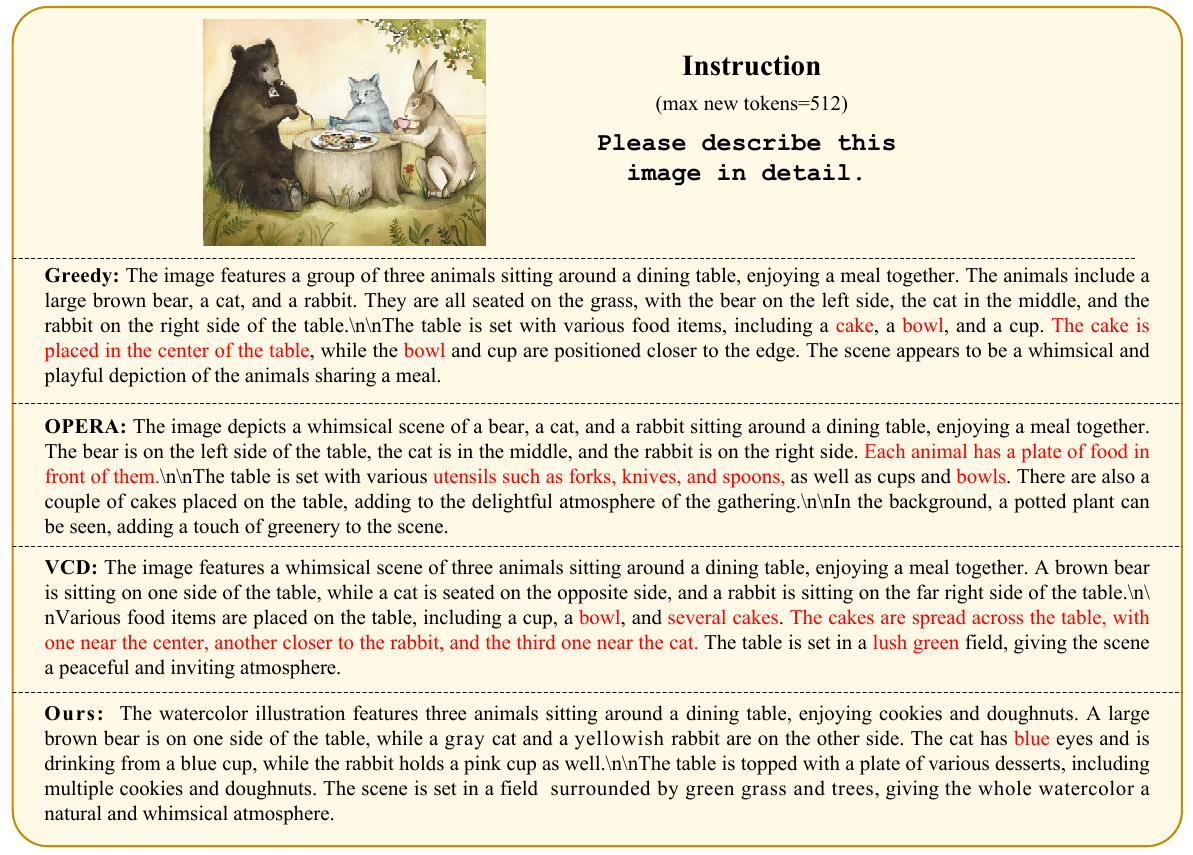}
  \caption{\textbf{Case Study from LLaVA-Bench-in-the-Wild} on LLaVA-1.5 7B. Hallucinations are marked in \textcolor{red}{red}. } 
  \label{case2}
\end{figure*}

\begin{figure*}[ht]
  \centering
  \includegraphics[width=1\textwidth]{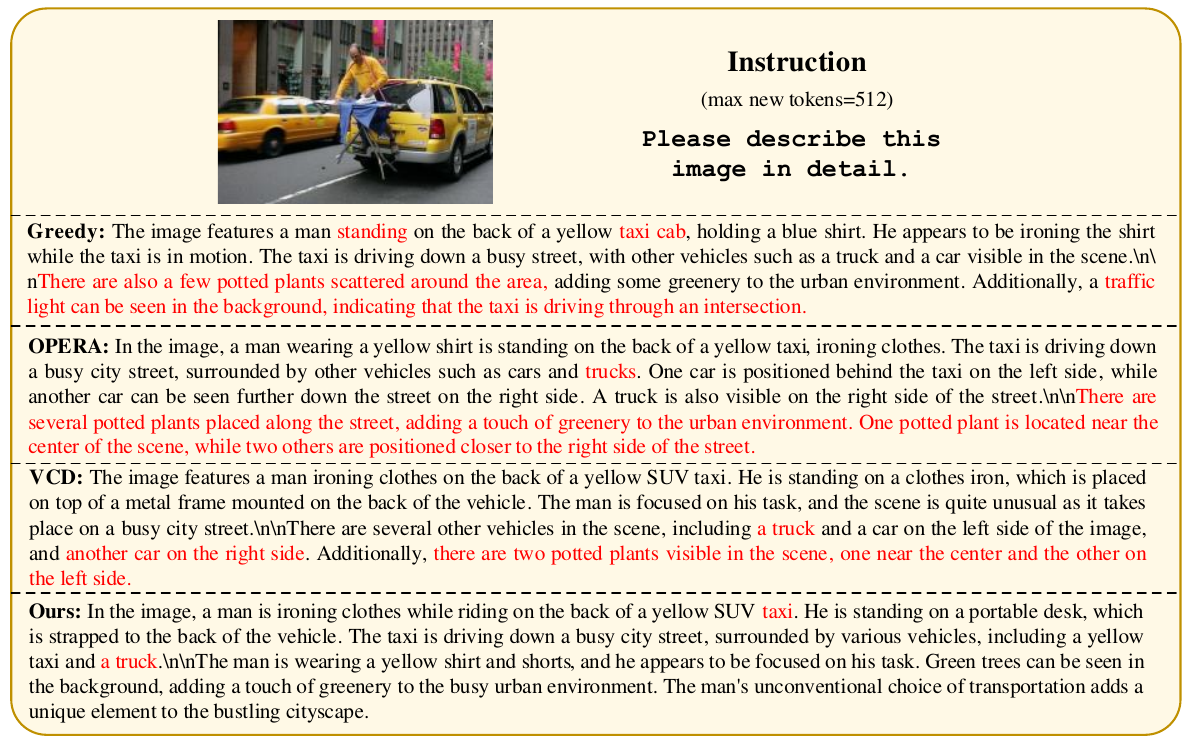}
  \caption{\textbf{Case Study from LLaVA-Bench-in-the-Wild} on LLaVA-1.5 7B. Hallucinations are marked in \textcolor{red}{red}.}
  \label{case3}
\end{figure*}

\textbf{Case Study.} 
In addition to using crafted metrics (CHAIR and POPE), GPT-4/GPT4-V-aided evaluations, and MME \citep{mme} and MMBench \citep{mmbench} benchmarks, we qualitatively present several case studies of SID's hallucination alleviation ability from LLaVA-Bench-in-the-Wild dataset \citep{llava}. As illustrated in Figure  \ref{case1}, \ref{case2}, and \ref{case3}, SID effectively mitigates hallucination in these challenging scenes by dynamically amplifying vision-and-text association hallucinations. Meanwhile, it preserves the detailness of each image. As we propose a training-free decoding method that does not rely on auxiliary analysis networks, it inherently carries over the existing weaknesses of LVLMs. Intuitive case studies, as illustrated in Figure  \ref{case1}, \ref{case2}, and \ref{case3}, reveal that SID still generates some hallucinations, particularly in finer details such as eye color and vehicle identification specifics. These failures may be attributed to the vision encoder's relatively limited visual perception ability. 
For future work, it is promising to integrate SID with InternVL \citep{internvl}, which scales the vision encoder up to 6B,  or consider leveraging auxiliary analysis networks like Grounding DINO \citep{dino} or OWLv2 \citep{owlv2} to mitigate LVLMs' internal weaknesses.

\section{Ablation Analyses}\label{sec6}
In this section, we conduct ablation analyses in terms of the \textbf{Computation Efficiency} and \textbf{Hyperparameter Sensitivity}, \textbf{Larger-scale Backbones}, \textbf{Other Decoding Strategies}, and \textbf{Visual Enhancing Decoding Strategy}.

\begin{table} \centering
% \scriptsize
\begin{tabular}{lccc}
\hline
Methods  & \textbf{Time} $\downarrow$ & \textbf{Memory}$\downarrow$ & \textbf{Accuracy}$\uparrow$ \\ \hline
Normal  & 494  & 15673  & 79.11\\
VCD      & 904  & 16753  & 78.12     \\
ICD      & 974  & 16843  & 80.21     \\
OPERA    & 2643 & 21943  & 79.16        \\
\textbf{Ours}$_{40\%}$ & 704  & 15809  & 83.11     \\
\textbf{Ours}$_{10\%}$ & 668  & 15767  & 83.24     \\ \hline
\end{tabular}
\captionof{table}{\textbf{Efficiency Comparisons} on NVIDIA V100. 
$_{10\%}$ and $_{40\%}$ mean tokens preserved ratios.}
% on POPE adversarial setting with LLaVA-1.5 7B using an NVIDIA V100. Inference time (s) and peak GPU memory (MB) are recorded. {10\%} and {40\%} mean tokens preserved ratios.}
\label{table_e}
\end{table}

\begin{figure} \centering
\includegraphics[width=0.7\textwidth]{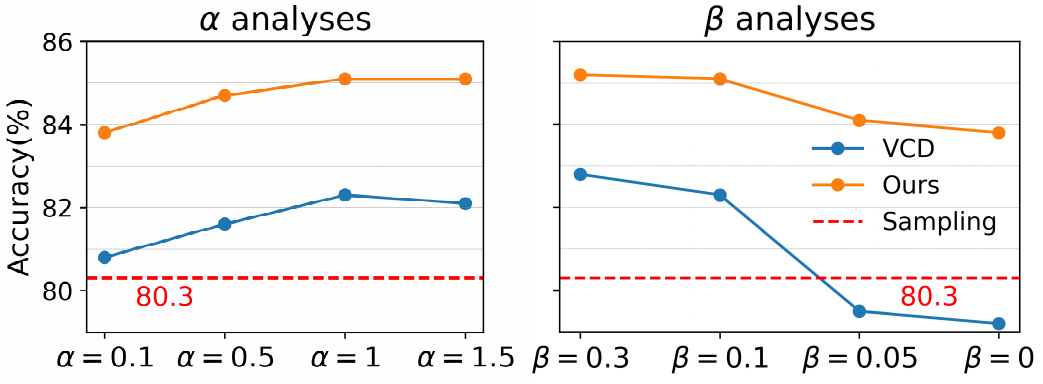}
\caption{\textbf{Hyperparameter Sensitivity of $\alpha$ and $\beta$} with POPE metric (under the sampling decoding).}
\label{figab}
\end{figure}

% \begin{figure}[htbp]
%   \centering
%   % 图片
%   % 表格
%   \begin{minipage}[b]{0.42\textwidth}
%     \scriptsize
%     \begin{tabular}{lccc}
%     \hline
%     Methods  & \textbf{Time} $\downarrow$ & \textbf{Memory}$\downarrow$ & \textbf{Accuracy}$\uparrow$ \\ \hline
%     Normal  & 494  & 15673  & 79.11\\
%     VCD      & 904  & 16753  & 78.12     \\
%     ICD      & 974  & 16843  & 80.21     \\
%     OPERA    & 2643 & 21943  & 79.16        \\
%     \textbf{Ours}$_{40\%}$ & 704  & 15809  & 83.11     \\
%     \textbf{Ours}$_{10\%}$ & 668  & 15767  & 83.24     \\ \hline
%     \end{tabular}
%     \captionof{table}{\textbf{Efficiency Comparisons} on NVIDIA V100. 
%     $_{10\%}$ and $_{40\%}$ mean tokens preserved ratios.}
%     % on POPE adversarial setting with LLaVA-1.5 7B using an NVIDIA V100. Inference time (s) and peak GPU memory (MB) are recorded. {10\%} and {40\%} mean tokens preserved ratios.}
%     \label{table_e}
%   \end{minipage}
% \hspace{0.1cm}   % 添加一些空白或者可以使用 \quad 来增加空间
%   \begin{minipage}[b]{0.52\textwidth}
%     \includegraphics[width=\textwidth]{Figures/alpha_beta}
%     \caption{\textbf{Hyperparameter Sensitivity of $\alpha$ and $\beta$} with POPE metric (under the sampling decoding).}
%     \label{figab}
%   \end{minipage}
% \end{figure}
% \vspace{-0.2cm}

\noindent\textbf{Computation Efficiency.} 
One primary concern of hallucination alleviation decoding methods is the computational burden. We evaluate the whole dataset inference time (seconds) and peak GPU memory (MB) on the LLaVA-1.5 7B under the POPE adversarial setting, as shown in Table \ref{table_e}. Contrastive Decoding (CD) methods \cite{vcd, icd} involve constructing distorted raw inputs, resulting in \textbf{twice} the inference complexity. OPERA \cite{opera} is based on beam-search decoding and maintains a set of beams to enlarge the candidate range. Additionally, roll back mechanism in the retrospection-reallocation strategy further exacerbates computational complexity. Our SID induces vision-and-text association hallucinations by pruning \textit{large-ratio} attention-important tokens in the \textit{early layers}, 
which greatly reduces the inference time of CD up to $\sim$30$\%$.

\noindent\textbf{Hyperparameter Sensitivity.} Beyond the sensitivity analyses in Figure  \ref{fig-i}, we validate the robustness of SID concerning $\alpha$ and $\beta$ of Equation \ref{eq2} and \ref{eq3}, compared to the contrastive decoding methods (i.e., VCD) on LLaVA-1.5 7B. From Figure  \ref{figab} (left), it is evident that as $\alpha$ decreases, the contrastive decoding mechanism diminishes. However, SID still achieves pleasant results, while VCD degrades close to Sampling when $\alpha$=0.1, as the CT$^2$S strategy induces informative \textit{vision-and-text association} hallucinations. When $\alpha$ increases, VCD degrades to some extent because holistic input disturbance does not always trigger contextual-related hallucination and might exacerbate uncertainty noise. Regarding $\beta$, a larger $\beta$ indicates more aggressive truncation of the output vocabulary. Figure  \ref{figab} (right) shows that VCD's performance heavily relies on large $\beta$ to retain only high-probability tokens. With mild or no adaptive plausibility constraint (Equation \ref{eq3}), VCD performs worse than the sample decoding strategy due to output logits influenced by distorted visual inputs. Ours is robust to the $\beta$ setting as the CT$^2$S strategy induces discriminative contrastive logits to generate plausible tokens.

\begin{table}[ht] \centering 
\caption{\textbf{Results on Larger-scale Backbones.} Sampling decoding is adopted and results average of three running times.}
\begin{tabular}{lcccl}
\hline
             & \multicolumn{2}{c}{\textbf{POPE}} & \multicolumn{2}{c}{\textbf{CHAIR}}                \\ \hline
Methods      & Accuracy        & F1 Score        & \  \  \   C$_S$ \  \  \                          &\  \  \   C$_I$ \  \  \             \\ \hline
LLaVA-1.5    & 81.60           & 80.31           & 49.6                              & 16.1          \\
+VCD         & 82.67           & 81.46           & 46.7                              & 16.4          \\
+OPERA       & 82.32           & 81.10           & \textbf{43.3}                     & 13.6          \\
+\textbf{Ours}        & \textbf{84.75}  & \textbf{83.17}  & 43.5                              & \textbf{12.7} \\ \hline
InstructBLIP & 77.26           & 79.23           & 50.8                              & 19.7          \\
+VCD         & 79.77           & 80.27           & 47.9                              & 17.6          \\
+OPERA       & 80.31           & 80.91           & 42.5                              & 14.3          \\
+\textbf{Ours}        & \textbf{81.97}  & \textbf{82.21}  & \textbf{41.7}                     & \textbf{13.3} \\ \hline
\end{tabular}
\label{table_large}
\end{table}

\noindent\textbf{Larger-scale LVLM Backbones.} 
We validate the effectiveness of SID in terms of 13B scale backbones on LLaVA-1.5 and InstructBLIP architectures. We choose POPE \cite{pope} and CHAIR \cite{chair} to validate the hallucination issues in both discrimination and open-end generation tasks. Table \ref{table_large} shows that SID remains effective as backbone networks scale up.

\begin{figure*}[]
  \centering
  \includegraphics[width=0.7\textwidth]{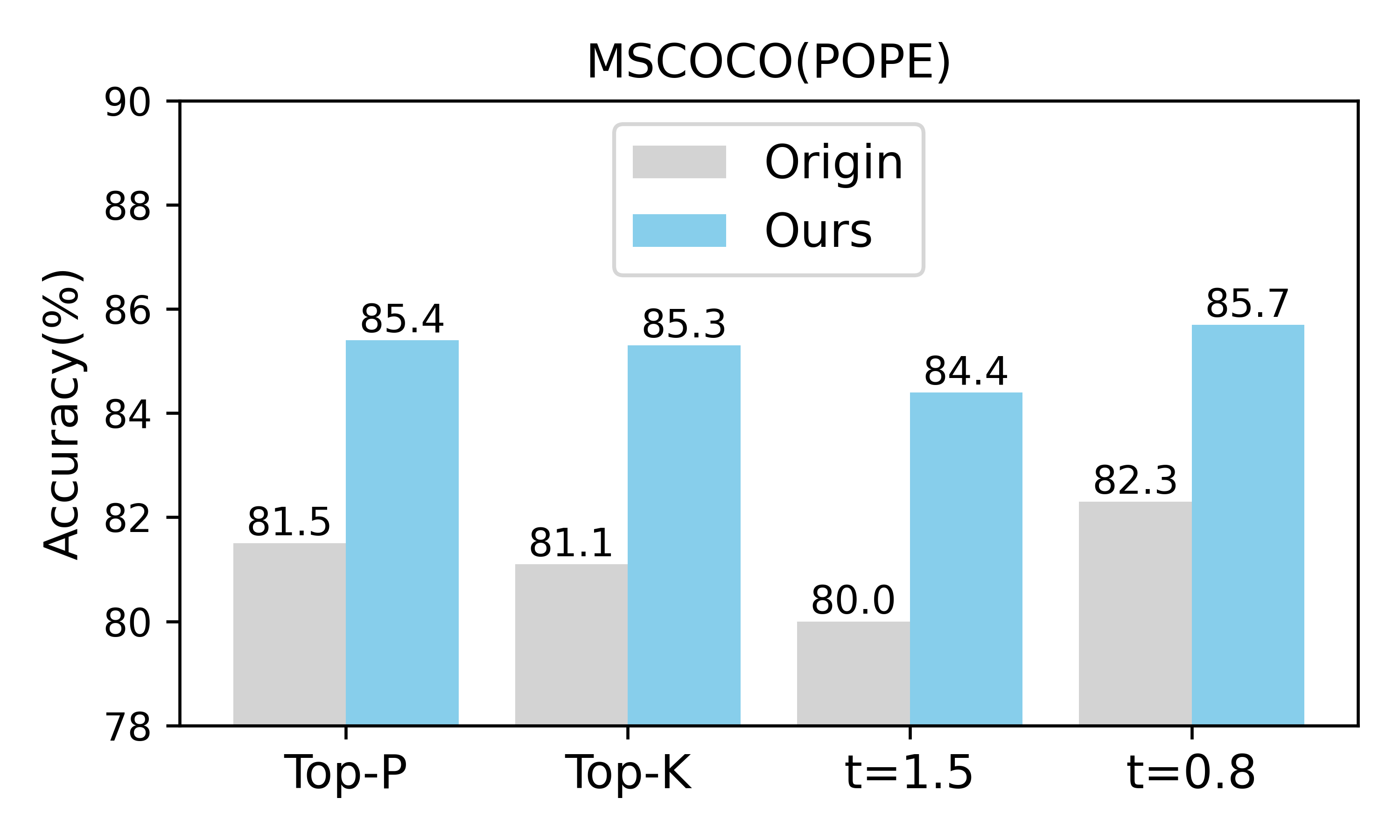}
  \caption{\textbf{Results of different decoding strategies.} }
  \label{fig_others}
\end{figure*}

\noindent\textbf{Adopting Other Decoding Strategies.} 
Meanwhile, besides direct sampling and greedy decoding, we conduct experiments on LLaVA-1.5 7B using the MSCOCO dataset with various decoding strategies, including Top-p sampling ($p$=0.9), Top-k sampling ($k$=50), Top-k sampling with varying temperature ($k$=50, $t$=1.5 and 0.8). Figure \ref{fig_others} shows that, regardless of the sampling strategy adopted, the application of SID consistently helps to alleviate hallucinations and improve the overall performance of LVLMs. This consistency highlights the versatility and effectiveness of SID across different sampling strategies.

\noindent\textbf{Visual Enhancing Decoding Strategy.} 
Although LVLMs can accurately recognize visual elements, LVLMs have difficulty fully interpreting those elements in the context of the input cue and effectively linking that recognition to their internal knowledge. We follow Visual Description Grounded Decoding (VDGD) \citep{vdgd} by first generating a detailed description of the image and appending it as a prefix to the instruction. The prompt template is adopted from \citep{vdgd}: 
\texttt{<image> I have been given this image to complete the task described as: {inst}. To help me complete the task, describe the given image in detail. In the case of real-world scenes, please include all foreground and background objects in the description, their properties (like color, shape, etc.), their relations with other objects, their count, and all other components in the image. In case of non-real-world scenes, like charts, graphs, tables, etc., please describe the table, mention all numbers (if any), mention the written text, and all other details.} Experiments are performed on hallucination evaluation benchmarks( i.e., POPE and CHAIR) and the general ability benchmark (i.e., MMbench). We re-implement VDGD based on official codes on LLaVA-1.5 7B. 
Table \ref{vdgd} demonstrates the effectiveness of VDGD \citep{vdgd} in LVLM's hallucination alleviation and general reasoning ability. However, the grounding visual descriptions, generated by LVLMs themself, enhance the visual perception reasoning capabilities while might inevitably contain hallucinations. Therefore, VDGD is inferior in the POPE ($\textbf{adversial}$) subset, which prioritizes \textbf{co-occurring} objects which are not present in the image. Meanwhile, 
VDGD shares somewhat similar motivations in enhancing vision information via Equation \ref{eq6} as we analyzed.
The experiments in Table \ref{figadd} are consistent with the above results, indicating that boosting the vision information is effective in mitigating hallucinations but is less effective in complex (i.e., adversarial) environments.

\begin{table}[ht] \centering
\begingroup
\caption{\textbf{Comparisons with Visual Enhancing Strategy (VDGD).} $^{\star}$ denotes employing greedy decoding strategy. }
\vspace{-0.2cm}
\begin{tabular}{cccccc}
\hline
       & \begin{tabular}[c]{@{}c@{}}POPE$\uparrow$\\ (random)\end{tabular} & \begin{tabular}[c]{@{}c@{}}POPE$\uparrow$\\ (adversarial)\end{tabular} & CHAIRs$\downarrow$        & CHAIRi$\downarrow$        & MMbench$\uparrow$       \\ \hline
Greedy & 88.8                                                    & 79.1                                                         & 49.6          & 14.4          & 64.4          \\
OPERA & 88.9                                                    & 79.2                                                         & 45.2          & 12.7          & 64.4          \\
VDGD$^{\star}$ & 89.0                                                    & 79.4                                                         & 46.7          & 13.7          & \textbf{65.2}          \\
SID$^{\star}$ & \textbf{89.3}                                           & \textbf{83.3}                                            & \textbf{44.2} & \textbf{12.2} & 65.0 \\ \hline
\end{tabular}
\label{vdgd}
\endgroup
\end{table}

\section{Discussion}

\begin{table}[] 
\centering \small
\caption{\textbf{Analyses of Different Token Selection Strategies} with POPE on MSCOCO dataset and CHAIR metrics. We select the high importance scores (Equation \ref{c5_eq5}) of vision tokens (-\textbf{Top}) and random vision tokens (-\textbf{Random}) for contrastive decoding. Experiments are conducted on LLaVA-1.5 7B.}
\begin{tabular}{l|cccc|cc}
\hline
\multicolumn{1}{c|}{\multirow{2}{*}{\textbf{Setting}}} & \multicolumn{2}{c|}{\textit{Random}}     & \multicolumn{2}{c|}{\textit{Adversarial}} & \multirow{2}{*}{CHAIRs $\downarrow$} & \multirow{2}{*}{CHAIRi $\downarrow$} \\ 
\multicolumn{1}{c|}{}                                  & Accuracy $\uparrow$ &  \multicolumn{1}{c|}{F1 Score $\uparrow$} & Accuracy $\uparrow$           & F1 Score $\uparrow$           &                         &                         \\ \hline
\textbf{Greedy}                                        & 88.8     & 88.6                          & 79.3               & 80.9               & 49.6                    & 14.4                    \\\cdashline{1-7} 
Ours                                          & 89.3     & 89.5                          & 83.3               & 82.5               & 44.2                    & 12.2                    \\
-High                                         &  87.0   &  87.3                         & 76.5               & 79.4               & 57.9                   & 25.6                    \\
-Random                                          & 88.4    & 87.2                         &  80.9             &  81.5             &  48.6                       & 13.5                        \\
AVISC                                          & 88.4     &88.1                &79.8       &80.5               &  45.3                       &  14.7                      \\\hline

\textbf{Sampling}                                        & 84.9     & 83.2                          & 78.7               & 78.9               & 51.3                    & 16.8                    \\\cdashline{1-7} 
Ours                                          & 88.8     & 88.7                          & 82.6               & 82.1               & 45.0                    & 11.7                    \\
AVISC                                          & 87.9   &  87.9          & 77.5             &  79.6              &  46.6                       &  12.5            \\\hline

\end{tabular}
\label{table_dif}
\end{table}

The core motivation is based the Context and Text-aware Token Selection (CT$^2$S) strategy. Here, we further analyze the efficacy of token selection strategies. Concretely, To validate the effectiveness of SID in selecting low attention scores to induce vision-and-text association hallucination, we further conduct quantitative experiments under different vision token selection strategies with the same preserved vision token number and Layer $i$=3 as ours. Table \ref{table_dif} shows that vision tokens with high attention scores degrade obviously, as it does not amplify contextual hallucinations rather than retain original important information. Contrastive decoding does not benefit from subtracting hallucinations amplified by the disturbed inputs rather than suffers from the potential disturbance noise.
% These experiments results further demonstrate the rationality of vision token importance sores.
% Meanwhile, the token-level disturbance also induces uncertainty noise, resulting in the inferior performance to greedy decoding. 
Selecting random vision tokens brings improvements in the \textit{adversarial} setting because randomly selected vision tokens amplify the over-reliance on statistical bias and language priors, similar to Vision CD \cite{vcd} and Instruction CD \cite{icd}. However, token-level random disturbance also induces uncertainty noise, resulting in the inferior performance in the \textit{random} setting to greedy decoding. Moreover, AVISC \cite{avisc}, in contrast to ours, preserves outlier high attention tokens (named 'blind token') and substracts output logits to counteract the overemphasis of 'blind token.' In this way, AVISC promotes balanced consideration of all tokens to alleviate hallucinations. However, Table \ref{table_dif} illustrates that Top-100 vision tokens with high attention scores can largely maintain the original performance. 'blind token' tends to have a high probability of target class logits, and contrastive decoding does not improve the target class's probability while might bring extra noise. Table \ref{table_dif} indicates AVISC still degrades the greedy decoding to some extent, which indicates the attentional vision re-calibration strategy of AVISC induces some annoying noise. Overall, these experiments further validate the rationality of our token selection strategy based on attention sores.

\section{Chapter Summary}

In Chapter \ref{Chapter5}, we firstly re-think contrastive decoding in LVLMs and empirically find that vision-and-text-agnostic input disturbances in CD do not always amplify desired hallucinations rather than induce potential uncertainty noise. To mitigate these issues, we propose a training-free decoding strategy named Self-Introspective Decoding (SID). 
By developing Context and Text-aware Token Selection (CT$^2$S) strategy,
SID amplifies \textit{vision-and-text association} hallucinations to guide LVLMs in contrastive decoding, thereby improving faithfulness. Extensive experiments validate the effectiveness and robustness of SID. Aligned with the thesis’s focus on robust open-world reasoning, SID’s self-corrective mechanism extends Chapter \ref{Chapter3}’s cross-primitive compatibility robustness and Chapter \ref{Chapter4}’s crossmodal knowledge transfer harmonization to introspective hallucination suppression of multimodal LLMs. 
As for Future Work: \textbf{1)} As the pruning ratios and layer are set manually, we consider training the external network to automatically determine optimal hyperparameters, inspired by \cite{diffrate}. In addition, to enhance the interpretability of hallucination alleviations, we consider resorting to pre-trained analysis networks to intuitively locate spurious related vision regions. \textbf{2)} Moreover, given that SID amplifies fine-grained hallucinations, we consider leveraging the CT$^2$S strategy to automatically construct high-quality negative instruction for robust visual instruction tuning rather than relying on expensive GPT-4 \cite{rit, hadpo}. Note that the self-generated hallucination dataset ensures $\textit{style consistency}$, which is crucial for preference learning \cite{hadpo}.
% In addition, Instruction tuning with different datasets might comprise 

\chapter{Conclusion and Suggestions for Future Research}

\label{Chapter6}

\section{Conclusion}
\label{c6.1}

Despite the success of neural networks, modern neural networks remain inadequate for open-world deployment due to limitations in flexibility, multimodal robustness, and trustworthiness. 
This thesis bridges these gaps by addressing three critical challenges in multimodal learning: (1) compositional robustness in multiple modality primitives, (2) efficient cross-modal knowledge transfer under modality incompleteness, and (3) hallucination suppression in multimodal large language models (LLMs). The research framework is illustrated in Figure \ref{c1_framework}.
In summary, this thesis is mainly composed of three following parts:

\begin{itemize}

\item To enhance modality composition generalization robustness, we revisit the primitive prediction approach and propose a novel method, termed Progressive Cross-primitive Compatibility (ProCC), to mimic the human learning process for OW-CZSL tasks. 
Specifically, the cross-primitive compatibility module \emph{explicitly} learns to model the interactions of state and object features with the trainable memory units, which efficiently acquires cross-primitive visual attention to reason high-feasibility compositions, \emph{without} the aid of external knowledge.
Moreover, to alleviate the invalid cross-primitive interactions, especially for partial-supervision conditions (pCZSL), we design a progressive training paradigm to optimize the primitive classifiers conditioned on pre-trained features in an easy-to-hard manner. 
Extensive experiments on three widely used benchmark datasets demonstrate that our method outperforms other representative methods on both OW-CZSL and pCZSL settings by large margins.

\item To ensure robustness under modality missing, we focus on studying crossmodal knowledge distillation to handle modality-missing situations. We empirically reveal that the modality gap, i.e., modality imbalance and soft label misalignment, incurs the ineffectiveness of traditional KD methods in CMKD.
As a solution, we propose a novel \textbf{\underline{C}}ustomized \textbf{\underline{C}}rossmodal \textbf{\underline{K}}nowledge \textbf{\underline{D}}istillation (C$^2$KD).
Specifically, to alleviate the modality gap, the pre-trained teacher performs bidirectional distillation with the student to provide customized knowledge. 
The On-the-Fly Selection Distillation(OFSD) strategy is applied to selectively filter out the samples with misaligned soft labels, where we distill cross-modal knowledge from non-target classes to avoid the modality imbalance issue. 
To further provide receptive cross-modal knowledge, proxy student and teacher, inheriting unimodal and cross-modal knowledge, is formulated to progressively transfer cross-modal knowledge through bidirectional distillation. 
Experimental results on audio-visual, image-text, and RGB-depth datasets demonstrate that our method can effectively transfer knowledge across modalities, achieving superior performance against traditional KD by a large margin.

\item To balance modality priors to mitigate hallucinations of multimodal LLMs, 
we first re-think contrastive decoding in LVLMs and empirically find that vision-and-text-agnostic input disturbances in CD do not always amplify desired hallucinations rather than induce potential uncertainty noise. To mitigate these issues, we propose a training-free decoding strategy named Self-Introspective Decoding (SID). 
By developing Context and Text-aware Token Selection (CT$^2$S) strategy,
SID amplifies \textit{vision-and-text association} hallucinations to guide LVLMs in contrastive decoding, thereby improving faithfulness. This train-free approach reduces hallucinations by 12–20$\%$ on metrics like POPE and CHAIR while cutting inference costs by 30$\%$ compared to methods like VCD \cite{vcd} and ICD \cite{icd}. Crucially, SID preserves LVLMs’ general abilities, as evidenced by strong MME and MMBench scores. By rebalancing modality priors without compromising functionality, SID advances modality-level robustness, ensuring trustworthy outputs in open-world deployment.

\end{itemize}

\section{Suggestions for Future Research}
\label{c6.2}

Developing robust machine learning for multiple modalities is not a trivial task. Beyond the above challenges, the open-world ever-changing environments also have has challenges such as adversarial adversarial shifts, out of distribution, novel class discovery, and more.
To achieve more flexible and efficient machine learning across multiple modalities, future research can explore at least the following three directions:

\subsection{Multimodal Test-time Adaptation} 
\label{c6.2.1}
The degradation of multimodal inputs under extreme environmental conditions (e.g., night, snowy, or foggy settings) introduces severe cross-modal misalignment, where modality-specific corruption patterns (e.g., obscured visuals vs. stable LiDAR signals) create imbalanced feature distributions. Traditional multimodal Test-time Adaptation (TTA) methods \cite{yang2024rfra, shin2022mm, cao2024reliable} partially mitigate this by globally tuning network parameters, yet they overlook two fundamental issues: fine-grained modality interactions and cross-modal consistency preservation. The first issue emphasizes the heterogeneous degradation levels across modalities (e.g., 73$\%$ pixel loss in RGB vs. 12$\%$ point cloud sparsity in LiDAR) demand dynamic modality weighting rather than uniform adaptation. The latter issue underlines current approaches that fail to enforce semantic coherence between corrupted and intact modalities during adaptation, risking error propagation. For future work, I think the robust multimodal test-time adaptation frameworks must prioritize \textit{hierarchical} adaptation frameworks that dynamically estimate modality-specific and modality-general degradations via entropy minimization, enforce semantic consistency through contrastive alignment of robust primitives (e.g., edge features in vision, spectral peaks in audio), and recalibrate fusion weights in real-time based on modality reliability scores. Such approaches would enable models to adaptively balance multimodal information under open-world volatility, bridging the gap between controlled laboratory performance and real-world resilience.

\subsection{Task-Aware Adaptation of Multimodal LLMs}
\label{c6.2.2}
The rapid evolution of large-scale vision-language models (VLMs) such as CLIP \cite{clip}, Flamingo \cite{flam}, and LLaVA \cite{llava} has unlocked unprecedented zero-shot generalization capabilities, positioning them as foundational tools for open-world multimodal learning. However, adapting these models to downstream tasks—from specialized domains like medical diagnostics to dynamic environments like autonomous navigation—requires overcoming critical barriers: the inherent misalignment between their pretraining objectives (e.g., generic image-text matching) and task-specific goals (e.g., fine-grained anomaly detection), the computational impracticality of full fine-tuning for billion-parameter architectures, and the inability to dynamically prioritize modalities based on contextual relevance (e.g., emphasizing thermal over RGB data in low-light conditions). To harness their potential, future work must innovate task-aware adaptation frameworks that inject domain knowledge through lightweight neural modules, optimize modality interactions via dynamic gating mechanisms, and enforce causal invariance to suppress spurious correlations. Early successes, such as adapter-based tuning \cite{hu2022lora, NEURIPS2023_3340ee1e, dora} methods improving rare-class detection in satellite imagery by a large margin, underscore the viability of such approaches. Yet achieving human-level adaptability—where a single LVLM seamlessly transitions from interpreting ambiguous medical queries to generating context-aware robot instructions—demands unifying their embodied knowledge with symbolic reasoning, bridging the gap between static pretrained information and the dynamic demands of downstream deployments. This direction not only amplifies the utility of foundation models but also advances the thesis’s broader vision of robust, open-world multimodal systems.

\subsection{Multimodal Agent as Experts} 
\label{c6.2.3}
Large Language Models (LLMs), with their expansive parameter scales and encyclopedic knowledge, are revolutionizing how neural networks adapt to complex tasks. By leveraging their deep understanding of contextual relationships and procedural reasoning, LLMs can dynamically guide the optimization, regularization, and architectural adjustments of neural networks. This capability enables adaptive learning systems that respond intelligently to diverse data distributions, resource constraints, and performance objectives. Recent advancements in LVLMs-empowered multimodal agents, such as those by \cite{xie2024large}, \cite{liu2024llava}, and \cite{zhang2023appagent}, have demonstrated remarkable proficiency in tool utilization, embodied AI, and cross-modal reasoning. For instance, Llava-plus \cite{liu2024llava} showcases LVLMs' ability to interpret visual inputs, generate executable code for robotic manipulation, and self-correct actions through iterative feedback. Similarly, Appagent \cite{zhang2023appagent} highlights how multimodal agents can autonomously navigate mobile interfaces by combining screen comprehension, language parsing, and gesture prediction. These innovations underscore the potential of LVLMs to act as versatile "experts" capable of orchestrating intricate workflows across domains. For future work, we want to propose a paradigm where multimodal agents serve as \textit{intelligent coordinators} for multimodal learning systems, enhancing their flexibility and trustworthiness. As coordinators, these agents perform three critical roles. For example, By analyzing real-time inputs (e.g., visual, textual, sensor data), LVLMs agents autonomously reconfigure model architectures, select optimal pretrained sub-networks, or adjust hyperparameters to match evolving task requirements. With the development of multimodal agents, I believe that human developers are totally free of designing complex algorithms for the multimodal learning system.

% \input{Chapter_7}

%\appendix

%\nocite{astl2,perry1,pollardsag1}
%\nocite{einstein,latexcompanion,knuthwebsite}

\bibliographystyle{plain}
\bibliography{reference}

\end{document}